\documentclass[10pt,journal,compsoc]{IEEEtran}

\ifCLASSINFOpdf
  \usepackage[pdftex]{graphicx}
\else
  \usepackage[dvips]{graphicx}
\fi

\usepackage{amsmath}
\usepackage{amssymb}
\usepackage{amsfonts} 

\ifCLASSOPTIONcompsoc
  \usepackage[caption=false,font=footnotesize,labelfont=sf,textfont=sf]{subfig}
\else
  \usepackage[caption=false,font=footnotesize]{subfig}
\fi

\usepackage{url}
\usepackage{color}
\usepackage{colortbl}
\usepackage{soul}
\usepackage{tabularx}
\usepackage{hyphenat}
\usepackage{longtable}
\usepackage[breaklinks]{hyperref}

\usepackage{float}

\usepackage{etoolbox}
\makeatletter
\patchcmd{\@makecaption}
  {\scshape}
  {}
  {}
  {}
\makeatother

\usepackage{algorithm}      
\usepackage[noend]{algpseudocode} 
\usepackage{cleveref} 

\usepackage{pifont}
\newcommand{\cmark}{\ding{51}}%

\definecolor{LightRed}{rgb}{1,0.92,0.92}
\definecolor{LightOrange}{rgb}{1,0.95,0.88}
\definecolor{LightYellow}{rgb}{1.0,1.0,0.84}
\definecolor{LightGreen}{rgb}{0.9,1.0,0.88}
\definecolor{LightCyan}{rgb}{0.9,1,1}
\definecolor{LightBlue}{rgb}{0.9,0.94,1}
\definecolor{LightIndigo}{rgb}{0.92,0.9,1}
\definecolor{LightMagenta}{rgb}{0.96,0.86,1}
\definecolor{DirtyWhite}{rgb}{0.96,0.96,0.96}

\usepackage[utf8]{inputenc} 
\usepackage[T1]{fontenc}    
\usepackage{url}            
\usepackage{booktabs}       
\usepackage{amsfonts}       
\usepackage{nicefrac}       
\usepackage{xcolor}         
\usepackage{graphicx}
\usepackage{multirow}
\usepackage{tikz}
\usepackage{wrapfig}
\usepackage{amsmath}
\usepackage{setspace}
\usepackage{makecell}
\usepackage{arydshln}
\usepackage{colortbl}
\usepackage{mathtools}
\usepackage{float}
\usepackage[figuresright]{rotating}
\usepackage{bm}
\usepackage{amsmath}
\usepackage{amssymb}
\definecolor{mycolor_blue}{HTML}{E7EFFA}
\definecolor{mycolor_green}{HTML}{E6F8E0}
\definecolor{mycolor_gray}{HTML}{ECECEC}
\definecolor{pearDark}{HTML}{2980B9}
\definecolor{lightblue}{rgb}{0.9,0.94,1}
\usepackage{graphicx} 		
\usepackage{pifont}
\usepackage{natbib}
\setcitestyle{numbers,square}
\usepackage{nccmath}

\usepackage{hyperref}
\usepackage{fontawesome5}
\newcommand{\Paper}[1]{\href{#1}{\faFilePdf}}
\newcommand{\Code}[1]{\href{#1}{\faGithub}}
\newcommand{\NA}{\textit{N/A}}
\newcommand{\faHuggingFace}{%
  \raisebox{-0.13em}{%
    \includegraphics[height=1em]{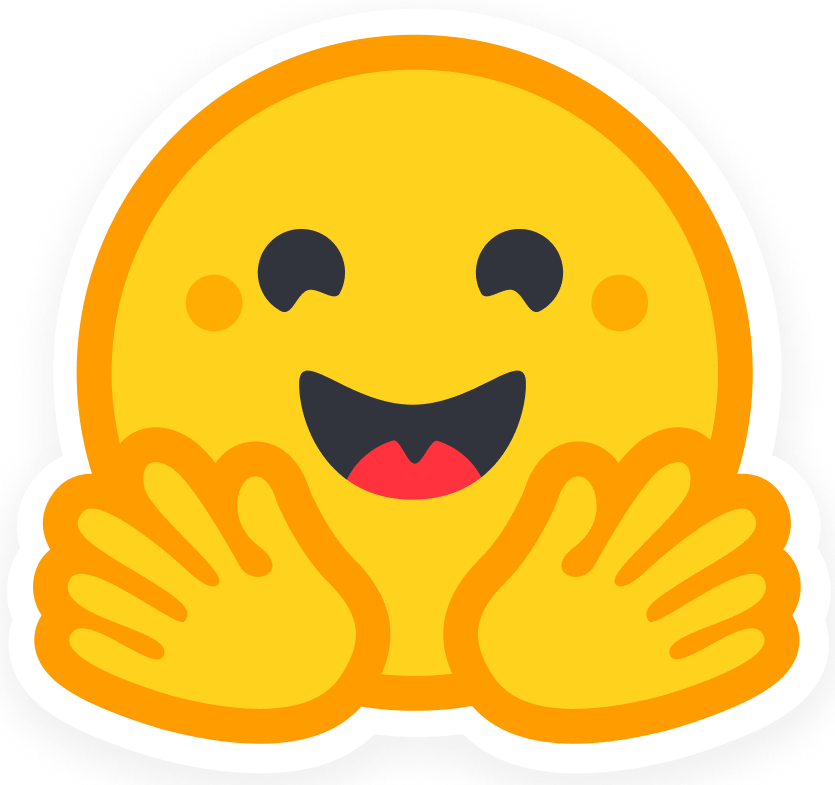}%
  }%
}

\begin{document}
%
\title{Reinforcement Learning-driven Diffusion Models: Language, Vision, and Beyond}

\markboth{IEEE Transactions on Pattern Analysis and Machine Intelligence, Manuscript Submission, September~2026}%
{Zhiyuan Ma \MakeLowercase{\textit{et al.}}: Efficient Diffusion Models: A Comprehensive Survey from Principles to Practices}

\author{
Zhiyuan Ma,
Jiabao Wei,
Tianyu Wang,
Hanqing Wang,
Jiadong Pan,
Haowen Xu,
Liangliang Zhao,
Ruixun Liu,
Yuhan Wang,
Yuwei Zhang,
Liang Feng,
Junhao Xiao,
Kaiyan Zhang,
Xuekai Zhu,\\
Daizong Liu,
Biqing Qi,
Yihao Liu,
Ganqu Cui,
Ning Ding,
Linfeng Zhang,
Jianjun Li,
Kun He,\\
Xiang Bai,~\IEEEmembership{Fellow,~IEEE},
Bowen Zhou,~\IEEEmembership{Fellow,~IEEE}

\IEEEcompsocitemizethanks{
\IEEEcompsocthanksitem Z. Ma, J. Xiao, J. Li, K. He, and X. Bai are with the Huazhong University of Science and Technology. (first author's e-mail: \{mzyth\}@hust.edu.cn). 

\IEEEcompsocthanksitem J. Wei is with the Beijing Institute of Technology. (e-mail: weijiabao@bit.edu.cn). Z. Ma and J. Wei contributed equally to this research.

\IEEEcompsocthanksitem T. Wang, Y. Wang, K. Zhang, N. Ding are with the Tsinghua University.

\IEEEcompsocthanksitem H. Xu, L. Feng, X. Zhu, L. Zhang are with the Shanghai Jiaotong University. Z. Ma, J. Wei and L. Zhang are project Leaders.

\IEEEcompsocthanksitem H. Wang is in Hong Kong University of Science and Technology (GZ).

\IEEEcompsocthanksitem J. Pan is with the University of Chinese Academy of Sciences.

\IEEEcompsocthanksitem L. Zhao is with the Fudan University \& Shanghai AI Laboratory.

\IEEEcompsocthanksitem R. Liu is with the Peking University.

\IEEEcompsocthanksitem Y. Zhang is with the Tongji University.

\IEEEcompsocthanksitem D. Liu is with the Wuhan University.

\IEEEcompsocthanksitem B. Qi, Y. Liu, G. Cui are with the Shanghai AI Laboratory.

\IEEEcompsocthanksitem B. Zhou is with the Tsinghua University \& Shanghai AI Laboratory. X. Bai and B. Zhou are co-corresponding authors. (xbai@hust.edu.cn, zhoubowen@tsinghua.edu.cn)
}

}


\IEEEtitleabstractindextext{%
\begin{abstract}
Diffusion models (DMs) have steadily demonstrated outstanding performance in high-fidelity, controllable generation across diverse scenarios, following the ``pretraining-finetuning'' paradigm established in LLMs.
Despite these advances, conventional DMs still face critical limitations in sophisticated reasoning, autonomous exploration of generation pathways, and adaptability to complex, human-aligned objectives.
Inspired by RLHF's success in LLMs, RL has recently emerged as a pivotal post-training mechanism to iteratively refine DMs through reward-driven trajectory optimization.
This paper presents the first comprehensive review of RL-driven DMs and  empirical analysis of how RL reshapes the diffusion process towards experience-centric, goal-directed generation.
We propose a novel technical panorama elucidating RL's multifaceted roles in diffusion-based language, multimodal, robotic, and scientific models—with granular focus on diffusion trajectory optimization and reward engineering.
Through extensive experiments on multiple task-specific benchmarks, we systematically evaluate representative methods to intuitively highlight the advantages of different RL post-training strategies.
Furthermore, we reveal and dissect three crucial bottlenecks that currently hinder the real-world deployment of RL-driven DMs: reward design fragility, exploration-exploitation trade-off in high-dimensional generation spaces, and training instability under sparse rewards.
Based on these perspectives, we provide a principled roadmap for the integration of RL into DMs and outline promising future research directions for robust, interpretable RL-diffusion systems.
We also construct a public repository to continuously track the latest developments in this rapidly evolving field: \textcolor{blue}{\url{https://github.com/MAIR-Lab-HUST/Awesome-RL-for-Diffusion-Models}}.
\end{abstract}
}

\maketitle

\setlength{\abovedisplayskip}{3.5pt}
\setlength{\belowdisplayskip}{3.5pt}

\IEEEdisplaynontitleabstractindextext

\IEEEpeerreviewmaketitle

\section{Introduction}
\label{sec:intro}

Diffusion models (DMs)~\cite{ho2020DDPM,rombach2022LDM,liu2022flow} have already emerged as a cornerstone paradigm in generative AI, demonstrating exceptional generation performance across a wide spectrum of domains, \textit{e.g.}, language, image, video, and audio.
This remarkable advancement stems not only from their inherent capacity to model complex data distributions but also from their adoption of the development pathway pioneered by LLMs, where the \textbf{\textit{``pretraining-finetuning''}} route has effectively guided DMs towards high-fidelity, controllable, and generalizable generation.
However, these conventional DMs are purely data-driven imitation learning, struggling to handle dynamic, multi-step decision problems and align with human preferences or task objectives, thereby limiting their potential in complex reasoning and specialized generation scenarios.
Specifically, the denoising process derived from imitation is primarily bound by static training-data distributions and the reconstruction-centric MSE loss, reaching a performance ceiling that prevents DMs from autonomously planning and exploring the optimal diffusion trajectory.

\begin{figure}[t]
\centering
\includegraphics[width=0.48\textwidth]{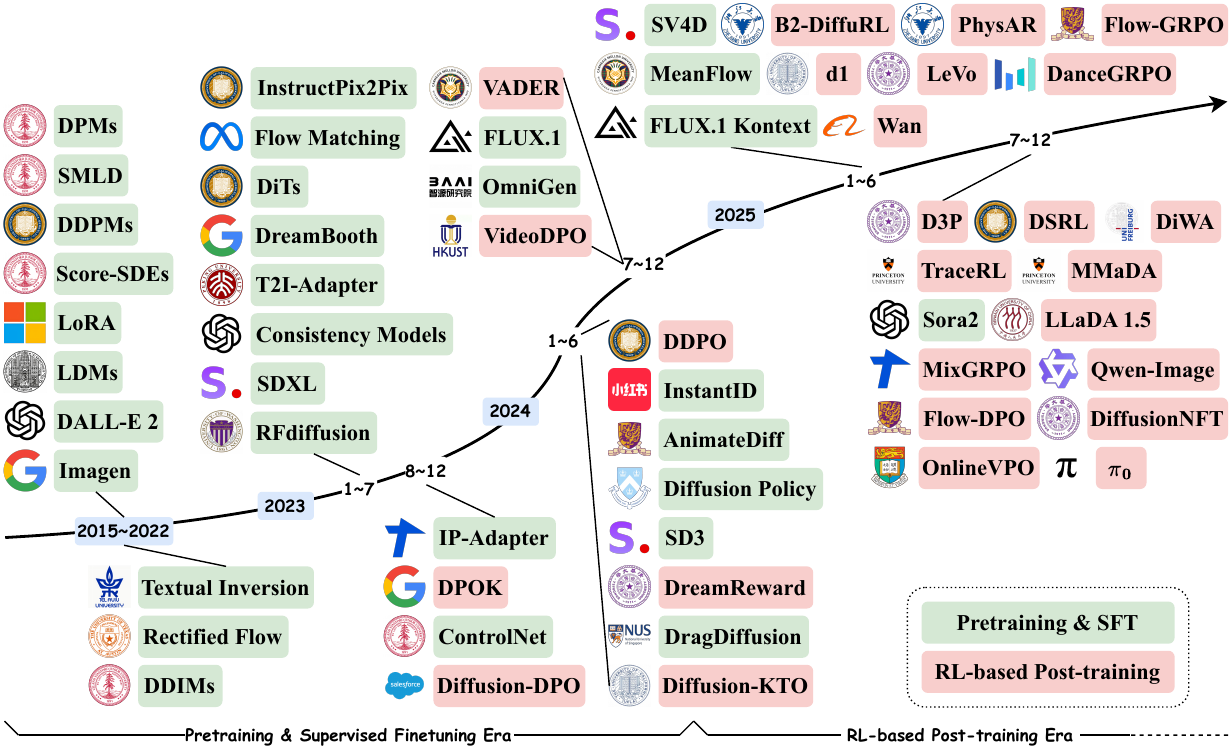}
\vspace{-1em}
\caption{A complete timeline of the development of DMs. At present, DMs have already passed the ``Pretraining $\&$ Supervised Finetuning'' era and entered the ``RL-based Post-training'' era. Our survey focuses on the latest development stage and reviews recent work on RL for DMs.
}
\vspace{-1.6em}
\label{fig:DM_develop}
\end{figure}
In recent years, reinforcement learning (RL)~\cite{sutton1998RL} has risen to prominence as a post-training strategy for human alignment in LLMs.
As shown in Fig.~\ref{fig:DM_develop}, RL is also progressively incorporated into DMs in an attempt to address the above challenges.
Widely utilized RL techniques such as Reinforcement Learning from Human Feedback (RLHF)~\cite{christiano2017RLHF} and Direct Preference Optimization (DPO)~\cite{rafailov2023direct} further finetune pre-trained foundational DMs to follow complex instructions and embody human preferences, thereby enhancing semantic coherence and aesthetic value while maintaining diversity.
More recently, a new trend has emerged: training DMs using reinforcement learning with verifiable rewards (RLVR), which generalizes DMs towards more specialized robot and science domains.
RLVR aims not merely to align behavior but to incentivize DMs' reasoning ability by incorporating \textbf{\textit{explicit reward signals}} for dynamic, rule-based optimization. Furthermore, RLVR is typically accompanied by large-scale RL, and specifically, Group Relative Policy Optimization (GRPO)~\cite{shao2024deepseekmath} can elicit sophisticated, long-form reasoning while markedly improving efficiency.
To sum up, RL has led DMs into a \textit{\textbf{``post-training era''}} which signals more intelligent, specialized, and versatile generation.


RL-driven DMs treat the denoising steps as the \textbf{\textit{``thinking''}} trajectory to implicitly perform reasoning.
This transformation manifests via three evolutionary shifts:
First, the driving force transitions from reliance on statistical distributions in static pretraining data to experience-driven dynamic interactions with external environment, enabling direct maximization of arbitrarily formalizable reward signals.
Second, the generation logic evolves from open-loop, single-path imitation to closed-loop planning, reflection, and self-correction.
RL empowers DMs to perform multi-trajectory scaling during inference and autonomously explore the optimal policy based on cumulative reward estimation, substantially enhancing capabilities in complex, long-horizon dependency tasks.
Third, the optimization objectives undergo fundamental reshaping—from minimizing simple MSE loss to maximizing specific reward functions—thereby achieving alignment with human preferences or task utility.

Meanwhile, further scaling RL-driven DMs for practical deployments introduces new constraints beyond computational resources, including reward engineering, algorithm efficiency, and exploration-exploitation trade-offs.
The pathways of effectively incorporating RL into DMs to achieve high-level generative intelligence and real-world utility remain significant, unanswered issues.
Hence, we think that it is imperative to revisit and systematize this emerging domain and explore more scalable RL strategies to advance DMs towards more intelligent and general-purpose generators.
Here we provide a comprehensive review as follows:
\begin{itemize}
    \item We first introduce the basic principles of standard (non-RL) DMs (\S~\ref{preliminary:Standard_DM}) and the preliminary definitions of RL-driven DMs (\S~\ref{preliminary:DM_RL}) to illustrate the Markovian theoretic commonality of DMs and RL.
    \item We introduce the paradigm of diffusion-based LLMs (dLLMs) (\S~\ref{sec:dLLMs}), including the masking/unmasking process and the ELBO objective, and review recent literature on RL-driven dLLMs along two orthogonal axes (\S~\ref{sec: RL's role in dLLMs}): trajectory level and reward density.
    \item We review recent RL-driven diffusion-based multimodal models (dMMs) from two different perspectives: \textbf{\textit{1)}} RL technology development ($\S$~\ref{sec:RL's role in dMMs}); and \textbf{\textit{2)}} RL-related multimodal tasks ($\S$~\ref{sec:RL4dMMs applications}).
    \item We discuss the foundations of RL for diffusion in robotics (\S~\ref{sec:pre_RL4Robo}) and review two mainstream diffusion-based embodied intelligence studies, including diffusion policy-based
    RL training (\S~\ref{sec:app_RLinDP}) and generative experts in Vision-Language-Action models (\S~\ref{sec:RLindVLA}).
    \item We review recent advances in RL-driven diffusion-based scientific models from a novel perspective of data modality, including discrete molecular sequence ($\S$~\ref{sec:Discrete Sequence Generation}), continuous 3D structure ($\S$~\ref{sec:Continuous 3D Structure Generation}), periodic crystal structure ($\S$~\ref{sec:Periodic Structure Generation}), continuous physics-constrained field ($\S$~\ref{sec:Continuous Fields}), and scientific imaging ($\S$~\ref{sec:Scientific Imaging}).
    \item Finally, we discuss future research directions for robust, interpretable RL-diffusion systems.
\end{itemize}

This survey serves as a roadmap, highlighting potential avenues for future research and applications, and fostering a deeper comprehension of the challenges and opportunities that lie ahead in the promising domain of RL-driven DMs.

\section{Background and Preliminaries}
\label{sec:preliminary}

\subsection{Standard Diffusion: From Discrete to Continuous}
\label{preliminary:Standard_DM}

Diffusion models (DMs) have emerged as the core backbone of modern generative AI. To provide context for RL-driven enhancements, we first briefly review their theoretical foundations, ranging from discrete-time probabilistic models to continuous-time score and flow generalizations.

\subsubsection{Discrete-Time Paradigm: DDPMs}
\label{sec:DDPMs}
Denoising Diffusion Probabilistic Models (DDPMs)~\cite{ho2020DDPM} generate data through a stochastic process analogous to thermodynamic diffusion, comprising a learning-free forward \textbf{\textit{diffusion}} and a learnable reverse \textbf{\textit{denoising}} process.
Specifically, in the forward process, the initial data $\mathbf{x}_0$ is gradually transformed to standard Gaussian noise $\mathbf{x}_T\sim\mathcal{N}(0,\mathbf{I})$ as
\begin{equation}
q(\mathbf{x}_t|\mathbf{x}_{t-1}):=\mathcal{N}(\mathbf{x}_t; \sqrt{1-\beta_t}\mathbf{x}_{t-1},\beta_t \mathbf{I}).
\end{equation}
Based on this, the diffusion process from $\mathbf{x}_0$ to $\mathbf{x}_t$ can be succinctly expressed via a single-step cumulative formula: $q(\mathbf{x}_t|\mathbf{x}_0):=\mathcal{N}(\mathbf{x}_t;\sqrt{\bar{\alpha}_t}\mathbf{x}_0,(1-\bar{\alpha}_t)\mathbf{I})$, where $\alpha_t=1-\beta_t$ and $\bar{\alpha_t}=\prod_{s=1}^t\alpha_s$.
Moreover, the reverse process recovers samples from noise $\mathbf{x}_T\sim\mathcal{N}(0,\mathbf{I})$ by approximating inverse transitions $p_\theta(\mathbf{x}_{t-1}|\mathbf{x}_t):=\mathcal{N}(\boldsymbol{\mu}_\theta(\mathbf{x}_t,t), \boldsymbol{\Sigma}_\theta(\mathbf{x}_t,t))$.
Training minimizes the noise-prediction objective, which can be simplified to a mean square error form as
\begin{equation}
L_{\text{simple}}(\theta)=
\mathbb{E}_{t,\mathbf{x}_0,\boldsymbol{\epsilon}}
\big[
\lVert
\boldsymbol{\epsilon}-\boldsymbol{\epsilon}_\theta(\sqrt{\bar{\alpha}_t}\mathbf{x}_0+\sqrt{1-\bar{\alpha}_t}\boldsymbol{\epsilon},t)
\rVert_2^2
\big].
\end{equation}
Latent Diffusion Models (LDMs)~\cite{rombach2022LDM} further apply this process in a compressed latent space to enhance efficiency.
\begin{figure}[t]
\centering
\includegraphics[width=1\linewidth]{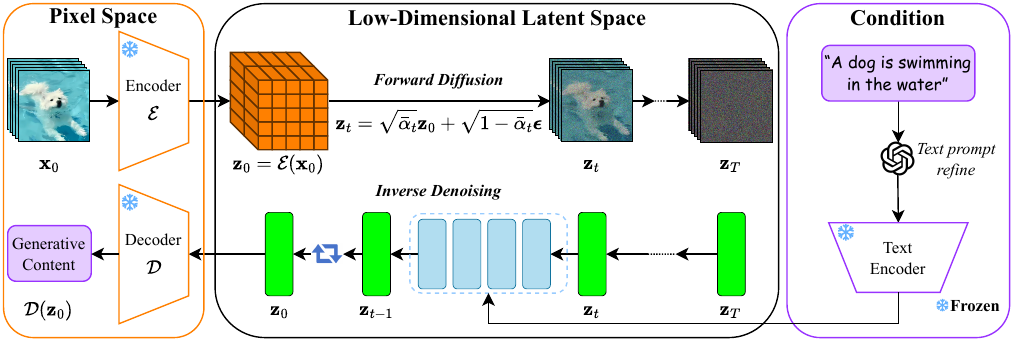}
\vspace{-2.2em}
\caption{
The overview of text-guided visual content generation.
}
\vspace{-1.5em}
\label{fig:overview_ldm}
\end{figure}
Fig.~\ref{fig:overview_ldm} visualizes the text-guided generation process of LDMs.

\subsubsection{Continuous Generalization: Score and Flow Matching}
\label{sec:SM_FM}
DDPMs~\cite{ho2020DDPM} can be theoretically unified by score matching (SM)~\cite{song2020SDE} and flow matching (FM)~\cite{lipman2022flow}, which both treat generation as continuous-time dynamics.
Notably, the noise-prediction loss is equivalent to estimating the score function $\nabla_{\mathbf{x}_t}\log q(\mathbf{x}_t)$. The reverse-time SDE recovers data via
\begin{equation}
\text{d}\mathbf{x}_t=\big[f(\mathbf{x}_t,t)-g^2(t)\nabla_{\mathbf{x}_t}\log q(\mathbf{x}_t)\big]\text{d}t+g(t)\text{d}\bar{\textbf{w}}.
\end{equation}
To generalize this, flow matching constructs a deterministic probability flow via a vector field $v_t$, transforming a prior $p_0$ to data $q(x)$. During training, it minimizes the distance between the learned vector field $v_t(x;\theta)$ and a target $u_t(x|x_1)$ constructed via Conditional Flow Matching (CFM):
\begin{equation}
    \mathcal{L}_{\text{CFM}}(\theta)=\mathbb{E}_{t,x_1\sim q, x\sim p_t(x|x_1)}\|v_t(x;\theta)-u_t(x|x_1)\|_2^2.
\label{eq:cfm}
\end{equation}
Rectified Flow~\cite{liu2022flow} simply defines the ``noised'' data $x_t$ as $x_t=(1-t)x_0+tx_1$ and directly regresses the velocity field $v_t(x;\theta)$ via eq.~\ref{eq:cfm}, where the target velocity field is $x_1-x_0$.

\subsection{Reinforcement Learning-driven Diffusion Models}\label{preliminary:DM_RL}
\subsubsection{Why Do Diffusion Models Require RL?}

RL provides a principled way to optimize DMs via rewards by treating the model as a policy guided by human feedback or task-specific signals. This enables explicit alignment of generation with desired objectives.
An alternative is reward-weighted finetuning, which biases the model towards high-reward samples via a learned reward estimator $\hat{r}(x)$:
\begin{equation}
    \mathbb{E}_{t,\mathbf{x}_0\sim q(\mathbf{x}_0), \epsilon\sim \mathcal{N}(0,\mathbf{I}) }[\hat{r}(x_0) \|\boldsymbol{\epsilon}-\boldsymbol{\epsilon}_{\theta}(\sqrt{\bar{\alpha}_t}\mathbf{x}_0+\sqrt{1-\bar{\alpha}_t}\boldsymbol{\epsilon},t)\|_2^2].
\end{equation}
However, it struggles to discover out-of-distribution (OOD) solutions and typically requires paired data ${c^i, x^i}$ for conditional models, limitations that RL avoids by directly optimizing the reward functions. Similarly, rejection sampling selects high-reward outputs from a pre-trained model but is effective only if such samples already have sufficient probability, whereas RL actively updates the model parameters to generate high-reward samples efficiently.

Beyond these alternatives, RL can directly optimize non-differentiable or sparse rewards, align with the sequential structure of diffusion by treating each denoising step as a policy action, and allow flexible reward design that carries human feedback or task-specific objectives. By combining exploration and reward-driven post-training, RL effectively guides DMs towards better alignment with downstream tasks, OOD generalization, and autonomous reasoning.
 
\begin{figure}[t]
\centering
\includegraphics[width=1\linewidth]{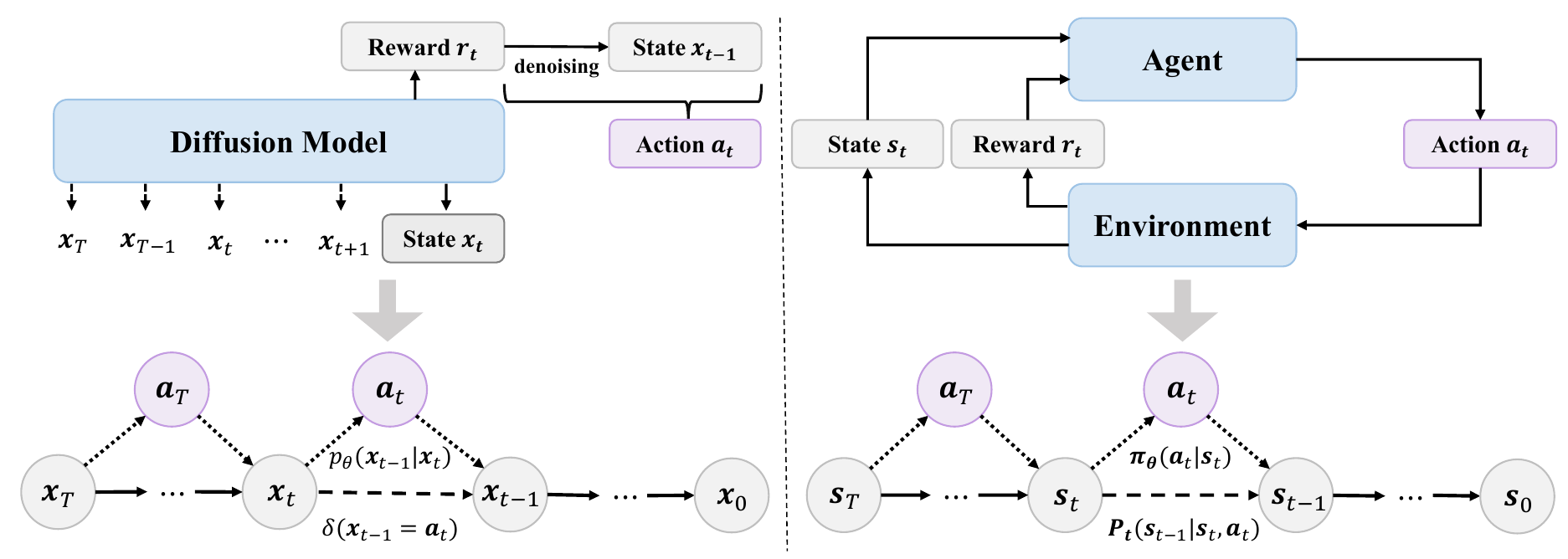}
\vspace{-2em}
\caption{
A schematic comparison of the diffusion model denoising process and the RL problem formulated as an MDP.
}
\vspace{-1.6em}
\label{fig:compare_dm_rl}
\end{figure}

\subsubsection{Preference-Based RL Optimization}
\label{sec:prefer_rl}

Preference-based RL effectively aligns DMs with user intent when struggling to define explicit rewards, using pairwise or ranking feedback as a proxy reward for policy updates.

Specifically, RLHF~\cite{christiano2017RLHF} trains a reward model $r_\phi$ from preference triples $(c, y_A, y_B)$ via a Bradley-Terry likelihood, producing a dense proxy for human feedback. The entire process typically involves supervised pretraining, reward model training, and preference-alignment optimization:
\begin{equation}
\begin{aligned}
\mathcal{L}_{RM}=-\sum_{(c,y_A,y_B)} \Big[  \, y &\log \sigma \big(r_\phi (c,y_A)-r_\phi(c,y_B)\big) \\
 +   (1-y) &\log \sigma \big(r_\phi (c,y_B)-r_\phi(c,y_A)\big) \Big].
\end{aligned}
\end{equation}

DPO~\cite{rafailov2023direct} aligns models directly with pairwise preference data without an explicit reward model. Given a prompt $c$ and pairwise preference $(y_A, y_B)$, it optimizes
\begin{equation}
\begin{aligned}
\mathcal{L}_{\mathrm{DPO}} &= -\mathbb{E}_{c,y_A,y_B}\big[ \log \sigma (\beta \Delta_\theta(c,y_A,y_B))\big], \\
\Delta_\theta(c,y_A,y_B) &= \log \frac{\pi_\theta(y_A|c)}{\pi_{\theta_{\rm ref}}(y_A|c)} 
                             - \log \frac{\pi_\theta(y_B|c)}{\pi_{\theta_{\rm ref}}(y_B|c)}.
\end{aligned}
\end{equation}
For image-generating diffusion models, this encourages
\begin{equation}
\pi_\theta(y^+|c) > \pi_\theta(y^-|c)
\end{equation}
while keeping $\pi_\theta$ close to the reference model $\pi_{\rm ref}$.

\subsubsection{Policy-Based RL Optimization}
\label{sec:policy_rl}

While RLHF~\cite{christiano2017RLHF} has been widely adopted for human alignment, finetuning DMs is more naturally cast as an RLVR problem by regarding the denoising trajectory as a Markov Decision Process (MDP)~\cite{haarnoja2017reinforcement,neu2017unified,geist2019theory}.
As depicted in Fig.~\ref{fig:compare_dm_rl}, each denoising step corresponds to an action, and the entire generative trajectory forms a decision-making sequence.

Formally, an MDP is defined as $\{\mathcal{S},\mathcal{A},\{P_t\},\{r_t\},p_0\}$, where $\mathcal{S}$ denotes the state space, $\mathcal{A}$ the action space, $P_t$ the transition dynamics, $r_t$ the reward function, and $p_0$ the initial state distribution. A policy $\pi_t:\mathcal{S}\rightarrow\mathcal{A}$ maps states to actions, and RL maximizes the expected cumulative reward:
\begin{equation}
    \mathbb{E}_{\pi} \left[ \sum_{t=0}^T r_t(s_t,a_t)\right].
\end{equation}

When casting DMs into this RL framework, both states and actions lie in the model input space $\mathcal{X}$. The transition dynamics are deterministic, following the pre-defined denoising schedule. Importantly, under this RLVR paradigm, rewards are assigned solely at the terminal step and are derived from verifiable criteria (\textit{e.g.}, rule-based checks or automated evaluators), thereby ensuring objective and stable optimization signals. The policy $\pi_t$ corresponds to the denoising distribution $p(\mathbf{x}_{T-t}|\mathbf{x}_{T+1-t})$, while the initial state distribution $p_0$ matches the Gaussian prior $p(\mathbf{x}_T)$.

This formulation tightly aligns RL optimization with the stepwise structure of diffusion-based generation, enabling policy improvement via trajectory-level, verifiable feedback. In practice, policy-based algorithms such as PPO~\cite{schulman2017proximal} and GRPO~\cite{shao2024deepseekmath} can be adopted to optimize the denoising policy under the RLVR objective.
\begin{figure}[t]
\centering
\includegraphics[width=1\linewidth]{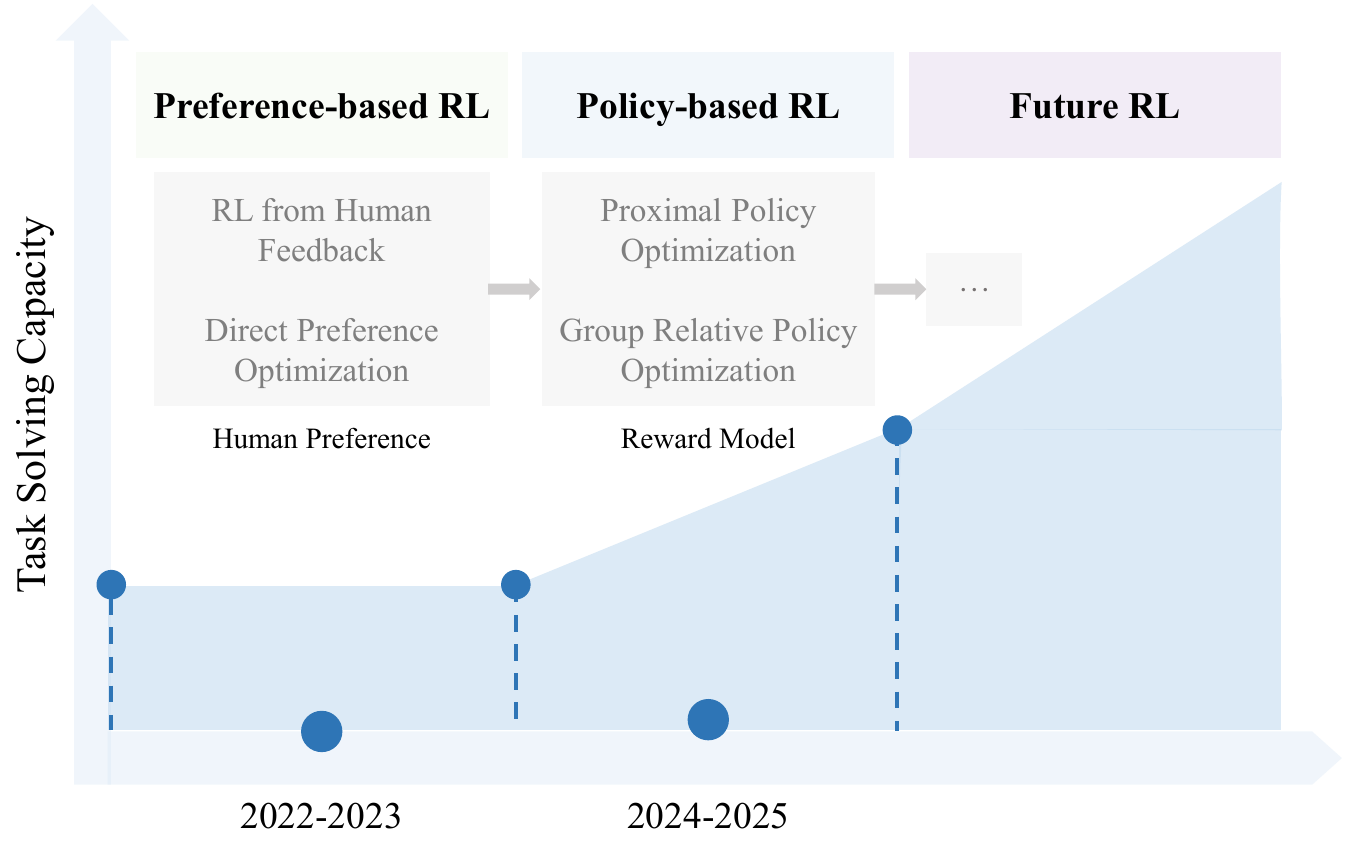}
\vspace{-2em}
\caption{
RL for DMs includes preference-/policy-based optimization.
}
\label{fig:value_preference_rl}
\end{figure}
Specifically, PPO~\cite{schulman2017proximal} is a policy-gradient approach that effectively balances training stability and simplicity by constraining update magnitudes through a clipped surrogate objective. For the probability ratio
\begin{equation}
    r_t(\theta)=\frac{\pi_\theta(a_t \mid s_t)}{\pi_{\theta_{\text{old}}}(a_t \mid s_t)},
\end{equation}
PPO~\cite{schulman2017proximal} replaces the standard objective $\mathbb{E}[r_t(\theta)\hat{A}_t]$ with a clipped version, which can be expressed as
\begin{equation}
    \mathcal{L}^{\text{CLIP}}(\theta)=\mathbb{E}\big[\min(r_t(\theta)\hat{A}_t,\, \text{clip}(r_t(\theta),1-\epsilon,1+\epsilon)\hat{A}_t)\big],
\end{equation}
where the advantage $\hat{A}_t$ is typically estimated via Generalized Advantage Estimation (GAE)~\citep{schulman2015high} using
\begin{equation}
    \delta_t = r_t + \gamma V(s_{t+1}) - V(s_t),\quad
    \hat{A}_t = \delta_t + \gamma \lambda \, \hat{A}_{t+1},
\end{equation}
which provides a tunable bias–variance trade-off. Notably, the corresponding return is $\hat{G}_t = \hat{A}_t + V(s_t)$.

Moreover, GRPO~\cite{shao2024deepseekmath} is another policy-gradient method originally developed for LLM finetuning and can be naturally extended to diffusion RL. Unlike standard actor-critic methods, GRPO~\cite{shao2024deepseekmath} removes the separate value network and computes \textit{group-relative advantages} across multiple outputs for the same condition $c$. For $G$ outputs $\{o_1,\dots,o_G\}$ with corresponding rewards $\{r_1,\dots,r_G\}$, we have
\begin{equation}
    \mu = \frac{1}{G}\sum_{i=1}^{G} r_i,\quad
    \sigma = \sqrt{\frac{1}{G}\sum_{i=1}^{G} (r_i-\mu)^2},\quad
    \hat A_i = \frac{r_i - \mu}{\sigma}.
\end{equation}
Similar to PPO~\cite{schulman2017proximal}, GRPO~\cite{shao2024deepseekmath} maximizes a clipped surrogate objective with optional KL regularization to a reference policy $\pi_{\rm ref}$.
The adjusted objective is as follows: 
\begin{equation}
\resizebox{\linewidth}{!}{$
\begin{aligned}
\mathcal{J}_{\rm GRPO}(\theta) &= 
\mathbb{E}_{c, \{o_i\} \sim \pi_{\theta_{\rm old}}} 
\Bigg[ 
\frac{1}{G} \sum_{i=1}^{G} 
\frac{1}{|o_i|} \sum_{t=1}^{|o_i|} 
\min\Bigg(
\frac{\pi_\theta(o_{i,t}\mid c, o_{i,<t})}
{\pi_{\theta_{\rm old}}(o_{i,t}\mid c, o_{i,<t})} \hat A_i, \\
&\quad \text{clip}\Big(
\frac{\pi_\theta(o_{i,t}\mid c, o_{i,<t})}
{\pi_{\theta_{\rm old}}(o_{i,t}\mid c, o_{i,<t})}, 1-\varepsilon, 1+\varepsilon
\Big) \hat A_i
\Bigg)
\Bigg] \\
&\quad - \beta D_{\rm KL}(\pi_\theta \| \pi_{\rm ref}).
\end{aligned}
$}
\end{equation}
By using group-relative advantages, GRPO~\cite{shao2024deepseekmath} provides a lightweight, stable alternative to PPO~\cite{schulman2017proximal} for generative RL.
Fig.~\ref{fig:value_preference_rl} visually presents the development of RL for DMs.

\section{RL for Diffusion Large Language Models}
\label{sec:dLLM}

\subsection{Preliminaries of Diffusion-based LLMs}
\label{sec:dLLMs}
Diffusion-based LLMs (dLLMs) have emerged as a promising \textit{\textbf{non-autoregressive}} alternative to traditional autoregressive (AR) transformers for text generation~\cite{nie2025large,Li-2022-DiffusionLM,gong2022diffuseq,gong2023diffuseqv2,strudel2022self}.
They treat text generation as an iterative unmasking trajectory, thereby enhancing controllability (\textit{e.g.}, output length) and enabling non-sequential intermediate steps~\cite{huang2025reinforcingdiffusionchainlateral}.
Here we outline the generation process of dLLMs as follows:

\noindent \textbf{Inverse Unmasking Process.} Unlike AR methods, dLLMs generate all tokens in parallel via iterative unmasking~\cite{li2023diffusion,zhao2025d1scalingreasoningdiffusion}.
Specifically, analogous to diffusion denoising, the unmasking process takes a completely masked state $\mathbf{x}_T$ as input to generate text $\mathbf{x}_0$.
At timestep $t$, the partially unmasked sequence $\mathbf{x}_t=(x^1_t,x^2_t,...,x^n_t)$ is refined by filling masked tokens.
Let $p_\theta(x^i_t|\hat{\mathbf{x}}_{t+1},c,t)$ denote the model’s predicted probability for the $i$-th token at timestep $t$, where $\hat{\mathbf{x}}_{t+1}$ represents the latent sequence variable predicted at timestep $t+1$.
The joint probability is given by $p_\theta(\mathbf{x}_0|c)=\prod_{t=0}^{T-1}\prod_{i=1}^np_\theta(x^i_t|\hat{\mathbf{x}}_{t+1},c,t)$ \cite{li2023diffusion}.


\noindent \textbf{Forward Masking Process.} 
Moreover, the masking process corrupts tokens at timestep $t \in [0,T]$ according to a probability distribution function.
Let a noise schedule $\alpha_t\in [0,1]$ denote the cumulative probability that a token remains unmasked by timestep $t$.
As positions are corrupted independently, we have the conditional distribution $q_{t|0}(\mathbf{x}_t | \mathbf{x}_0)=\prod_{i=0}^{L} q_{t|0}(x_t^i|x_0^i)$ where
\begin{equation}
    q_{t|0}(x_t^i|x_0^i) = \begin{cases}
                    1 - \alpha_t, & x_t^i = \texttt{[mask]}; \\
                    \alpha_t, & x_t^i = x_0^i.
                 \end{cases}
\end{equation}
We want to unmask an intermediate state $s$ from a corrupted state $t$ ($s<t$). The conditional distribution of a partially corrupted sequence $\mathbf{x}_s$ (compared to $\mathbf{x}_t$) is derived via case analysis and Bayes' rule~\cite{zhao2025d1scalingreasoningdiffusion,shi2024simplified,nie2025large}:
\begin{equation}
\begin{medsize}
    q_{s|t}(x^i_s|\mathbf{x}_t) =\begin{cases}
                    1, & x_t^i \neq \texttt{[mask]},\ x_s^i = x_t^i; \\
                    \frac{1 - \alpha_s}{1 - \alpha_t}, & x_t^i = \texttt{[mask]},\ x_s^i = \texttt{[mask]}; \\
                    \frac{\alpha_s - \alpha_t}{1 - \alpha_t}q_{0|t}(x_s^i|x_t), & x_t^i = \texttt{[mask]},\ x_s^i \neq \texttt{[mask]}; \\
                    0, & \text{otherwise.} \\
                 \end{cases}
\end{medsize}
\end{equation}

\noindent \textbf{ELBO Objective.}
dLLMs are trained by maximizing the evidence lower bound (ELBO), aligning the reverse unmasking process $p_\theta$ with the forward masking process $q$.
This can be written as minimizing the KL divergence between $p_\theta(x^i_{t}|\mathbf{x}_{t+1}, c)$ and $q_{t|t+1}(x^i_t|\mathbf{x}_{t+1}, x^i_0)$.
Assuming absorbing transition~\cite{austin2021structured,campbell2022continuous,tang2025wd1weightedpolicyoptimization}, the token-level ELBO optimization objective for a given data example~\cite{zhan2025principledtractablerlreasoning} is
\begin{equation} \label{eq:elbo} 
\mathcal L(\theta)
=
\mathbb{E}_{\substack{
t\sim U[0,1]\\
\mathbf{x}_0\sim p_{data}\\
\mathbf{x}_t\sim q_{t|0}(\mathbf{x}_t\mid\mathbf{x}_0)
}}
\left[
\frac{1}{t}
\sum  _{x_t^i=\texttt{[mask]}}
\log p_\theta(x_0^i\mid \mathbf{x}_t)
\right].
\end{equation}
which intuitively trains the model to predict the original token $x^i_0$ for every masked position at a random timestep $t$.
\begin{table*}[t]
\setlength{\tabcolsep}{1pt}
\centering
\caption{Summary of reinforcement learning methods for diffusion-based LLMs as of 2026.08.}
\vspace{-1em}
\label{tab:dllm-rl-methods}
\scalebox{0.734}{\begin{tabular}{ccccccc}
\toprule
\textbf{Methods} & \textbf{Date} &  \textbf{Backbone} & \textbf{Policy} & \textbf{Key Contributions} & \textbf{Primary Task} & \textbf{Links} \\

\midrule
\multicolumn{7}{c}{\cellcolor{lightblue}\textbf{\textit{GRPO-Related Post Training \& Likelihood Estimation}}}\\
\midrule

d1~\cite{zhao2025d1scalingreasoningdiffusion} & 2025.04 & LLaDA & On & First policy-gradient RL for dLLMs; masked-prompt likelihood approx & Math/Logic Reasoning & \Paper{https://arxiv.org/abs/2504.12216} \Code{https://github.com/dllm-reasoning/d1} \\
DiffuCoder~\cite{gong2025diffucoderunderstandingimprovingmasked} & 2025.06 & Qwen-2.5-Coder & On & Complementary masks for low-variance token likelihood & Code & \Paper{https://arxiv.org/abs/2506.20639} \Code{https://github.com/apple/ml-diffucoder}\,\href{https://huggingface.co/apple/DiffuCoder-7B-Base}{\faHuggingFace}\\
wd1~\cite{tang2025wd1weightedpolicyoptimization} & 2025.07 & LLaDA & On & Weighted-likelihood objective; denominator-free update; much faster & Math Reasoning & \Paper{https://arxiv.org/abs/2507.08838} \Code{https://github.com/xiaohangt/wd1} \\
BGPO~\cite{lin2025boundaryguidedpolicyoptimizationmemoryefficient} & 2025.10 & LLaDA & On & Boundary-guided lower bound enabling large MC with constant memory & Math/Code/Planning & \Paper{https://arxiv.org/abs/2510.11683} \Code{https://github.com/THU-KEG/BGPO}\,\href{https://huggingface.co/collections/THU-KEG/llada-8b-bgpo}{\faHuggingFace} \\
DMPO~\cite{zhu2025enhancingreasoningdiffusionllms} & 2025.10 & LLaDA & Off & Forward-KL via weighted denoising cross-entropy & Math Reasoning & \Paper{https://arxiv.org/abs/2510.08233} \Code{https://github.com/yuchen-zhu-zyc/DMPO} \\
GDPO~\cite{rojas2025improvingreasoningdiffusionlanguage} & 2025.10 & LLaDA & On & Semi-deterministic Monte Carlo for ELBO & Math Reasoning & \Paper{https://arxiv.org/pdf/2510.08554} \Code{https://github.com/kevinrojas1499/GDPO}\\
AGRPO~\cite{zhan2025principledtractablerlreasoning} & 2025.10 & LLaDA & On & Unbiased MC per-timestep gradient without ad-hoc ratios & Math Reasoning & \Paper{https://arxiv.org/abs/2510.04019} \Code{https://github.com/probablyabot/agrpo} \\
ESPO ~\cite{ou2025principledrldiffusionllms} & 2025.12 & LLaDA, Dream & On & Entire sequence generation as a single action; ELBO as sequence-likelihood& Math/Code/Planning & \Paper{https://arxiv.org/pdf/2512.03759} \Code{https://github.com/ML-GSAI/ESPO}  \href{https://huggingface.co/collections/GSAI-ML/espo}{\faHuggingFace}\\
d-TreeRPO~\cite{pan2026dtreerporeliablepolicyoptimization} & 2025.12 & LLaDA(-MoE) & On & Tree rollouts for step-wise credit; time-scheduled self-distillation loss & Math/Planning & \Paper{https://arxiv.org/abs/2512.09675v3}\,\Code{https://github.com/THU-BPM/d-TreeRPO} \\
LFPO~\cite{wei2026lfpolikelihoodfreepolicyoptimization}
& 2026.03
& \footnotesize{\smash{\makecell[c]{LLaDA, Dream,\\[-2pt]DiffuCoder}}}
& Off
& Flow-matching in token space without likelihood estimation
& Math/Code
& \Paper{https://arxiv.org/abs/2603.01563v2}\,\Code{https://github.com/kithib/LFPO}
\\
StableDRL~\cite{zhong2026stabilizingreinforcementlearningdiffusion} & 2026.03 & LLaDA, SDAR & On & Unconditional clipping + self-normalization stabilize RL updates & Math/Planning & \Paper{https://arxiv.org/abs/2603.06743v1}\,\Code{https://github.com/JianyuanZhong/StableDRL} \\
RSPO~\cite{yu2026relativescorepolicyoptimization} & 2026.05 & LLaDA & On & Reward-implied target ratios calibrate noisy likelihood estimates & Math/Planning & \Paper{https://arxiv.org/abs/2605.10218v1} \\
GDSD~\cite{tang2026gdsdreinforcementlearningguided} & 2026.05 & LLaDA, Dream & Off & Advantage-guided denoiser self-distillation without ELBO & Math/Code/Planning & \Paper{https://arxiv.org/abs/2605.29398v1}\,\Code{https://github.com/GaryBall/GDSD}\,\href{https://huggingface.co/diffusion-reasoning/gdsd_code_llada}{\faHuggingFace} \\
\midrule
d2~\cite{wang2025d2improvedtechniquestraining} & 2025.09 & LLaDA & On & AnyOrder unbiased estimator; StepMerge to merge diffusion steps & Reasoning; Infer Efficiency & \Paper{https://arxiv.org/abs/2509.21474} \Code{https://github.com/kuleshov-group/d2}\\
IGPO~\cite{zhao2025inpaintingguidedpolicyoptimizationdiffusion} & 2025.09 & LLaDA & On & Inpainting hints inject partial CoT during online sampling & Math Reasoning & \Paper{https://arxiv.org/abs/2509.10396} \Code{https://github.com/facebookresearch/igpo} \\
DiFFPO~\cite{zhao2025diffpotrainingdiffusionllms} & 2025.10 & LLaDA & Off & Off-policy surrogate + importance sampling; RL-trained inference budget & Math Reasoning; Sampler & \Paper{https://arxiv.org/abs/2510.02212} \\
dUltra~\cite{chen2026dultraultrafastdiffusionlanguage} & 2025.12 & LLaDA & On & Learned unmasking planner; rewards for accuracy and fewer steps & Math/Code & \Paper{https://arxiv.org/abs/2512.21446v2}\,\Code{https://github.com/chinsengi/dUltra-os}\,\href{https://huggingface.co/sengi/dUltra-coding}{\faHuggingFace} \\
STP~\cite{liu2026efficientstablereinforcementlearning} & 2026.02 & LLaDA & On & Prune token updates and late denoising steps for faster, stabler RL & Reasoning & \Paper{https://arxiv.org/abs/2602.08905v1}\,\Code{https://github.com/Lolo1222/STP} \\
ERILS~\cite{lee2026onpolicyexplorationintegratingexternal} & 2026.08 & LLaDA & Both & External policy rollouts expand exploration; separate reward rules & Math/Planning & \Paper{https://arxiv.org/abs/2608.01717v2} \\
MAPG~\cite{raajesh2026maskawarepolicygradientsdiffusion} & 2026.07 & LLaDA, Dream & On & dLLM generation as a MDP; joint policy gradients & Math/Code/Planning & \Paper{https://arxiv.org/abs/2607.15200v1}\,\Code{https://github.com/Haran71/mask-aware-policy-gradients} \\

\midrule
\multicolumn{7}{c}{\cellcolor{lightblue}\textbf{\textit{Human Preference Alignment}}}\\
\midrule
LLaDA 1.5~\cite{zhu2025llada15variancereducedpreference} & 2025.05 & LLaDA & Off & Variance-reduced preference optimization; Antithetic sampling & Alignment + Reasoning & \Paper{https://arxiv.org/abs/2505.19223} \Code{https://github.com/ML-GSAI/LLaDA-1.5}\,\href{https://huggingface.co/GSAI-ML/LLaDA-1.5}{\faHuggingFace} \\
ELBO-KTO~\cite{jindal2025aligningdiffusionlanguagemodels} & 2025.10 & LLaDA & Off & ELBO-based preference alignment using unpaired like/dislike feedback & Alignment + Reasoning & \Paper{https://arxiv.org/abs/2510.23658v2}\,\Code{https://github.com/vaibhavjindal/elbo-kto} \\
\midrule
\multicolumn{7}{c}{\cellcolor{lightblue}\textbf{\textit{Boosting Reasoning Abilities}}}\\
\midrule
DCoLT~\cite{huang2025reinforcingdiffusionchainlateral} & 2025.05 & SEDD, LLaDA & On & Optimize whole latent reasoning chain (lateral thought) & Math/Code & \Paper{https://arxiv.org/abs/2505.10446} \Code{https://github.com/maple-research-lab/LLaDOU}\,\href{https://huggingface.co/maple-research-lab/LLaDOU-v0-Math}{\faHuggingFace} \\
MDPO~\cite{he2025mdpoovercomingtraininginferencedivide} & 2025.08 & LLaDA & On & Progressive-schedule RL with intermediate rewards; RCR remasking & Bridging schedule mismatch & \Paper{https://arxiv.org/abs/2508.13148} \Code{https://github.com/autonomousvision/mdpo} \\
SAPO~\cite{xie2025stepawarepolicyoptimizationreasoning} & 2025.10 & LLaDA & On & Step-aware process reward aligning denoising hierarchy & Complex Reasoning & \Paper{https://arxiv.org/abs/2510.01544} \\
\midrule
\multicolumn{7}{c}{\cellcolor{lightblue}\textbf{\textit{Frameworks \& Pipeline improvements}}}\\
\midrule
TraceRL~\cite{wang2025revolutionizingreinforcementlearningframework} & 2025.09 & Dream, SDAR & On & Trajectory-aware RL; Blocked diffusion \& attention & Math/Code; Framework & \Paper{https://arxiv.org/abs/2509.06949} \Code{https://github.com/Gen-Verse/dLLM-RL}\,\href{https://huggingface.co/collections/Gen-Verse/trado-series}{\faHuggingFace} \\
RFG~\cite{chen2025rfgtesttimescalingdiffusion} & 2025.09 & LLaDA, Dream & \NA & Log-likelihood ratio guidance of enhanced \& ref dLLM & Math/Code at inference time & \Paper{https://arxiv.org/abs/2509.25604} \\
SPG~\cite{wang2025spgsandwichedpolicygradient} & 2025.10 & LLaDA & On & ELBO/EUBO sandwich + block-wise masking & Math/Code Reasoning & \Paper{https://arxiv.org/abs/2510.09541} \Code{https://github.com/facebookresearch/SPG} \\
DiRL \cite{zhu2025dirl}& 2025.12 & SDAR & On & Unbiased logits of blockwise dLLM; FlexAttention & Math & \Paper{https://arxiv.org/abs/2512.22234} \Code{https://github.com/OpenMOSS/DiRL}\\
DARE~\cite{yang2026darediffusionlargelanguage} & 2026.04 & LLaDA,Dream,SDAR & Both & Unified RL training and evaluation across masked and blockwise dLLMs & Math/Code; Framework & \Paper{https://arxiv.org/abs/2604.04215v1}\,\Code{https://github.com/yjyddq/DARE} \\

\bottomrule
\end{tabular}}
\vspace{-1.25em}
\end{table*}

\noindent\textbf{Overview of Training and Inference.}
Using LLaDA~\cite{nie2025large} as an example, it involves three stages: \textbf{\textit{1) Pre-training:}} training on text with randomly masked tokens, where the masking ratio is determined by $t\sim U[0,1]$. \textbf{\textit{2) SFT:}} similar to \textit{\textbf{1)}} but merely randomly masking response tokens in structured ``instruction-response'' pairs. \textit{\textbf{3) Sampling:}} simulating a diffusion process from $t=1$ (masked) to $t=0$ (unmasked).

\subsection{The Role of RL in Diffusion-based LLMs}
\label{sec: RL's role in dLLMs}

While RL has proven effective for fine-tuning AR LLMs~\cite{bai2022traininghelpfulharmlessassistant,zhang2025survey}, its integration requires rethinking credit assignment due to \textit{\textbf{non-AR}} generation via iterative unmasking, especially \emph{how to estimate logits, policy, and loss}.
We originally categorize existing RL4dLLMs along two orthogonal axes that directly address this adaptation: \textbf{\textit{trajectory level}} (defining the granularity of credit assignment) and \textbf{\textit{reward density}} (specifying the distribution of supervisory signals).

\textbf{Trajectory Level} characterizes the granularity at which policy updates assign credit to likelihood terms.

For a dLLM policy inducing a trajectory $\tau=(s_{0:T},a_{0:T})$ over a discrete reverse-diffusion process, the loss function takes the form:
\[
\mathcal{L}(\pi_\theta)\sim
\begin{cases}
\sum_{t} \mathbb{E}\big[\ell_\theta(a_t \mid s_t)\big], & \text{token-level};\\[2pt]
\sum_{k} \mathbb{E}\big[\ell_\theta(a_{u_k:v_k} \mid s_{u_k:v_k})\big], & \text{phrase-level};\\[2pt]
\mathbb{E}\big[\ell_\theta(a_{0:T} \mid s_{0:T})\big], & \text{sequence-level},
\end{cases}
\]
where $\{(u_k,v_k)\}$ are masked spans and $\ell_\theta$ denotes a surrogate loss (\textit{e.g.}, ELBO-based objective, GRPO-based objective, likelihood ratio, or policy gradient).
Specifically, \textbf{\textit{token-level}} methods apply GRPO-style policy gradients to token-wise log-probabilities, which depend on individual token probability.
\textit{\textbf{Phrase-level}} methods treat contiguous masked spans (\textit{e.g.}, reasoning clauses, code snippets) as atomic units.
Even when final updates are sequence-level, many studies focus on phrase granularity via explicit inpainting of partial traces or masked phrases.
\textbf{\textit{Sequence-level}} methods, by contrast, optimize the full reverse trajectory, with advantages typically derived from holistic outcome rewards (\textit{e.g.}, verifier scores).


\textbf{Reward Density} characterizes the distribution of supervisory signals across the trajectory via
\[
R(\tau)\sim
\begin{cases}
r_T, & \text{Single Maximum (Outcome RM)};\\
r_T^{+}-r_T^{-}, & \text{Double-polar};\\
\frac{1}{T}\sum_{t=1}^{T} w_tr_t, & \text{Mean Reward};\\
\{r_{t,k}\}, & \text{Dense (Process RM)},
\end{cases}
\]
where $r_T$ is an Outcome Reward Model (ORM) computed only from the final output and $\{r_t\}$ is stepwise Process Reward Models (PRMs) that provide dense feedback.
Specifically, \textit{\textbf{Single Maximum}} RL maximizes the terminal reward $r_T$, optimizing the advantage $A_i$ via GRPO~\cite{shao2024deepseekmath}, i.e., $\min \left(\bar{\rho_i}^t A_i, \operatorname{clip}\left(\bar{\rho_i}^t, 1-\varepsilon, 1+\varepsilon\right) A_i\right)$.
\textbf{\textit{Double-polar}} methods use a bipartite objective to steer sequences toward preferred outcomes.
\textit{\textbf{Mean reward}} models aggregate stepwise weighted rewards.
\textit{\textbf{Dense reward}} models provide intermediate supervision $r_{t,k}$ across steps $t$ and sample groups $N$.
\begin{figure}[t]
    \centering
    \includegraphics[width=\linewidth]{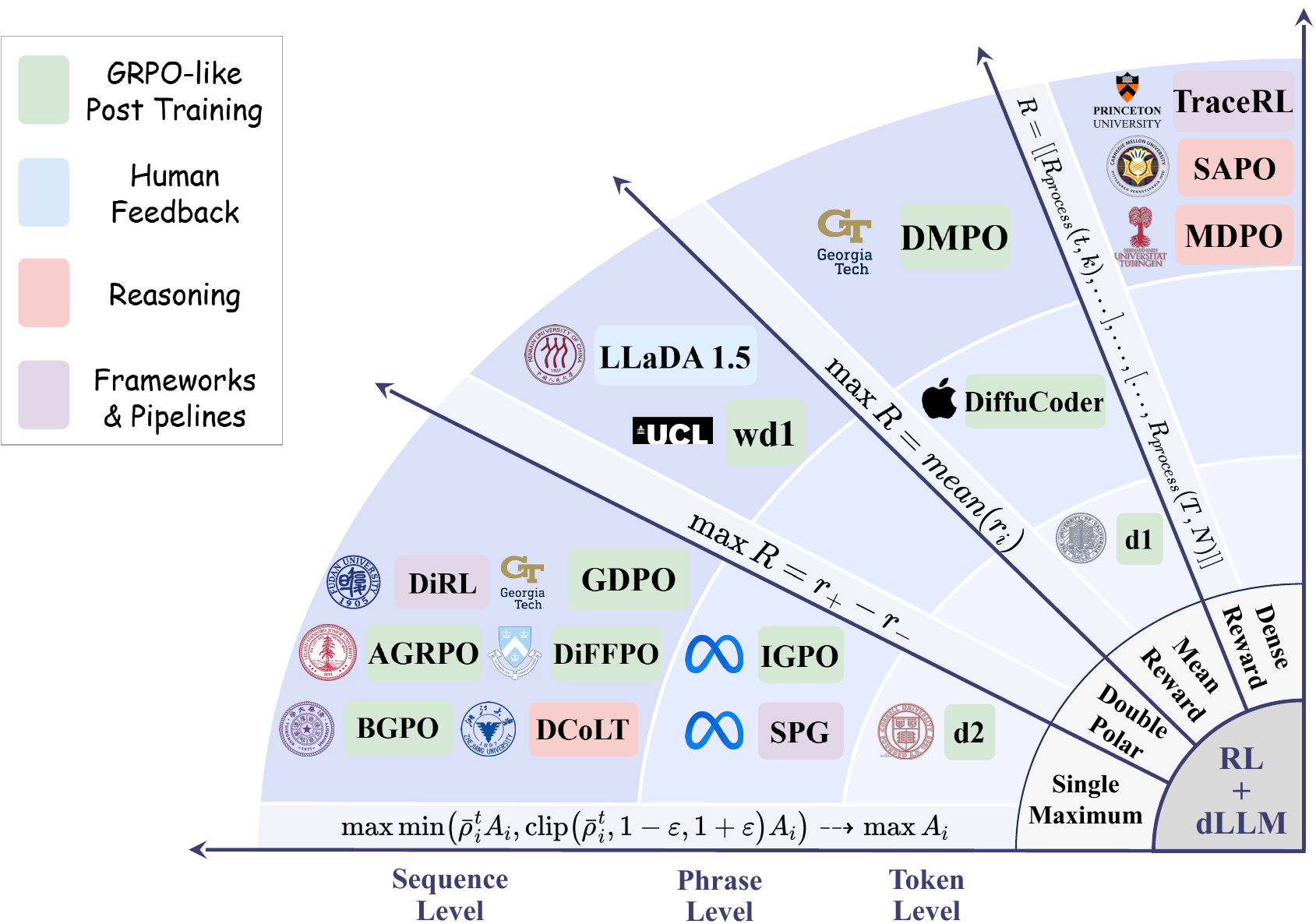}
    \vspace{-2em}
    \caption{An overview of representative works on RL4dLLMs.}
    \label{fig:dLLM-overview}
    \vspace{-1em}
\end{figure}

Fig.~\ref{fig:dLLM-overview} visualizes this taxonomy as a 2D coordinate chart, where representative RL methods are organized by trajectory level and reward density. Complementarily, Table \ref{tab:dllm-rl-methods} summarizes methods and Table \ref{tab:dllm-bench} compares benchmark performance.
Below, we review these developments by technical insights and target use cases.

\subsubsection{GRPO-Related Post Training}
\label{sec:GRPO-Related Post Training}
Efficient policy gradient estimation for dLLMs requires accurate computation of \textbf{\textit{per-token}} and \textbf{\textit{sequence}} completion probability, which is a computationally intensive challenge.
Recent methods have extended GRPO~\cite{shao2024deepseekmath} to address this, with key advances summarized below.

\noindent \textbf{Variance-Reduced Likelihood Estimation.}
d1~\cite{zhao2025d1scalingreasoningdiffusion}, the first application of policy gradient with masked prompts, randomly masks portions of the input prompt $q$ (yielding $q'$) while keeping the completion fully masked.
Each training epoch runs $n$ iterations of this process, unmasking only one timestep per iteration to estimate per-token log probability $\phi^{\pi_\theta}(o^k|q')$ and then optimize GRPO’s~\cite{shao2024deepseekmath} policy gradient objective.
Despite effective fine-tuning for math and logical reasoning, estimation error may induce high variance.

To effectively reduce variance, DiffuCoder~\cite{gong2025diffucoderunderstandingimprovingmasked} samples paired completions for each batch with a complementary masking pattern (i.e., high-confidence tokens in one completion are masked in the other) and then averages log-likelihoods over the two trajectories to cancel masking-induced noise.
Apart from this, wd1~\cite{tang2025wd1weightedpolicyoptimization} reformulates RL into a double-polar objective, where completions with higher advantage get larger weight. It requires merely current-model per-token probabilities $\pi_{\theta}$ instead of old or reference policy probabilities ($\pi_{\theta_\text{old}}/\pi_{\text{ref}}$) in importance-weighted gradients.
In short, this mechanism avoids high-variance importance sampling, reduces denominator bias, and allows direct RL fine-tuning from scratch.


Other methods further refine estimation: BGPO~\cite{lin2025boundaryguidedpolicyoptimizationmemoryefficient} constructs a tight ELBO for larger Monte Carlo (MC) samples and more accurate likelihood approximations;
DMPO~\cite{zhu2025enhancingreasoningdiffusionllms} aligns the dLLM policy with the reward-tilted distribution via cross-entropy optimization;
GDPO~\cite{rojas2025improvingreasoningdiffusionlanguage} employs Semi-deterministic MC to curb ELBO variance;
AGRPO~\cite{zhan2025principledtractablerlreasoning} treats token summation as an expectation over the diffusion timesteps $t$ and derives a policy gradient estimator via MC sampling to avoid heuristic likelihood ratios; SEPO~\cite{zekri2025finetuningdiscretediffusionmodels} uses importance sampling through gradient flow; and d-TreeRPO \cite{pan2026dtreerporeliablepolicyoptimization} structure rollouts as a tree to compute advantage and propose a time-scheduled self-distillation objective.

\noindent\textbf{Speed-Performance Trade-off Optimization.}
DiFFPO~\cite{zhao2025diffpotrainingdiffusionllms} targets efficiency through \textbf{\textit{1)}} an off-policy RL paradigm with a surrogate policy (using two-stage likelihood approximation with importance-sampling correction) and \textit{\textbf{2)}} an RL-trained controller that dynamically sets budget thresholds per prompt, thereby reducing denoising steps.
d2~\cite{wang2025d2improvedtechniquestraining} introduces d2-StepMerge (merging diffusion steps for computational savings) and d2-AnyOrder (an attention-inspired mechanism requiring any-order causality during decoding). More recently, dUltra \cite{chen2026dultraultrafastdiffusionlanguage} introduces an unmasking planner head for superior accuracy-efficiency trade-offs, STP \cite{liu2026efficientstablereinforcementlearning} compresses the redundancy through pruning the sequence length for
ELBO, as well as refinement steps.



\noindent \textbf{Knowledge-Enhanced Sampling.}
IGPO~\cite{zhao2025inpaintingguidedpolicyoptimizationdiffusion} injects partial ground-truth reasoning traces into the unmasking trajectory as hints during sampling.
This avoids wasted samples on non-rewarding intermediate sequences and significantly increases the probability of high-reward completions, thereby boosting training efficiency. Moreover, ERILS \cite{lee2026onpolicyexplorationintegratingexternal} addresses weak on-policy exploration by incorporating higher-reward rollouts from a stronger external policy.
\begin{table}[t]
\centering
\setlength{\tabcolsep}{2pt}
\caption{Quantitative comparison of benchmark results for RL+dLLM methods. The values are rounded to three decimal places. For tasks with multiple sequence length evaluations, the best reported result is selected. Notably, $^\dagger$ denotes third-party results. Other results are obtained from the original paper, where the best result across all reported evaluation protocols is chosen. The best and the second-best results of dLLMs in each column are demonstrated in \colorbox{green!16}{\textbf{green}} and \colorbox{gray!20}{\underline{gray}}.}
\label{tab:dllm-bench}
\vspace{-1em}
\scalebox{0.666}{\begin{tabular}{lcccccc}
\toprule
\textbf{Methods} & \textbf{GSM8K} & \textbf{MATH500} & \textbf{Countdown} & \textbf{Sudoku} & \textbf{HumanEval} & \textbf{MBPP} \\
\midrule
\multicolumn{7}{c}{\cellcolor{lightblue}\textbf{\textit{Non-RL diffusion-based LLMs}}}\\
\midrule
LLaDA-8B-Instruct~\cite{nie2025large} & 78.2$^\dagger$ & 36.2$^\dagger$ & 20.7$^\dagger$ & 11.7$^\dagger$ & 49.4 & 41.0\\
\midrule
\multicolumn{7}{c}{\cellcolor{lightblue}\textbf{\textit{RL-driven diffusion-based LLMs}}}\\
\midrule
d1~\cite{zhao2025d1scalingreasoningdiffusion} & 82.1 & 40.2 & 42.2 & 22.1 & 39.0 & 45.5 \\
DiffuCoder~\cite{gong2025diffucoderunderstandingimprovingmasked} & \NA & \NA & \NA & \NA & \cellcolor{gray!20}\underline{73.2} & \cellcolor{gray!20}\underline{78.6} \\
wd1~\cite{tang2025wd1weightedpolicyoptimization} & 82.3 & 39.0 & 51.2 & {76.4} & 39.0 & 39.2 \\
BGPO~\cite{lin2025boundaryguidedpolicyoptimizationmemoryefficient} & 84.3 & {45.7} & \cellcolor{green!16}\textbf{87.5} & 26.9 & 47.6 & 41.7 \\
DMPO~\cite{zhu2025enhancingreasoningdiffusionllms} & 85.2 & 42.8 & {79.7} & {32.8} & \NA & \NA \\
GDPO~\cite{rojas2025improvingreasoningdiffusionlanguage} & 82.3 & 38.2 & 67.2 & 25.1 & 39.6 & {50.6} \\
AGRPO~\cite{zhan2025principledtractablerlreasoning} & {88.1} & 42.0 & {82.4} & \cellcolor{gray!20}\underline{95.2} & \NA & \NA \\
ESPO~\cite{ou2025principledrldiffusionllms} & 83.7 & 43.4 & 82.0 & 92.7 & 50.0 & 47.4\\
d2 ~\cite{wang2025d2improvedtechniquestraining} & 85.0 & 41.6 & 56.6 &{91.9} & \NA & \NA \\
IGPO~\cite{zhao2025inpaintingguidedpolicyoptimizationdiffusion} & {86.4} & {47.4} & \NA & \NA & \NA & \NA \\
DiFFPO~\cite{zhao2025diffpotrainingdiffusionllms} & 83.5 & 38.6 & 59.0 & {45.7} & \NA & \NA \\
LLaDA 1.5~\cite{zhu2025llada15variancereducedpreference} & 83.3 & 35.8$^\dagger$ & 26.2$^\dagger$ & 15.2$^\dagger$ & {52.4} & 42.8 \\
DCoLT~\cite{huang2025reinforcingdiffusionchainlateral} & 88.1 & \NA & \NA & \NA & {59.1} & {51.6}\\
MDPO~\cite{he2025mdpoovercomingtraininginferencedivide} & \NA & 38.6 & 70.7 & \NA & \NA & \NA \\
SAPO~\cite{xie2025stepawarepolicyoptimizationreasoning} & 82.2 & 40.0 & 56.3 & 22.4 &\NA &\NA \\
TraDo-8B-Thinking~\cite{wang2025revolutionizingreinforcementlearningframework} & \cellcolor{green!16}\textbf{94.2} & \cellcolor{green!16}\textbf{87.4} & \NA & \NA & \NA & \NA\\
SPG~\cite{wang2025spgsandwichedpolicygradient} & 86.1 & 41.8 & 71.5  &  {94.0} & 41.5 &50.6\\
DiRL-8B-Instruct \cite{zhu2025dirl} & \cellcolor{gray!20}\underline{93.1} & \cellcolor{gray!20}\underline{85.1} &  \NA & \NA & \NA & \NA\\
d-TreeRPO~\cite{pan2026dtreerporeliablepolicyoptimization} & 82.6 & 46.3 & 71.1 & 92.9 & \NA & \NA \\
dUltra~\cite{chen2026dultraultrafastdiffusionlanguage} & 81.5 & 35.6 & \NA & \NA & 36.8 & 39.1 \\
STP~\cite{liu2026efficientstablereinforcementlearning} & 81.0 & 36.6 & 66.0 & \NA & \NA & \NA \\
ERILS~\cite{lee2026onpolicyexplorationintegratingexternal} & \NA & 40.2 & 73.8 & 91.7 & \NA & \NA \\
LFPO~\cite{wei2026lfpolikelihoodfreepolicyoptimization} & 79.6 & \NA & \NA & \NA & \cellcolor{green!16}\textbf{75.6} & \cellcolor{green!16}\textbf{81.6} \\
StableDRL~\cite{zhong2026stabilizingreinforcementlearningdiffusion} & 92.4 & 79.2 & 84.8 & 92.4 & \NA & \NA \\
RSPO~\cite{yu2026relativescorepolicyoptimization} & 82.6 & 42.1 & 78.8 & 92.1 & \NA & \NA \\
GDSD~\cite{tang2026gdsdreinforcementlearningguided} & 86.4 & 44.6 & \cellcolor{gray!20}\underline{85.6} & 92.0 & 45.7 & 43.6 \\
Mask-Aware PG~\cite{raajesh2026maskawarepolicygradientsdiffusion} & 87.1 & 52.9 & 71.4 & \cellcolor{green!16}\textbf{96.2} & 44.0 & 53.4 \\
ELBO-KTO~\cite{jindal2025aligningdiffusionlanguagemodels} & 82.8 & \NA & \NA & \NA & 42.7 & \NA \\
\midrule
\multicolumn{7}{c}{\cellcolor{lightblue}\textbf{\textit{Non-dLLM baselines}}}\\
\midrule
Qwen2.5-7B-Instruct~\cite{qwen2.5} & 91.6  & 72.8$^\dagger$ & 6.2$^\dagger$ & 21.0$^\dagger$ & 84.8 & 79.2 \\
GPT-4o~\cite{hurst2024gpt} & 92.9$^\dagger$ & 76.6$^\dagger$ & \NA & \NA & 90.2 & 86.8$^\dagger$\\
GPT o4-mini & 99.0$^\dagger$ & 98.2$^\dagger$ & \NA & \NA & 97.3$^\dagger$ & 94.9$^\dagger$\\
\bottomrule
\end{tabular}}
\vspace{-16pt}
\end{table}

\subsubsection{Beyond Standard Token-level GRPO}
In order to adapt RL to dLLM, there are also emerging methods to escape the computation of per-token probability and group advantage. For instance, ESPO~\cite{ou2025principledrldiffusionllms} treats \textit{the entire generated sequence} as a single action and uses the ELBO as a sequence-likelihood proxy; LFPO~\cite{wei2026lfpolikelihoodfreepolicyoptimization} maps flow matching to discrete token space and directly optimizes denoising logits; 
StableDRL~\cite{zhong2026stabilizingreinforcementlearningdiffusion} introduces unconditional clipping and self-normalization to stabilize optimization;
RSPO~\cite{yu2026relativescorepolicyoptimization} calibrates noisy likelihood estimates using reward-implied target ratios;
GDSD~\cite{tang2026gdsdreinforcementlearningguided} derives an advantage-guided self-teacher from reverse-KL-regularized RL and directly distills the denoiser; and Mask-Aware Policy Gradients \cite{raajesh2026maskawarepolicygradientsdiffusion} formulate the generation process as a two-stage
action MDP.

\subsubsection{Aligning with Human Feedback}
\label{sec:Aligning with Human Feedback}
A critical advancement involves aligning dLLMs with human preference, analogous to RLHF in traditional AR methods~\cite{ouyang2022SFT+RLHF}.
LLaDA 1.5~\cite{zhu2025llada15variancereducedpreference} proposes VRPO to mitigate high variance in gradient estimates arising from ELBO approximations while aligning with human feedback.
Specifically, VRPO employs three strategies for unbiased variance reduction: \textbf{\textit{1)}} increasing the number of samples for gradient estimation, \textbf{\textit{2)}} allocating the full budget to timesteps, and \textbf{\textit{3)}} sharing timesteps and masks across policies. In contrast, ELBO-KTO \cite{jindal2025aligningdiffusionlanguagemodels} uses unpaired feedback to reduce dependence on costly preference pairs.

\subsubsection{Advancing Chain-of-Thought Reasoning Abilities}
\label{sec:Boosting Reasoning Abilities}
dLLMs treat intermediate generation steps as latent reasoning states rather than byproducts of text decoding.
Recent RL-based methods have optimized entire Chain-of-Thought (CoT) trajectories for complicated reasoning.
DCoLT~\cite{huang2025reinforcingdiffusionchainlateral} reframes the unmasking process as a latent thinking sequence with outcome-based CoT optimization and then utilizes top-\textit{K} ranking to predict unmasking positions during inference.
SAPO~\cite{xie2025stepawarepolicyoptimizationreasoning} addresses the flawed reasoning in outcome-based RL using a process-based reward to evaluate incremental progress between two timesteps—a larger reward difference indicates a greater contribution to the answer, thereby aligning unmasking steps with the latent reasoning hierarchy.
Beyond reward maximization, DMPO~\cite{zhu2025enhancingreasoningdiffusionllms} enhances CoT through distribution matching, avoiding per-token likelihood estimation via a weighted denoising cross-entropy loss (an off-policy implementation of forward KL).




\subsubsection{RL-Adapted Frameworks}
\label{sec:RL-adapted Frameworks}
TraceRL~\cite{wang2025revolutionizingreinforcementlearningframework} presents an RL framework that integrates a diffusion-based value model, TraDo, in order to reduce variance and enhance training stability.
Rather than relying only on the final response, TraceRL derives rewards from \emph{\textbf{intermediate traces}} generated via dLLMs and aggregates neighboring steps to align the post-training objective with the inference trajectory while utilizing blocked attention for efficient sliced training.
To further improve efficiency and effectiveness, DiRL~\cite{zhu2025dirl} employs unbiased GRPO and FlexAttention in its framework. 
Besides them, SPG~\cite{wang2025spgsandwichedpolicygradient} introduces EUBO, a provable log-likelihood upper bound, to penalize incorrect outputs and overcome ELBO’s one-sided limitation for negative rewards, augmented by a block-wise MC masking strategy.
RFG~\cite{chen2025rfgtesttimescalingdiffusion} generates pseudo-reward signals via log-likelihood ratios between RL/SFT-enhanced and reference dLLMs, enabling reward-free trajectory biasing and outperforming both models without additional training. DARE~\cite{yang2026darediffusionlargelanguage} provides a unified post-training and evaluation framework for dLLMs.

\section{RL for Diffusion Multimodal Models}
\label{sec:dMMs}

In addition to text generation, RL also empowers diffusion-based multimodal models (dMMs) to achieve complex, subjective objectives beyond what standard SFT can capture. By leveraging rewards from human feedback, aesthetic evaluators, or pretrained vision-language models, RL guides the generation process towards higher visual quality, semantic consistency, and cross-modal relevance.
In this section, we review recent RL-driven dMMs from two different perspectives: \textbf{\textit{1)}} RL technology development, i.e., the evolution of task-solving capacity; \textbf{\textit{2)}} RL-related multimodal tasks.


\subsection{Underlying RL Technology Development}
\label{sec:RL's role in dMMs}
\begin{figure}[t]
\centering
\includegraphics[width=0.48\textwidth]{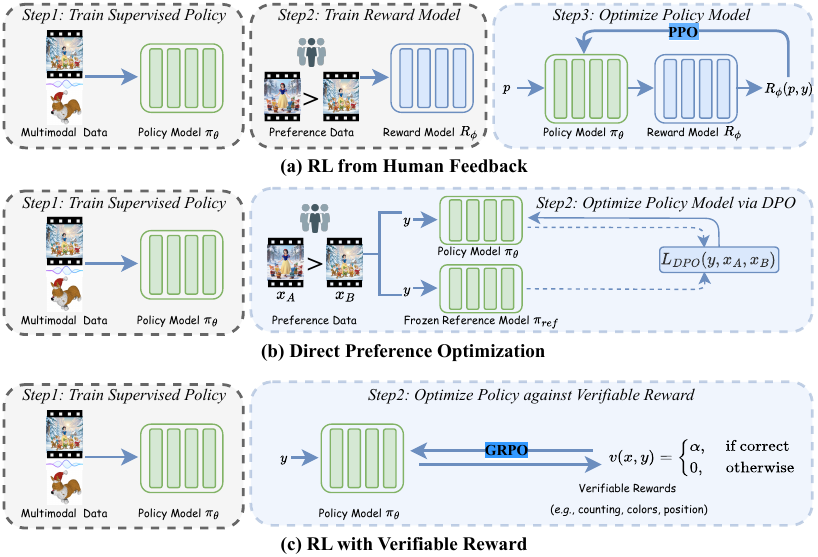}
\vspace{-1em}
\caption{
RLHF and DPO have been the two predominant RL methodologies for human alignment in recent years. In contrast, RLVR represents an emerging trend in RL for dMMs, significantly enhancing their capacity for complex task solving and scalable long-horizon reasoning.
}
\vspace{-1.8em}
\label{fig:rl_role}
\end{figure}

The integration of RL into dMMs has become a promising paradigm for aligning probabilistic generation with human preferences and verifiable, task-specific goals. Existing RL-driven dMMs can be broadly grouped into three complementary categories:
\textbf{\textit{1) Reward-based RLHF}}, which treats the denoising process as a stochastic policy and iteratively aligns outputs with aesthetic, semantic, or factual preferences using human-provided or proxy reward signals.
\textbf{\textit{2) Reward-free DPO}}, which aligns model outputs with human preferences by directly comparing favored and disfavored samples without explicit policy-gradient updates, offering greater stability and sample efficiency compared to conventional RLHF.
\textbf{\textit{3) Rule-based RLVR}}, which introduces scalable and reliable training signals from rule-based verifiable rewards to guide denoising, sampling, or policy generation.
Refer to Fig.~\ref{fig:rl_role} for an overview of these three paradigms.

\subsubsection{Reward-based RLHF}
Traditional DMs are typically trained with reconstruction-driven objectives, such as noise prediction in DDPM~\cite{ho2020DDPM} or velocity prediction in Rectified Flow~\cite{liu2022flow}. While effective for low-level visual fidelity, these objectives provide limited supervision over semantic coherence and human-aligned aesthetics, revealing a gap between pixel-perfect generation and subjective human preferences. RLHF~\cite{christiano2017RLHF}, adapted from LLMs like InstructGPT~\cite{ouyang2022SFT+RLHF}, addresses this gap by framing preference alignment as an optimization problem guided by human feedback, typically instantiated via preference datasets, learned reward models, and RL-based fine-tuning.
And scalable human preference resources, such as SAC~\cite{pressman2022simulacra}, multi-attribute annotations~\cite{lee2023aligning}, PickScore~\cite{kirstain2023pick}, ImageReward~\cite{xu2023imagereward}, and HPSv2~\cite{wu2023human}, standardize evaluation and enable reward learning for visual generation. RLHF-based methods optimize DMs by maximizing learned reward signals, using policy gradient fine-tuning~\cite{fan2023dpok}, Markov Decision Process interpretations of denoising trajectories~\cite{black2024training}, or direct reward-driven gradient propagation~\cite{clark2023directly,prabhudesai2023aligning}, and have also been extended to multi-dimensional video generation~\cite{liu2025improving}. Despite early success, reliance on static reward models and fixed feedback distributions can limit generalization and induce instabilities such as reward hacking, motivating the development of more robust and intrinsic alignment strategies beyond explicit reward optimization.

\subsubsection{Reward-free DPO}
DPO~\cite{rafailov2023direct} addresses the limitations of static reward models in RLHF~\cite{christiano2017RLHF} by directly optimizing DMs via pairwise human preferences, simplifying training and improving stability.
Conceptually, DPO~\cite{rafailov2023direct} acts as an implicit RL framework, aligning distributions of preferred and dispreferred outputs without explicit reward estimation, reducing bias and enhancing sample efficiency.
Specifically, extensions to diffusion tasks include Diffusion-DPO~\cite{wallace2024diffusion} for text-to-image generation, Diffusion-KTO~\cite{li2024aligning} via KL-target optimization to remove reward modeling, VideoDPO~\cite{liu2025videodpo} for temporal generation, and OnlineVPO~\cite{zhang2024onlinevpo} along with online preference collection for iterative alignment.
Moreover, industrial-scale models such as Stable Diffusion 3~\cite{esser2024scaling}, MovieGen~\cite{polyak2024movie}, HunyuanVideo~\cite{kong2024hunyuanvideo}, StepVideo~\cite{ma2025step}, Cosmos~\cite{agarwal2025cosmos}, and Wan~\cite{wan2025wan} further adopt RLHF or DPO post-training to improve alignment beyond pretraining. These advances mark a shift from reward-centric RL to preference-driven direct optimization. Notably, reliance on static preference data limits adaptability, driving exploration of dynamic and verifiable methods.

\subsubsection{Rule-based RLVR}
Existing human alignment methods, such as RLHF~\cite{christiano2017RLHF} and DPO~\cite{rafailov2023direct}, depend on static annotations or fixed reward models, which limit the optimization space and struggle with complex objectives like multi-attribute consistency, compositional constraints, or dynamic task rewards. To address these limitations, RL with verifiable and dynamic rewards has been proposed, incorporating external signals during or after sampling for adaptive, task-driven optimization. Introducing verifiable rewards shifts DMs from static alignment to dynamic optimization, though challenges arise from the deterministic and continuous nature of diffusion and flow matching. Existing solutions fall into three categories: probabilistic reparameterization, introducing learnable stochasticity for differentiable reward propagation (\textit{e.g.}, F5R-TTS~\cite{sun2025f5r}); sampling randomization and policy modeling, reformulating diffusion as a stochastic decision process for GRPO-style trajectory optimization (\textit{e.g.}, Flow-GRPO~\cite{liu2025flow}, DanceGRPO~\cite{xue2025dancegrpo}, MixGRPO~\cite{li2025mixgrpo}); and implicit policy improvement using structured reward guidance or energy-based formulations for interpretable optimization (\textit{e.g.}, DiffusionNFT~\cite{zheng2025diffusionnft}, Nabla-GFlowNet~\cite{liu2024efficient}). Post-training inference optimization frameworks such as ReNo~\cite{eyring2024reno} and SDPO~\cite{zhang2024aligning}, along with adaptive sampling control methods~\cite{zhang2023adadiff,wang2023learning}, further demonstrate verifiable feedback without full retraining. Collectively, these advances mark a shift toward feedback-driven, verifiable RL for diffusion-based generation.

\subsection{RL-driven Diffusion Multimodal Tasks}
\label{sec:RL4dMMs applications}
\begin{figure}[t]
\centering
\includegraphics[width=0.48\textwidth]{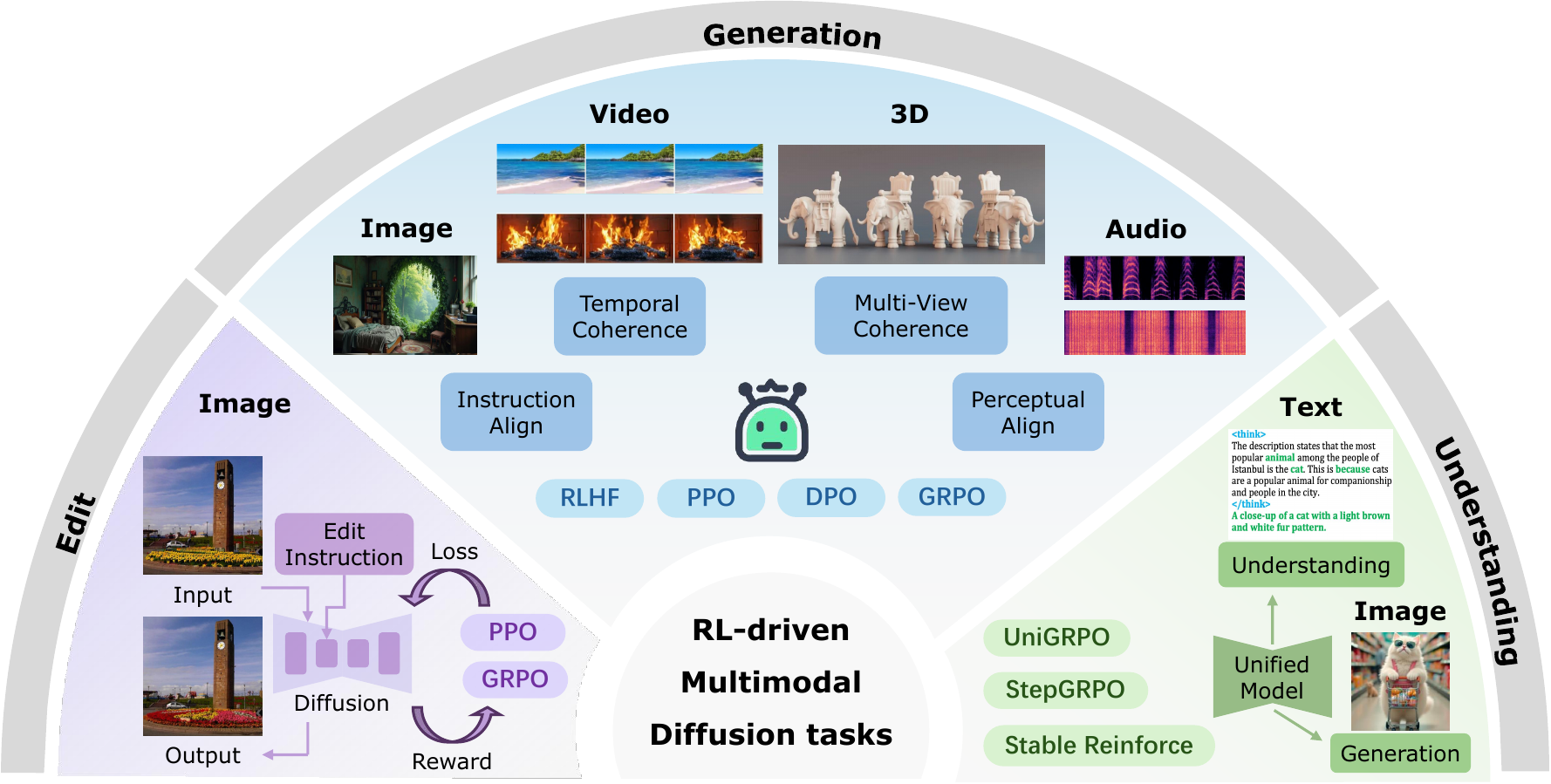}
\vspace{-6pt}
\caption{
The overview of RL-driven diffusion multimodal tasks.
}
\vspace{-1.8em}
\label{fig:MM_tasks}
\end{figure}

The combination of RL and DMs provides a critical tool for fine-grained control across diverse modalities like image, video, 3D, and audio. Here we review how RL-based strategies are formulated to solve three multimodal tasks (Fig.~\ref{fig:MM_tasks}): optimizing \textbf{\textit{multimodal generation}}, enabling precise \textbf{\textit{content editing}}, and fostering \textbf{\textit{unified understanding and generation}}.

\subsubsection{RL-driven Multimodal Generation}
\label{sec:generation}

In multimodal generation, RL has been increasingly adopted as a general alignment and control tool for DMs beyond text-only conditions.
This subsection reviews existing works according to the modality-specific challenges encountered in generative modeling and discusses how RL improves generation quality, controllability, and efficiency in different scenarios, including image, video, 3D, and audio.

\vspace{1em} 
\noindent \textbf{RL for Image Generation.}
\textcolor{black}{Reward and preference optimization complement supervised finetuning for generation quality and task-specific constraints.}
This leads the community to formalize three core tasks, each supported by a dedicated evaluation ecosystem: \textcolor{black}{\textbf{\textit{human-preference and visual-quality alignment}}}, assessed via pretrained reward models such as PickScore~\cite{kirstain2023pick}, ImageReward~\cite{xu2023imagereward}, and HPSv2~\cite{wu2023human}; \textbf{\textit{compositional reasoning}}, \textcolor{black}{evaluated using GenEval~\cite{ghosh2023geneval} and TIFA~\cite{hu2023tifa}}; and \textbf{\textit{inference acceleration}}, which characterizes the quality-efficiency trade-off.
As shown in Table~\ref{tab:rl4diff-image-methods}, we categorize existing methods according to their underlying optimization paradigms.
And Table~\ref{tab:image-generation-quantitative} summarizes paired base-to-aligned improvements reported in the original papers for representative \textbf{\textit{image generation}} approaches on multiple widely used metrics.

\begin{table}[t!]
\centering
\scriptsize
\setlength{\tabcolsep}{2pt}
\caption{\textcolor{black}{Image-generation alignment and related methods. Non-exclusive columns: Online refreshes samples; RLHF includes human/AI preference and quality feedback; DPO denotes direct-preference methods; RLVR uses target/rule feedback, possibly model-assisted. RePrompt trains a prompt policy; VARD's offline stage pretrains its value model.}}
\vspace{-6pt}
\label{tab:rl4diff-image-methods}
\begin{tabular}{ccccccccccc}
\toprule
\multirow{2}{*}{\textbf{Methods}} & \multicolumn{2}{c}{\textbf{Training Strategy}} & \multicolumn{3}{c}{\textbf{Feedback Paradigm}} & \multirow{2}{*}{\textbf{Links}} \\
\cmidrule(r){2-3} \cmidrule(r){4-6}
& \textbf{Online} & \textbf{Offline} & \textbf{RLHF} & \textbf{DPO} & \textbf{RLVR} & \\

\midrule
\multicolumn{7}{c}{\cellcolor{lightblue}\textit{\textbf{Outcome-level Alignment (Global Reward Optimization)}}} \\
\midrule

Diffusion-DPO~\cite{wallace2024diffusion} & & \cmark & \textcolor{black}{\cmark} & \cmark & & \Paper{https://arxiv.org/abs/2311.12908} \Code{https://github.com/SalesforceAIResearch/DiffusionDPO} \\

DPOK~\cite{fan2023dpok} & \cmark & & \cmark & & & \Paper{https://arxiv.org/abs/2305.16381} \Code{https://github.com/google-research/google-research/tree/master/dpok} \\

DDPO~\cite{black2024training} & \cmark & & \cmark & & \textcolor{black}{\cmark} & \Paper{https://arxiv.org/abs/2305.13301} \Code{https://github.com/jannerm/ddpo}\,\href{https://huggingface.co/kvablack/ddpo-compressibility}{\faHuggingFace} \\

LargeRL~\cite{zhang2024largerl} & \cmark & & \cmark & & \textcolor{black}{\cmark} & \Paper{https://arxiv.org/abs/2401.12244} \Code{https://github.com/pinterest/atg-research} \\


Diffusion-KTO~\cite{li2024aligning} &  & \cmark & \textcolor{black}{\cmark} & \cmark & & \Paper{https://arxiv.org/abs/2404.04465} \Code{https://github.com/jacklishufan/diffusion-kto} \,\href{https://huggingface.co/jacklishufan/diffusion-kto}{\faHuggingFace} \\

\textcolor{black}{DiverseRL~\cite{Miao2024DiverseRL}} & \cmark & & & & & \Paper{https://openaccess.thecvf.com/content/CVPR2024/html/Miao_Training_Diffusion_Models_Towards_Diverse_Image_Generation_with_Reinforcement_Learning_CVPR_2024_paper.html} \\

Parrot~\cite{lee2024parrot} & \cmark & & \cmark & & & \Paper{https://arxiv.org/abs/2401.05675} \\



\textcolor{black}{DSPO~\cite{ICLR2025_7f70331d}} & & \textcolor{black}{\cmark} & \textcolor{black}{\cmark} & \textcolor{black}{\cmark} & & \Paper{https://proceedings.iclr.cc/paper_files/paper/2025/hash/7f70331dbe58ad59d83941dfa7d975aa-Abstract-Conference.html} \Code{https://github.com/huaishengzhu/DSPO} \\

HERO~\cite{hiranaka2025hero} & \cmark & & \cmark & & & \Paper{https://arxiv.org/abs/2410.05116} \Code{https://github.com/sony/hero} \\

B$^2$-DiffuRL~\cite{hu2025betteralignment} & \cmark & & \cmark & & & \Paper{https://arxiv.org/abs/2503.11240} \Code{https://github.com/hu-zijing/B2-DiffuRL} \\


\textcolor{black}{FiFA~\cite{yang2025fifa}} & & \textcolor{black}{\cmark} & \textcolor{black}{\cmark} & \textcolor{black}{\cmark} & & \Paper{https://arxiv.org/abs/2410.10166} \Code{https://github.com/YangYongJin/FiFA-Official-Repo} \\

\textcolor{black}{VARD~\cite{dai2025vard}} & \textcolor{black}{\cmark} & \textcolor{black}{\cmark} & \textcolor{black}{\cmark} & & \textcolor{black}{\cmark} & \Paper{https://arxiv.org/abs/2505.15791} \\

\textcolor{black}{Score-Action~\cite{zhao2025scoreaction}} & \textcolor{black}{\cmark} & & \textcolor{black}{\cmark} & & & \Paper{https://arxiv.org/abs/2502.01819} \\

CaPO~\cite{lee2025CAPO} & & \cmark & \textcolor{black}{\cmark} & \cmark & & \Paper{https://arxiv.org/abs/2502.02588} \\

FreeLunch~\cite{xian2025freelunch} & & \cmark & \textcolor{black}{\cmark} & \cmark & & \Paper{https://arxiv.org/abs/2509.25771} \Code{https://github.com/DSL-Lab/T2I-Free-Lunch-Alignment} \\

Diff-LPO~\cite{bai2025listwisepreference} & & \cmark & \textcolor{black}{\cmark} & \cmark & & \Paper{https://arxiv.org/abs/2510.01540} \\

\textcolor{black}{Flow-GRPO~\cite{liu2025flow}} & \textcolor{black}{\cmark} & & \textcolor{black}{\cmark} & & \textcolor{black}{\cmark} & \Paper{https://arxiv.org/abs/2505.05470} \Code{https://github.com/yifan123/flow_grpo}\,\href{https://huggingface.co/collections/jieliu/sd35m-flowgrpo}{\faHuggingFace} \\

\textcolor{black}{DiffusionNFT~\cite{zheng2025diffusionnft}} & \textcolor{black}{\cmark} & & \textcolor{black}{\cmark} & & \textcolor{black}{\cmark} & \Paper{https://arxiv.org/abs/2509.16117} \Code{https://github.com/NVlabs/DiffusionNFT}\,\href{https://huggingface.co/worstcoder/SD3.5M-DiffusionNFT-MultiReward}{\faHuggingFace} \\

\textcolor{black}{AWM~\cite{xue2025awm}} & \textcolor{black}{\cmark} & & \textcolor{black}{\cmark} & & \textcolor{black}{\cmark} & \Paper{https://arxiv.org/abs/2509.25050} \Code{https://github.com/scxue/advantage_weighted_matching} \\

\noalign{\nocite{dufour2025miro}}

Adv-GRPO~\cite{mao2025advgrpo} & \cmark & & \cmark & & \cmark & \Paper{https://arxiv.org/abs/2511.20256} \Code{https://github.com/showlab/Adv-GRPO}\,\href{https://huggingface.co/benzweijia/Adv-GRPO}{\faHuggingFace} \\

{V-GRPO~\cite{tang2026vgrpoonlinereinforcementlearning}} & {\cmark} & & {\cmark} & & {\cmark} & {\Paper{https://arxiv.org/abs/2604.23380}} \\

{RVM~\cite{choi2026scalingreinforcementlearningdiffusion}} & {\cmark} & & {\cmark} & & {\cmark} & {\Paper{https://arxiv.org/abs/2608.23664}} \\

\textcolor{black}{DiffusionOPSD~\cite{zhou2026onpolicyselfdistillationdiffusionmodels}} & \textcolor{black}{\cmark} & & \textcolor{black}{\cmark} & & & {\Paper{https://arxiv.org/abs/2608.24646}} \\

{Qwen-Image-2.0-RL~\cite{xu2026qwenimage20rltechnicalreport}} & {\cmark} & & {\cmark} & & & {\Paper{https://arxiv.org/abs/2606.27608}} \\

\midrule
\multicolumn{7}{c}{\cellcolor{lightblue}\textit{\textbf{Process-level Control (Structural Reasoning)}}} \\
\midrule

\textcolor{black}{RePrompt~\cite{wu2025reprompt}} & \textcolor{black}{\cmark} & & \textcolor{black}{\cmark} & & \textcolor{black}{\cmark} & \Paper{https://arxiv.org/abs/2505.17540} \Code{https://github.com/microsoft/DKI_LLM/tree/main/RePrompt} \\

\textcolor{black}{SRRL~\cite{pan2025srrl}} & \textcolor{black}{\cmark} & & \textcolor{black}{\cmark} & & & \Paper{https://arxiv.org/abs/2505.22407} \Code{https://github.com/jadenpan0/SRRL} \\

\noalign{\nocite{jiang2025t2ir1}}

\noalign{\nocite{guo2025cotgenerationimage}}

\noalign{\nocite{liu2025cotlized}}

\textcolor{black}{Visual-CoG~\cite{li2025visualcog}} & \textcolor{black}{\cmark} & & \textcolor{black}{\cmark} & & \textcolor{black}{\cmark} & \Paper{https://arxiv.org/abs/2508.18032} \\

\textcolor{black}{HiCoGen~\cite{yang2025hicogen}} & \textcolor{black}{\cmark} & & \textcolor{black}{\cmark} & & \textcolor{black}{\cmark} & \Paper{https://arxiv.org/abs/2511.19965} \\

{DenseGRPO~\cite{deng2026densegrposparsedensereward}} & {\cmark} & & {\cmark} & & {\cmark} & {\Paper{https://arxiv.org/abs/2601.20218}} \\

{Stepwise-Flow-GRPO~\cite{savani2026stepwisecreditassignmentgrpo}} & {\cmark} & & {\cmark} & & {\cmark} & {\Paper{https://arxiv.org/abs/2603.28718}} \\

\textcolor{black}{OTCA~\cite{li2026learningcreditrightsteps}} & \textcolor{black}{\cmark} & & \textcolor{black}{\cmark} & & & {\Paper{https://arxiv.org/abs/2604.19234}} \\

\midrule
\multicolumn{7}{c}{\cellcolor{lightblue}\textit{\textbf{\textcolor{black}{Efficiency \& Trajectory Optimization}}}} \\
\midrule

RLCM~\cite{oertell2024rlcm} & \cmark & & \cmark & & \cmark & \Paper{https://arxiv.org/abs/2404.03673} \Code{https://github.com/Owen-Oertell/rlcm} \\

\textcolor{black}{SDPO~\cite{zhang2024aligning}} & \textcolor{black}{\cmark} & & \textcolor{black}{\cmark} & & & \Paper{https://arxiv.org/abs/2411.11727} \Code{https://github.com/ZiyiZhang27/sdpo} \\

\textcolor{black}{Diff-Sharp~\cite{tian2025diffusionsharpening}} & \textcolor{black}{\cmark} & & \textcolor{black}{\cmark} & \textcolor{black}{\cmark} & & \Paper{https://arxiv.org/abs/2502.12146} \Code{https://github.com/Gen-Verse/Diffusion-Sharpening} \\

\textcolor{black}{R0~\cite{luo2025rewardinstruct}} & \textcolor{black}{\cmark} & & \textcolor{black}{\cmark} & & & \Paper{https://arxiv.org/abs/2503.13070} \Code{https://github.com/Luo-Yihong/Reward-Instruct} \\

\textcolor{black}{Flash-DMD~\cite{chen2025flashdmd}} & \textcolor{black}{\cmark} & & \textcolor{black}{\cmark} & \textcolor{black}{\cmark} & & \Paper{https://arxiv.org/abs/2511.20549} \\

ProxT2I~\cite{fang2025proxt2i} & \cmark & & \cmark & & & \Paper{https://arxiv.org/abs/2511.18742} \\

\textcolor{black}{DMDR~\cite{jiang2025dmdr}} & \textcolor{black}{\cmark} & & \textcolor{black}{\cmark} & & & \Paper{https://arxiv.org/abs/2511.13649} \Code{https://github.com/vvvvvjdy/dmdr}\,\href{https://huggingface.co/Tongyi-MAI/Z-Image-Turbo}{\faHuggingFace} \\

\bottomrule
\end{tabular}
\vspace{-18pt}
\end{table}

For \textcolor{black}{\textbf{\textit{human-preference and visual-quality alignment}}}, the dominant framework is \textbf{\textit{outcome-level alignment}}, i.e., \textbf{\textit{global reward optimization}}, \textcolor{black}{which uses final-output feedback to shift the generation distribution towards high-reward regions.}
Early methods rely on policy-gradient updates (\textit{e.g.}, DDPO~\cite{black2024training}) or KL-regularized objectives (\textit{e.g.}, DPOK~\cite{fan2023dpok}) to optimize image-level rewards.
While effective for black-box metrics, these high-variance updates are suboptimal when rewards are differentiable.
In such cases, gradient-based optimization offers a lower-variance alternative.
\textcolor{black}{DRaFT~\cite{clark2023directly} backpropagates reward gradients through the denoising chain, using LoRA and gradient checkpointing to reduce memory costs; its DRaFT-K variant truncates backpropagation to the final $K$ sampling steps.}
Subsequent works recognize preference alignment as a multi-objective problem: \textcolor{black}{LargeRL~\cite{zhang2024largerl} scales RL to millions of prompts, whereas DiverseRL~\cite{Miao2024DiverseRL} explicitly promotes generation diversity}, while Parrot~\cite{lee2024parrot} models the Pareto frontier between visual quality and semantic fidelity.
Additionally, to address the potential mode collapse caused by reward maximization, $\nabla$-GFlowNet~\cite{liu2024efficient} employs the gradient-informed GFlowNet objective to align samples with reward distributions, explicitly balancing high rewards and the generative prior.

\begin{figure}[t]
\centering
\includegraphics[width=0.48\textwidth]{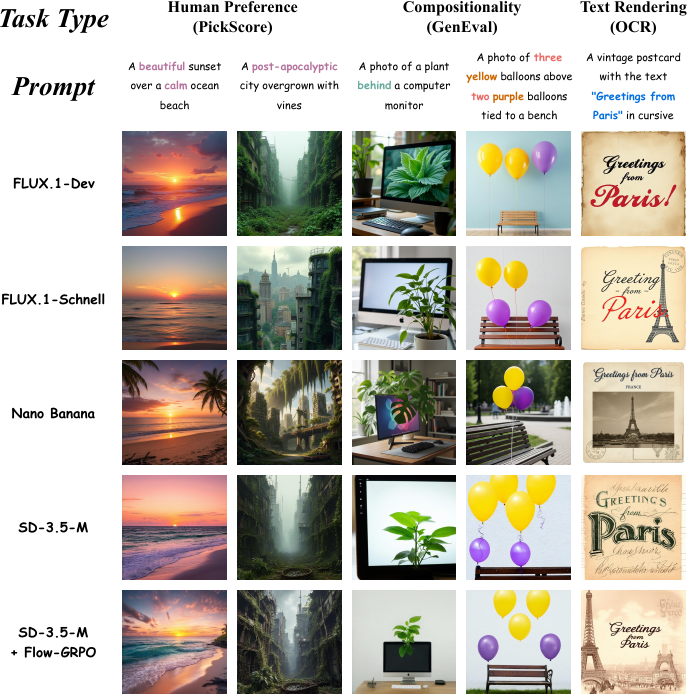}
\vspace{-4pt}
\caption{
\textcolor{black}{Qualitative examples from SD-3.5-M, SD-3.5-M with Flow-GRPO, FLUX.1-dev, FLUX.1-schnell, and Nano Banana. Task labels indicate prompt categories, not benchmark evaluations. The Flow-GRPO checkpoint configuration is unspecified; these examples are illustrative, not controlled benchmark comparisons or evidence of limitations of outcome-based rewards.}
}
\vspace{-1.8em}
\label{fig:flow_grpo_comparison}
\end{figure}

Recent advances further emphasize \textit{\textbf{algorithmic simplicity}} and \textbf{\textit{training stability}}.
\textcolor{black}{Diffusion-DPO~\cite{wallace2024diffusion} adapts DPO~\cite{rafailov2023direct} to diffusion models, whereas DSPO~\cite{ICLR2025_7f70331d} aligns text-to-image diffusion models through preference score matching; neither requires separately training an explicit reward model. HERO~\cite{hiranaka2025hero} instead learns a similarity-based reward from human feedback and fine-tunes the diffusion model using DDPO.}
And Diffusion-KTO~\cite{li2024aligning} explores the necessity of pairwise preference data and shows that alignment is achievable utilizing only \textit{unary (like/dislike) feedback}, thereby simplifying data collection and avoiding complex reward design.
Moreover, this paradigm has been extended to flow-based backbones, where Flow-GRPO~\cite{liu2025flow}, MixGRPO~\cite{li2025mixgrpo}, and Multi-GRPO~\cite{lyu2025multi} apply GRPO~\cite{shao2024deepseekmath} to continuous flow trajectories, \textcolor{black}{with mechanisms that improve training-rollout or optimization efficiency.}
Pushing the flow-based paradigm further, DiffusionNFT~\cite{zheng2025diffusionnft} proposes an online RL framework that performs policy improvement directly on the \emph{forward} process of flow matching by contrasting positive and negative generations, thereby avoiding trajectory-based optimization.
\textcolor{black}{Flow-GRPO~\cite{liu2025flow} improves the OCR score from 0.59 to 0.92 using an OCR-specific model with KL regularization (Table~\ref{tab:image-generation-quantitative}). Fig.~\ref{fig:flow_grpo_comparison} presents selected qualitative examples rather than aggregate performance.}

Rather than an explicit denoising-MDP objective, V-GRPO~\cite{tang2026vgrpoonlinereinforcementlearning} retains an ELBO-based ratio surrogate with variance reduction and controlled updates. RVM~\cite{choi2026scalingreinforcementlearningdiffusion} instead reinforces native velocity-field directions with an anchor controlling policy drift, without trajectory likelihoods.

And \textcolor{black}{for \textbf{\textit{compositional reasoning}}, \textbf{\textit{process-level control}} spans planning, verification, and stage-aware feedback. As a related autoregressive approach rather than a diffusion method, T2I-R1~\cite{jiang2025t2ir1} uses final-image rewards for reasoning and image-token generation. PARM~\cite{guo2025cotgenerationimage} provides test-time verification, distinct from the same study's iterative DPO. Visual-CoG~\cite{li2025visualcog} uses stage-aware rewards for planning and masked image-token generation.}

Process-level RL is also moving from external reasoning traces into the denoising dynamics themselves. Specifically, DenseGRPO~\cite{deng2026densegrposparsedensereward} scores intermediate clean estimates and utilizes step-wise reward gains with timestep-aware exploration, whereas Stepwise-Flow-GRPO~\cite{savani2026stepwisecreditassignmentgrpo} derives gain-based advantages from reward changes between Tweedie estimates. They both expose temporal credit assignment as a distinct process-control problem rather than assigning the terminal reward uniformly to every denoising step.

Finally, for \textbf{\textit{inference acceleration}}, the third paradigm, \textbf{\textit{efficiency-oriented alignment}}, \textcolor{black}{combines reward feedback with few-step generation, distillation, or adaptive sampling.}
\textcolor{black}{RLCM~\cite{oertell2024rlcm} and SDPO~\cite{zhang2024aligning} align few-step generators. Diffusion-Sharpening~\cite{tian2025diffusionsharpening} uses reward-guided trajectories for online supervised finetuning or preference optimization, not necessarily fewer sampling steps. AdaDiff~\cite{zhang2023adadiff} selects a predefined sampling-step budget using a prompt-conditioned policy.}
Another emerging direction explores \textit{joint distillation}, as in Flash-DMD~\cite{chen2025flashdmd} and \textcolor{black}{DMDR}~\cite{jiang2025dmdr}, where student models are trained to simultaneously match a teacher distribution and maximize reward, \textcolor{black}{improving the latency--quality trade-off in the reported settings.}
Lastly, distinct from these weight-updating methods, ReNO~\cite{eyring2024reno} introduces a \textit{training-free test-time alignment} strategy. Instead of updating weights, ReNO directly optimizes the initial latent noise via reward guidance, effectively trading increased inference-time computation for superior preference and compositional adherence.

A complementary system-level advance is Qwen-Image-2.0-RL~\cite{xu2026qwenimage20rltechnicalreport}. It combines task-specific composite rewards, scalable GRPO, prompt filtering, and category-wise reward calibration, then merges specialized text-to-image and editing policies into a single model through on-policy velocity distillation. This pipeline extends RL post-training beyond single-reward algorithm prototypes to a joint generation-and-editing deployment setting.

\begin{table}[t]
\footnotesize
\centering
\setlength{\tabcolsep}{2pt}
\caption{\textcolor{black}{Video alignment and related methods; columns follow Table~\ref{tab:rl4diff-image-methods}. Online includes direct reward-gradient tuning. VideoRM reports DRaFT-V; PISA combines offline/online stages; VideoAlign uses Flow-DPO/Flow-RWR. $^\dagger$: reward-weighted regression. Align-A-Video performs per-instance editing; RLVR includes imperfect proxies.}}
\vspace{-6pt}
\label{tab:rl4diff-video-methods}
\begin{tabular}{ccccccc}
\toprule
\multirow{2}{*}{\textbf{Methods}} & \multicolumn{2}{c}{\textbf{Training Strategy}} & \multicolumn{3}{c}{\textbf{Feedback Paradigm}} & \multirow{2}{*}{\textbf{Links}} \\
\cmidrule(r){2-3} \cmidrule(r){4-6}
& \textbf{Online} & \textbf{Offline} & \textbf{RLHF} & \textbf{DPO} & \textbf{RLVR} & \\

\midrule
\multicolumn{7}{c}{\cellcolor{lightblue}\textit{\textbf{Temporal \& Cross-modal Alignment (Motion Quality \& Coherence)}}} \\
\midrule

InstructVideo~\cite{Yuan_2024_CVPR} & \cmark & & \cmark & & & \Paper{https://arxiv.org/abs/2312.12490} \Code{https://github.com/ali-vilab/VGen/blob/main/doc/InstructVideo.md} \\

VideoDPO~\cite{liu2025videodpo} & & \cmark & \textcolor{black}{\cmark} & \cmark & & \Paper{https://arxiv.org/abs/2412.14167} \Code{https://github.com/CIntellifusion/VideoDPO} \\

\textcolor{black}{VADER~\cite{Prabhudesai_2024_VADER}} & \textcolor{black}{\cmark} & & \textcolor{black}{\cmark} & & \textcolor{black}{\cmark} & \Paper{https://arxiv.org/abs/2407.08737} {\Code{https://github.com/mihirp1998/VADER}} \\

\textcolor{black}{T2V-Turbo~\cite{Li_2024_NeurIPS}} & \textcolor{black}{\cmark} & & \textcolor{black}{\cmark} & & & \Paper{https://arxiv.org/abs/2405.18750} {\Code{https://github.com/Ji4chenLi/t2v-turbo}} \\

\textcolor{black}{VideoRM~\cite{Wu_2024_NeurIPS}} & \textcolor{black}{\cmark} & & \textcolor{black}{\cmark} & & & \Paper{https://openreview.net/pdf?id=3ivnixHy16} \\

\textcolor{black}{VideoAlign~\cite{liu2025improving}} & & \textcolor{black}{\cmark} & \textcolor{black}{\cmark} & \textcolor{black}{\cmark} & & \Paper{https://arxiv.org/abs/2501.13918} \Code{https://github.com/KlingTeam/VideoAlign}\,\href{https://huggingface.co/KlingTeam/VideoReward}{\faHuggingFace} \\

DenseDPO~\cite{wu2025densedpo} & & \cmark & \textcolor{black}{\cmark} & \cmark & & \Paper{https://arxiv.org/abs/2506.03517} \Code{https://github.com/snap-research/DenseDPO} \\

\textcolor{black}{cIPO~\cite{liu2026temporalconcentrationrollouterrors}} & \textcolor{black}{\cmark} & & & \textcolor{black}{\cmark} & & {\Paper{https://arxiv.org/abs/2607.28058}} \\

OmniNFT~\cite{zhang2026omninftmodalitywiseomnidiffusion} & {\cmark} & & {\cmark} & & & {\Paper{https://arxiv.org/abs/2605.12480}} \\

\midrule
\multicolumn{7}{c}{\cellcolor{lightblue}\textit{\textbf{Physical Dynamics (Causality \& Logic)}}} \\
\midrule

PhysAR~\cite{Lin_2025_PhysAR} & \cmark & & & & \cmark & \Paper{https://arxiv.org/abs/2504.15932} \\

\textcolor{black}{PhysMaster~\cite{ji2025physmaster}} & & \textcolor{black}{\cmark} & \textcolor{black}{\cmark} & \textcolor{black}{\cmark} & & \Paper{https://arxiv.org/abs/2510.13809} {\Code{https://github.com/KlingTeam/PhysMaster}} \\

PhysCorr~\cite{wang2025physcorr} & & \cmark & \textcolor{black}{\cmark} & \cmark & & \Paper{https://arxiv.org/abs/2511.03997} \\

\textcolor{black}{PISA~\cite{li2025pisa}} & \textcolor{black}{\cmark} & \textcolor{black}{\cmark} & & & \textcolor{black}{\cmark} & \Paper{https://arxiv.org/abs/2503.09595} \Code{https://github.com/vision-x-nyu/pisa-experiments}\,\href{https://huggingface.co/datasets/nyu-visionx/pisa-experiments}{\faHuggingFace} \\

World-R1~\cite{wang2026worldr1reinforcing3dconstraints} & {\cmark} & & {\cmark} & & {\cmark} & {\Paper{https://arxiv.org/abs/2604.24764}} \\

{VGGRPO~\cite{an2026vggrpoworldconsistentvideogeneration}} & {\cmark} & & & & {\cmark} & {\Paper{https://arxiv.org/abs/2603.26599}} \\

\midrule
\multicolumn{7}{c}{\cellcolor{lightblue}\textit{\textbf{Scalability \& Identity (GRPO \& Consistency)}}} \\
\midrule

\textcolor{black}{Align-A-Video~\cite{Wang_2025_CVPR}} & \textcolor{black}{\cmark} & & \textcolor{black}{\cmark} & & & \Paper{https://openaccess.thecvf.com/content/CVPR2025/papers/Wang_Align-A-Video_Deterministic_Reward_Tuning_of_Image_Diffusion_Models_for_Consistent_CVPR_2025_paper.pdf} \\

DanceGRPO~\cite{xue2025dancegrpo} & \cmark & & \cmark & & & \Paper{https://arxiv.org/abs/2505.07818} \Code{https://github.com/XueZeyue/DanceGRPO} \\

\textcolor{black}{GigaVideo-1$^\dagger$~\cite{bao2025gigavideo}} & & \textcolor{black}{\cmark} & \textcolor{black}{\cmark} & & & \Paper{https://arxiv.org/abs/2506.10639} \Code{https://github.com/GigaAI-research/GigaVideo-1} \\

\textcolor{black}{MagicID~\cite{li2025magicid}} & & \textcolor{black}{\cmark} & \textcolor{black}{\cmark} & \textcolor{black}{\cmark} & & \Paper{https://arxiv.org/abs/2503.12689} \Code{https://github.com/EchoPluto/MagicID} \\

\textcolor{black}{ID-Crafter~\cite{pan2025idcrafter}} & \textcolor{black}{\cmark} & & \textcolor{black}{\cmark} & & \textcolor{black}{\cmark} & \Paper{https://arxiv.org/abs/2511.00511} \Code{https://github.com/paulpanwang/IDCrafter} \\

{TAGRPO~\cite{wang2026tagrpoboostinggrpoimagetovideo}} & {\cmark} & & {\cmark} & & & {\Paper{https://arxiv.org/abs/2601.05729}} \\

\bottomrule
\vspace{-18pt}
\end{tabular}
\end{table}

\vspace{1em} 
\noindent\textbf{RL in video generation.}
Extending DMs from images to videos introduces temporal dynamics, substantially increasing alignment complexity. \textcolor{black}{Reward-based alignment targets temporal coherence and computational efficiency alongside supervised training. InstructVideo~\cite{Yuan_2024_CVPR} uses image-level rewards with segmental selection and temporal weighting, whereas VideoDPO~\cite{liu2025videodpo} constructs preferences using OmniScore, including motion smoothness, flicker, and subject consistency.} \textcolor{black}{For differentiable rewards, VADER~\cite{Prabhudesai_2024_VADER} backpropagates reward gradients through video denoising, using truncated backpropagation, LoRA, and gradient checkpointing to reduce optimization costs.} To consolidate these techniques, the VideoReward system~\cite{liu2025improving} proposes a holistic pipeline with \textcolor{black}{Flow-DPO and Flow-RWR for training} and Flow-NRG for inference-time guidance. Moreover, to address the limitations of static supervision, OnlineVPO~\cite{zhang2024onlinevpo} further employs a VQA-based reward model within an online DPO framework. Notably, the high cost and instability of video-level RL highlight the need for \textit{scalable RL optimization}. A key advance is DanceGRPO~\cite{xue2025dancegrpo}, which introduces critic-free GRPO to visual generation. By replacing explicit value networks with group-based baselines, DanceGRPO stabilizes large-scale RLHF across text-to-image, text-to-video, and image-to-video settings.

\begin{table*}[t]
\centering
\caption{Paired quantitative effects of representative RL/RL-family image generation methods. All values are collected from the reported tables in the original papers: DPOK's Table 1~\cite{fan2023dpok}, AlignProp's Table 1 \textcolor{black}{(v2)}~\cite{prabhudesai2023aligning}, DanceGRPO's Tables 2--4~\cite{xue2025dancegrpo}, and Flow-GRPO's Tables 1--2~\cite{liu2025flow}, together with Qwen-Image-2.0-RL's Table 1 and arena evaluation~\cite{xu2026qwenimage20rltechnicalreport}. Rows should be interpreted as paired base-to-aligned improvements under the corresponding source protocol rather than as a single cross-paper leaderboard. \textcolor{black}{SD v1.4 DDPO/ReFL/DanceGRPO HPS and CLIP results are DanceGRPO's reward-specific runs; its Pick-a-Pic/GenEval results use joint rewards. DPOK is evaluated on training prompts. Flow-GRPO uses three separately trained task-specific models with KL regularization; OCR score is based on normalized edit distance, not the fraction of images with entirely correct text.}}
\vspace{-1em}
\label{tab:image-generation-quantitative}
\footnotesize
\setlength{\tabcolsep}{5pt}
\begin{tabular}{lllccc}
\toprule
\textbf{Base Models} & \textbf{Methods} & \textbf{Metrics} & \textbf{Base} & \textbf{Aligned} & \textbf{$\Delta$} \\
\midrule
\multirow{6}{*}{Stable Diffusion v1.4}
& \multirow{2}{*}{DDPO~\cite{black2024training}} & HPS-v2.1 & 0.239 & 0.297 & +0.058 \\
& & CLIP & 0.363 & 0.381 & +0.018 \\
\cmidrule(lr){2-6}
& \multirow{2}{*}{ReFL~\cite{xu2023imagereward}} & HPS-v2.1 & 0.239 & 0.357 & +0.118 \\
& & CLIP & 0.363 & 0.418 & +0.055 \\
\cmidrule(lr){2-6}
& \multirow{2}{*}{DanceGRPO~\cite{xue2025dancegrpo}} & HPS-v2.1 & 0.239 & 0.365 & +0.126 \\
& & CLIP & 0.363 & 0.421 & +0.058 \\
\midrule
\multirow{4}{*}{Stable Diffusion v1.5}
& \multirow{2}{*}{DPOK~\cite{fan2023dpok}} & MS-COCO ImageReward & 0.22 & 0.55 & +0.33 \\
& & DrawBench ImageReward & 0.13 & 0.58 & +0.45 \\
\cmidrule(lr){2-6}
& \multirow{2}{*}{AlignProp \textcolor{black}{(v2)}~\cite{prabhudesai2023aligning}} & \textcolor{black}{Aesthetic (Animals test prompts)} & 5.64 & 8.71 & +3.07 \\
& & HPS-v2 reward (test) & 2.86 & 3.32 & +0.46 \\
\midrule
\multirow{2}{*}{Stable Diffusion v1.4}
& \multirow{2}{*}{DanceGRPO~\cite{xue2025dancegrpo}} & Pick-a-Pic & 0.202 & 0.215 & +0.013 \\
& & GenEval & 0.421 & 0.522 & +0.101 \\
\midrule
\multirow{4}{*}{FLUX.1-dev}
& \multirow{4}{*}{DanceGRPO~\cite{xue2025dancegrpo}} & HPS-v2.1 & 0.304 & 0.343 & +0.039 \\
& & CLIP & 0.405 & 0.427 & +0.022 \\
& & Pick-a-Pic & 0.224 & 0.228 & +0.004 \\
& & GenEval & 0.659 & 0.687 & +0.028 \\
\midrule
\multirow{3}{*}{Stable Diffusion 3.5-M}
& \multirow{3}{*}{Flow-GRPO~\cite{liu2025flow}} & GenEval & 0.63 & 0.95 & +0.32 \\
& & \textcolor{black}{OCR score} & 0.59 & 0.92 & +0.33 \\
& & PickScore & 21.72 & 23.31 & +1.59 \\
\midrule
\multirow{2}{*}{\textcolor{black}{Qwen-Image-2.0-Base}}
& \multirow{2}{*}{{Qwen-Image-2.0-RL~\cite{xu2026qwenimage20rltechnicalreport}}} & {Qwen-Image-Bench Overall} & 55.23 & {57.84} & {+2.61} \\
& & {T2I Arena Elo} & {1115} & {1193} & {+78} \\
\bottomrule
\end{tabular}
\vspace{-1em}
\end{table*}

Beyond perceptual coherence, recent works increasingly target \textit{\textbf{physical dynamics}} to \textcolor{black}{reduce physical inconsistencies in generated videos. PhysAR~\cite{Lin_2025_PhysAR} uses physics-aware rewards to optimize an autoregressive video-token policy with GRPO. PhysMaster~\cite{ji2025physmaster} uses staged Flow-DPO; its final stage freezes the DiT and updates only the physical head of PhysEncoder. PhysRVG~\cite{zhang2026physrvg} uses motion-mask and trajectory-based feedback to improve rigid-body motion generation. These methods improve physics-oriented video generation without guaranteeing general physical validity.} Please refer to Table~\ref{tab:rl4diff-video-methods} for a complete summary. And Table~\ref{tab:video-generation-quantitative} provides a quantitative comparison on \textbf{\textit{text-to-video generation}}, reporting paired results between each video generation baseline and its aligned counterpart under the metrics and protocols used in the original papers.

Geometry-aware rewards now extend video alignment from perceptual quality to 3D/4D world consistency.
World-R1~\cite{wang2026worldr1reinforcing3dconstraints} \textcolor{black}{uses feedback from 3D reconstruction and vision-language models to optimize a video generator with Flow-GRPO}, and periodic decoupled training balances geometric rigidity with non-rigid dynamics.
VGGRPO~\cite{an2026vggrpoworldconsistentvideogeneration} instead decodes 4D geometry directly from diffusion latents and rewards camera smoothness and reprojection consistency, thereby avoiding repeated RGB decoding.

Moreover, joint audio-video generation further expands credit assignment across modalities.
OmniNFT~\cite{zhang2026omninftmodalitywiseomnidiffusion} routes reward-specific advantages to \textcolor{black}{audio and video branches}, combines this routing with layer-wise gradient surgery, and reweights synchronization-critical regions. This replaces a global advantage with modality- \& region-aware feedback.

\vspace{1em} 
\noindent\textbf{RL in other modalities: 3D, audio, and motion.}
In addition to image and video modalities, RL also plays a crucial role in generative domains that deviate fundamentally from pixel space, including \textbf{\textit{3D geometry}}, \textbf{\textit{1D audio waveforms}}, and \textbf{\textit{embodied motion}}, where standard SFT objectives provide insufficient structural, logical, or physical guidance.

In \textbf{\textit{3D generation}}, the dominant challenge is \textbf{\textit{geometric consistency}}. Early methods relying on 2D priors typically suffer from the ``Janus problem'' due to the absence of 3D-aware supervision.
\textcolor{black}{DreamReward~\cite{ye2024dreamreward} combines Reward3D preference gradients with score distillation to optimize a NeRF, keeping the multi-view diffusion prior frozen.}
Nabla-R2D3~\cite{liu2025nablar2d3effectiveefficient3d} further addresses the complementary feedback bottleneck for 3D-native diffusion: it aligns 3D generators with mature 2D rewards via reward-gradient and score matching, reducing the need for specialized 3D preference data.

Moreover, in \textbf{\textit{audio generation}}, \textcolor{black}{alignment balances semantic correspondence, perceptual quality, and long-range structure.} Tango 2~\cite{majumder2024tango2aligningdiffusionbased} applies DPO~\cite{rafailov2023direct} on Audio-alpaca, a CLAP-filtered preference dataset, to improve semantic alignment in text-to-audio tasks. And for \textit{\textbf{music generation}}, \textcolor{black}{LeVo~\cite{lei2025levohighqualitysonggeneration} separately finetunes its autoregressive LeLM with DPO~\cite{rafailov2023direct} for lyric intelligibility, prompt consistency, and musicality, then merges the parameters; its diffusion-based music codec handles audio reconstruction rather than these preference updates.}
Resonate~\cite{li2026resonatereinforcingtexttoaudiogeneration} advances this line from offline pairwise preference optimization to online GRPO for a flow-based text-to-audio model. Specifically, its large-audio-language-model reward supplies finer temporal and compositional feedback than CLAP-style similarity scores, allowing exploration to improve both audio quality and semantic alignment.

Finally, \textbf{\textit{motion generation}} connects DMs to embodied intelligence, where the central requirement is \textbf{\textit{physical plausibility}} under textual control. Although DMs can produce diverse kinematic trajectories, they often violate physical constraints in simulation.
ReinDiffuse~\cite{han2024reindiffusecraftingphysicallyplausible} reformulates motion diffusion as a parameterized action distribution and applies RL with physics-based rewards to \textcolor{black}{reduce targeted motion artifacts. As a related autoregressive approach, MotionRL~\cite{liu2024motionrlaligntexttomotiongeneration} finetunes a text-to-motion generator with rewards for text adherence, motion quality, and human preference, via Pareto-based selection to balance these objectives.}


\subsubsection{RL-driven Image Editing}
\label{sec:rl_image_editing}
Not just multimodal AI generation, RL has also emerged as a powerful technique for enhancing image editing. Specifically, InstructRL4Pix~\cite{li2024instructrl4pix} \textcolor{black}{uses PPO with attention-map and CLIP-based rewards to fine-tune an instruction-based diffusion editor}.
To alleviate the training instability of online RL methods, EditScore~\cite{luo2025editscore} establishes a high-fidelity reward modeling paradigm for better alignment with user intentions. Furthermore, SPIE~\cite{benarous2025spie} integrates AI feedback to enhance the models' ability to preserve image structure while performing semantically meaningful edits.

\textcolor{black}{Recent editing work targets two complementary challenges: multi-turn consistency and editing-reward availability.}
MT-EditFlow~\cite{huang2026mteditflowreinforcementlearningmultiturn} treats successive user revisions as a multi-turn trajectory and designs turn-aware reward aggregation and advantage fusion to reduce error propagation.
Lever-Edit~\cite{ma2026performonlinerlimage} addresses scarce editing-specific rewards via counterfactual target captions that transfer established T2I rewards to online editing RL.

\begin{table*}[t]
\centering
\caption{Paired quantitative effects of representative RL/RL-family video generation methods. All values are collected from the reported tables in the original papers: InstructVideo's Table 1~\cite{Yuan_2024_CVPR}, VideoDPO's Table 1~\cite{liu2025videodpo}, T2V-Turbo's Table 1~\cite{Li_2024_NeurIPS}, DanceGRPO's Table 5~\cite{xue2025dancegrpo}, and DenseDPO's Table 1~\cite{wu2025densedpo}, together with VGGRPO's Table 1~\cite{an2026vggrpoworldconsistentvideogeneration}. VADER results are taken from the same VBench comparison protocol reported in VideoDPO's Table 1; DenseDPO uses the SFT baseline in its source table as the paired base. VBench-T/Q/S denote VBench Total, Quality, and Semantics; VQ/MQ/TA denote Visual Quality, Motion Quality, and Text Alignment. \textcolor{black}{Original metric scales are retained; comparisons are within paired rows, not across papers. T2V-Turbo-VC2 uses a four-step VCM-VC2 baseline, not the full-step teacher.}}
\vspace{-0.7em}
\label{tab:video-generation-quantitative}
\footnotesize
\setlength{\tabcolsep}{5pt}
\begin{tabular}{lllccc}
\toprule
\textbf{Base Models} & \textbf{Methods} & \textbf{Metrics} & \textbf{Base} & \textbf{Aligned} & \textbf{$\Delta$} \\
\midrule
\multirow{3}{*}{VideoCrafter2}
& \multirow{3}{*}{VADER~\cite{Prabhudesai_2024_VADER}} & VBench-T & 80.44 & 80.59 & +0.15 \\
& & VBench-Q & 82.20 & 82.46 & +0.26 \\
& & VBench-S & 73.42 & 73.09 & -0.33 \\
\midrule
\multirow{3}{*}{ModelScopeT2V}
& \multirow{3}{*}{InstructVideo~\cite{Yuan_2024_CVPR}} & HPS-v2 reward (in-domain) & 0.2506 & 0.2717 & +0.0211 \\
& & HPS-v2 reward (new animals) & 0.2502 & 0.2645 & +0.0143 \\
& & HPS-v2 reward (non-animals) & 0.2557 & 0.2682 & +0.0125 \\
\midrule
\multirow{3}{*}{VideoCrafter2}
& \multirow{3}{*}{VideoDPO~\cite{liu2025videodpo}} & VBench-T & 80.44 & 81.93 & +1.49 \\
& & VBench-Q & 82.20 & 83.07 & +0.87 \\
& & VBench-S & 73.42 & 77.38 & +3.96 \\
\midrule
\multirow{3}{*}{T2V-Turbo}
& \multirow{3}{*}{VideoDPO~\cite{liu2025videodpo}} & VBench-T & 80.95 & 81.80 & +0.85 \\
& & VBench-Q & 82.71 & 83.80 & +1.09 \\
& & VBench-S & 73.93 & 73.81 & -0.12 \\
\midrule
\multirow{3}{*}{VCM-VC2}
& \multirow{3}{*}{T2V-Turbo-VC2~\cite{Li_2024_NeurIPS}} & VBench-T & 73.97 & 81.01 & +7.04 \\
& & VBench-Q & 78.54 & 82.57 & +4.03 \\
& & VBench-S & 55.66 & 74.76 & +19.10 \\
\midrule
\multirow{4}{*}{HunyuanVideo}
& \multirow{4}{*}{DanceGRPO~\cite{xue2025dancegrpo}} & VQ & 4.51 & 7.03 & +2.52 \\
& & MQ & 1.37 & 3.85 & +2.48 \\
& & TA & 1.75 & 1.59 & -0.16 \\
& & VisionReward & 0.124 & 0.128 & +0.004 \\
\midrule
\multirow{3}{*}{SFT latent flow model}
& \multirow{3}{*}{DenseDPO~\cite{wu2025densedpo}} & VBench Dynamic Degree & 83.25 & 85.38 & +2.13 \\
& & VisionReward Text Alignment & 0.773 & 0.863 & +0.090 \\
& & VisionReward Visual Quality & 0.205 & 0.376 & +0.171 \\
\midrule
\multirow{2}{*}{{Wan2.2-5B}}
& \multirow{2}{*}{{VGGRPO~\cite{an2026vggrpoworldconsistentvideogeneration}}} & {VBench Imaging Quality} & {0.6402} & {0.6843} & {+0.0441} \\
& & {VBench Motion Smoothness} & {0.9467} & {0.9619} & {+0.0152} \\
\bottomrule
\end{tabular}
\vspace{-1em}
\end{table*}

\subsubsection{RL in Unified Understanding and Generation}
\label{sec:unification}
The unification of understanding and generation abilities within a single framework represents a significant advancement in multimodal AI. MMaDA~\cite{yang2025mmada} exemplifies this via a modality-agnostic diffusion architecture, and the proposed UniGRPO enables simultaneous optimization of comprehension and creation objectives, demonstrating RL's role in harmonizing diverse tasks within a unified system.
UniRL-Zero~\cite{wang2025unirl}, a robust foundation for advancing unified RL frameworks, further enhances the performance of complex generative tasks requiring tight LM-DM synergy.
GvU~\cite{pan2026learning}, an understanding-driven intrinsic rewarding framework for unified multimodal models, leverages internal comprehension signals to improve semantic alignment and instruction-following generation.
Moreover, AlphaGRPO~\cite{huang2026alphagrpounlockingselfreflectivemultimodal} closes the loop between understanding and generation in a unified AR-diffusion multimodal model. \textcolor{black}{Its decompositional verifiable reward evaluates the final generated image through atomic questions derived from the prompt; a shared advantage jointly optimizes the autoregressive and flow-based branches for reasoning-conditioned generation and self-reflective refinement.}

\textcolor{black}{Complementarily, UnifiedReward~\cite{wang2025unified} uses supervised learning for multimodal reward assessment, supporting downstream DPO and GRPO. Alpamayo-r1~\cite{wang2025alpamayo} combines RL-trained autoregressive reasoning and discrete trajectory prediction with a separately flow-matching-trained action expert. UnifiedReward-Think~\cite{wang2025unifiedthink} and R1-Reward~\cite{zhang2025r1} use GRPO and StableReinforce, respectively, to train multimodal reward models: RL updates the evaluator, not a diffusion generator.}

\section{RL for Diffusion Robotics}
\label{sec:dRobo}
RL has also demonstrated extremely powerful capabilities in the field of robotics. By enabling agents to interact with the environment and assisting with optimization through reward functions, RL has greatly promoted the development of robotics. Recently, generative models represented by DMs have also achieved remarkably impressive performance in robotics. They excel at multimodal action distribution modeling and high-dimensional action space representation, and such powerful capabilities can serve as strong priors to benefit RL. Therefore, a surge of research efforts has emerged in the community to integrate this RL-diffusion paradigm, thereby further advancing the field of robotics.

For \textbf{\textit{dRobotics}}, the trajectory-level optimization extends from visual or textual denoising paths to \textbf{\textit{continuous action-sequence refinement}}, where rewards are grounded in \textbf{\textit{closed-loop physical feedback}} (\textit{e.g.}, safety), further highlighting the importance of reward provenance, reliable verifiers, and safeguards against proxy reward over-optimizing.

\begin{figure}[t]
\centering
\includegraphics[width=0.48\textwidth]{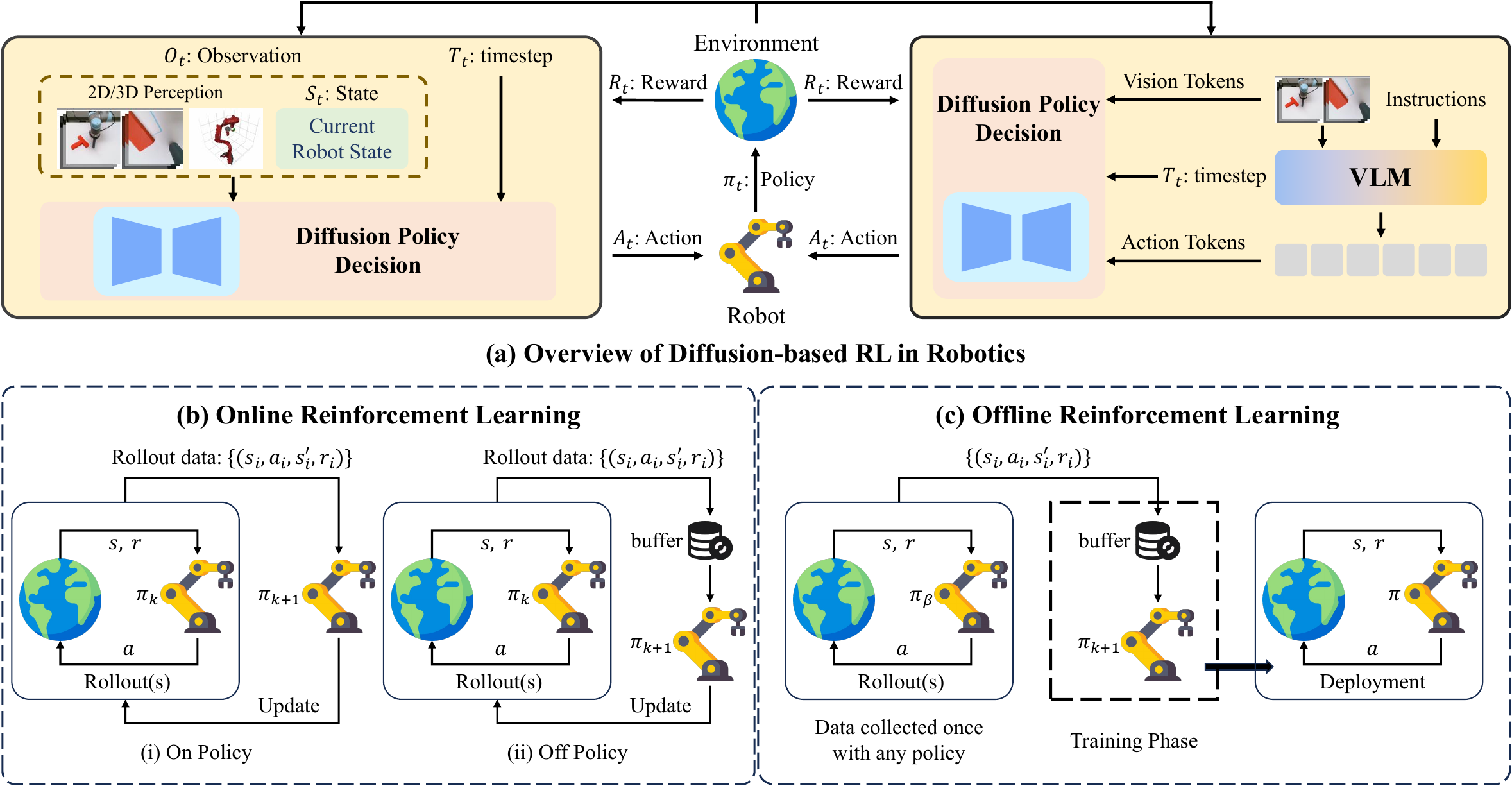}
\vspace{-0.25em}
\caption{
Overview of RL-driven Diffusion-based Robotics.
}
\vspace{-1.25em}
\label{fig:d_robo}
\end{figure}

\begin{figure}[t]
\centering
\includegraphics[width=0.48\textwidth]{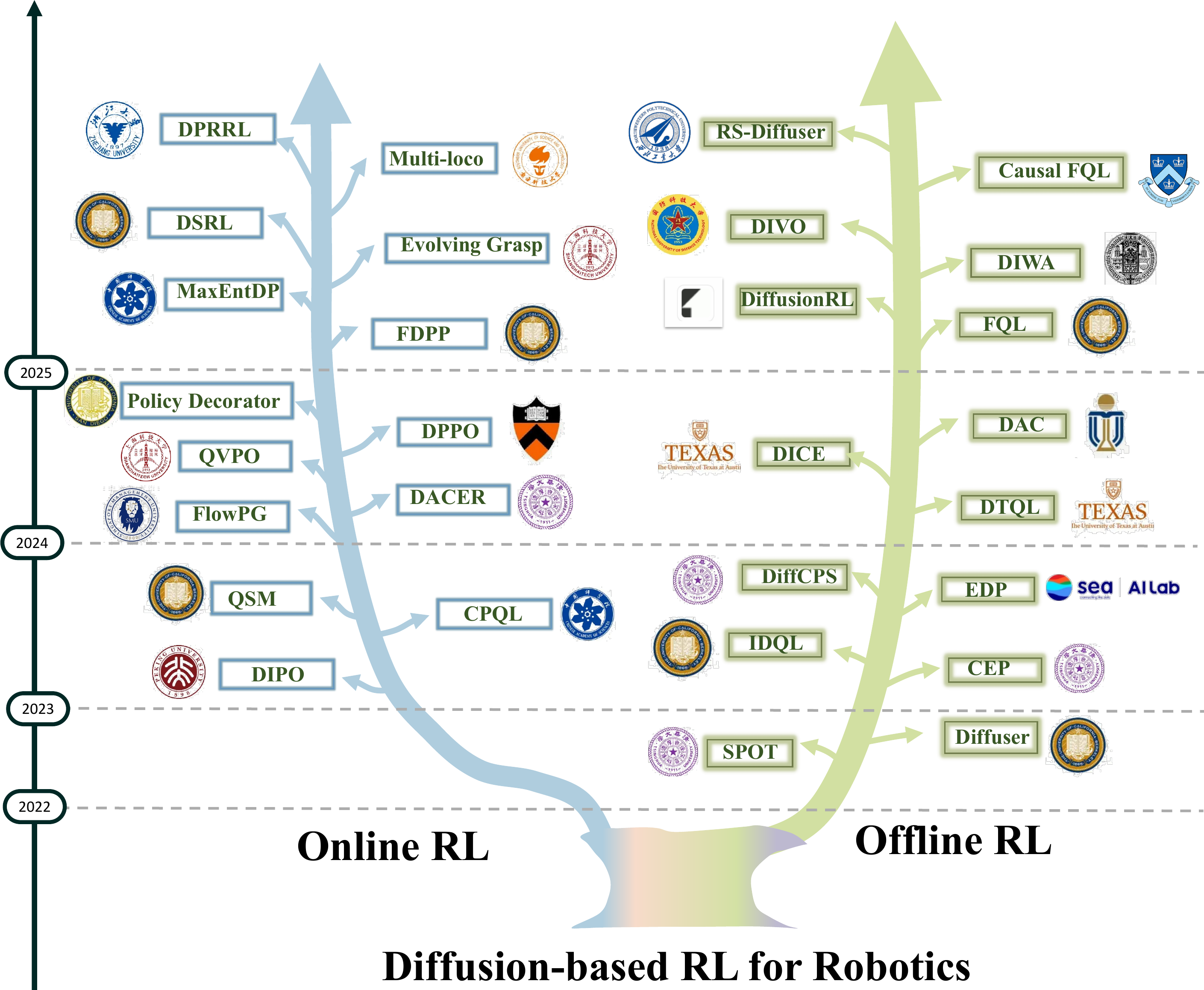}
\vspace{-0.75em}
\caption{
Representative research about diffusion-based RL for robotics.
}
\vspace{-1.25em}
\label{fig:paper_robo}
\end{figure}

There exist two mainstream integration approaches in embodied intelligence research: \textbf{\textit{1) Diffusion Policy-based RL training:}} DMs as policy backbones iteratively optimized via RL gradients; \textit{\textbf{2) Generative experts in Vision-Language-Action models:}} fusing multimodal perception with actions generated by diffusion (Fig.~\ref{fig:d_robo}).
Moreover, the first integration can be structured by two RL paradigms for distinct scenarios: \textbf{\textit{online diffusion RL}} and \textit{\textbf{offline diffusion RL}}. This section comprehensively sorts out recent RL literature in diffusion-based robotics along with 
\emph{online/offline} scenarios (Fig.~\ref{fig:paper_robo}) and systematically reviews theoretical foundations, integration frameworks, applications, and core challenges, aiming to clarify how this synergy advances robots towards adaptive, robust, human-level embodied intelligence.

\subsection{Foundations of RL for Diffusion in Robotics}
\label{sec:pre_RL4Robo}
\noindent\textbf{Diffusion Policy via DDPMs.}
\label{sec:Diffusion Policy Preliminaries}
Diffusion Policy (DP)~\cite{Chi2023DiffusionPV} first attempts to explore the diffusion-based policy $\pi_\theta$ parameterized by DDPMs~\cite{ho2020DDPM}, which takes in $s$ as a conditioning argument and then parameterizes $p_\theta(a^{k-1}|a^k, s)$ as $\mathcal{N}\left(a^{k-1}; \sqrt{\alpha_k} a^k + \sqrt{1 - \alpha_k} \varepsilon_\theta(a^k, s, k), \sigma_k^2 \mathbf{I}\right)$.
DPs can be trained via behavior cloning by fitting the conditional noise prediction $\varepsilon_\theta(a^k, s, k)$ to predict the added noise. Note that unlike standard policy parameterizations, \textit{e.g.}, unimodal Gaussian policies, DPs do not maintain an explicit likelihood of $p_\theta(a^0|s)$. Similar with image or language generation, DP's training also exploits the DDPM to approximate the conditional distribution $p(\mathbf{A}_t | \mathbf{O}_t)$:
\begin{equation}
    \label{eq:diffusion_policy_langevin}
    \mathbf{A}^{k-1}_t = \alpha(\mathbf{A}^k_t - \gamma\epsilon_\theta(\mathbf{O}_t,\mathbf{A}^k_t,k) + \mathcal{N} \bigl(0, \sigma^2 I \bigl)),
\end{equation}
where $\mathbf{A}_t$ is the robot action and $\mathbf{O}_t$ represents the input observation, while the training loss can be expressed as
\begin{equation}
    \label{eq:diffusion_policy_loss}
    \mathcal{L}=\text{MSE}(\mathbf{\epsilon}^k,\epsilon_\theta(\mathbf{O}_t, \mathbf{A}^0_t + \mathbf{\epsilon}^k, k)).
\end{equation}

\noindent\textbf{Diffusion VLA via Flow Matching.}
VLA models typically adopt DMs (particularly flow matching) as the action expert, which formulates action generation as conditional denoising. Diffusion-based methods naturally model diverse action distributions, enabling multiple valid trajectories from identical observations. In general, for each action $a_{t'}$ in the action chunk $ A_t$, a corresponding \emph{action token} will be fed into the action expert. During training, these action tokens are supervised using a conditional flow matching loss~\citep{lipman2022flow},
\begin{align*}
    L^\tau(\theta) &= E_{p(A_t | o_t), q(A_t^\tau | A_t)} || v_\theta(A_t^\tau, o_t) - u(A_t^\tau | A_t) ||^2,
\end{align*}
where subscripts denote robot timesteps and superscripts denote flow matching timesteps, with $\tau \in [0, 1]$.

\subsubsection{RL-driven Diffusion-based Robotics}
\label{sec:Diffusion Models based RL for robotics}
\noindent\textbf{Basics of RL in Robotics}. In robotic systems, RL empowers robots to learn task execution via environmental interaction and feedback, with its core concepts forming the learning loop.
Briefly: \textbf{\textit{Agent (Robot)}} perceives the environment and executes actions; \textbf{\textit{Environment}} is the robot's physical operating space, with states evolving via transition functions in response to robot's actions; \textbf{\textit{State}} describes the agent-environment status (\textit{e.g.}, position, sensor data) at a given time; \textbf{\textit{Action}} denotes executable operations (\textit{e.g.}, joint control); \textbf{\textit{Reward}} is scalar feedback guiding the agent to distinguish favorable behaviors; \textbf{\textit{Policy}} is the agent's decision-making rule mapping states to actions; and \textbf{\textit{Value function}} quantifies long-term reward potential (including state-value and action-value/Q-functions), directing optimal action selection for maximum cumulative rewards.

RL in robotics attempts to learn the optimal \emph{``perception-decision-action''} strategy via physical interaction. However, traditional RL policies (\textit{\textit{e.g.}}, Gaussian, MLP-parameterized policies) are constrained by predefined distribution forms, struggling to handle high-dimensional, multi-modal action spaces (\textit{\textit{e.g.}}, robotic manipulation, trajectory planning). By contrast, DMs capture complicated data distributions via the ``forward diffusion (noising) \& reverse denoising'' process, and DMs' strong conditional generation capability provides a novel policy representation paradigm for RL. Specifically, the combination requires no prior assumptions about action distributions but directly models state-action mappings via state-conditioned denoising. We theoretically systematize the core framework, mathematical formulations, and modular design logic of diffusion-based RL below.

\begin{table*}[t!]
    \centering
    \caption{Representative offline RL methods on the D4RL Gym-MuJoCo and AntMaze benchmarks, adapted from DIVO~\cite{ma2025diffusion}. All tasks use standardized D4RL v2 normalized scores, where higher is better. Gym-MuJoCo includes dense-reward locomotion tasks: HalfCheetah for planar running, Hopper for single-legged hopping and balance, and Walker2d for bipedal walking control; each is evaluated with medium, medium-replay, and medium-expert datasets. AntMaze includes sparse-reward long-horizon navigation tasks in U-Maze, Medium, and Large layouts, with play datasets collected from exploratory behavior and diverse datasets containing more varied goal-reaching trajectories. DIVO results are reported as mean $\pm$ standard error over the last 10 evaluations and 5 random seeds, and the aggregate rows sum normalized scores within each benchmark group.}
    \label{experi:d4rl-result}
    \vspace{-1em}
    \resizebox{\linewidth}{!}{
    \small
    \setlength{\tabcolsep}{4pt}
    \renewcommand{\arraystretch}{1.1}  
    \begin{tabular}{lccccccccc}
    \hline
    \toprule
    \multirow{2}{*}{\bf Task Name}  
    & \multicolumn{4}{c}{\bf Non-diffusion Methods} 
    & \multicolumn{5}{c}{\bf Diffusion-based Methods} \\
    \cmidrule(lr){2-5} \cmidrule(lr){6-10}
    & \bf TD3+BC~\cite{fujimoto2021minimalist} 
    & \bf BCQ~\cite{fujimoto2019off} 
    & \bf BEAR~\cite{kumar2019stabilizing} 
    & \bf CQL~\cite{kumar2020conservative} 
    & \bf EDP~\cite{kang2023efficient} 
    & \bf IDQL~\cite{hansen2023idql} 
    & \bf Diffusion-QL~\cite{wang2022diffusion} 
    & \bf DTQL~\cite{chen2024diffusion} 
    & \bf DIVO~\cite{ma2025diffusion} \\ 
    \midrule
        halfcheetah-medium & 48.3 & 47.0 & 41.0  & 44.0 & 52.1  & 51.0 & 51.1 (0.5) & 57.9 (0.13) & \cellcolor{green!16}\textbf{67.71} (0.64)\\ 
        hopper-medium & 59.3 & 56.7 & 51.9 & 58.5 & 81.9 & 65.4 & 90.5 (4.6) & 99.6 (0.87) & \cellcolor{green!16}\textbf{100.34} (0.44)  \\
        walker2d-medium & 83.7 & 72.6 & 80.9 & 72.5 & 86.9 & 82.5 & 87.0 (0.9) & 89.4 (0.13) & \cellcolor{green!16}\textbf{90.94} (1.63) \\ 
        \midrule
        halfcheetah-medium-replay & 44.6 & 40.4 & 29.7 & 45.5 & 49.4 & 45.9 & 47.8 (0.33) & 50.9 (0.11) & \cellcolor{green!16}\textbf{55.53} (0.77)\\ 
        hopper-medium-replay & 60.9 & 53.3 & 37.3 & 95.0 & 101.1 & 92.1 & \cellcolor{green!16}\textbf{101.3} (0.6) & 100.0 (0.13) & \cellcolor{green!16}\textbf{101.28} (0.40)\\ 
        walker2d-medium-replay & 81.8 & 52.1 & 18.5 & 77.2 & 94.9 & 85.1 & \cellcolor{green!16}\textbf{95.5} (1.5) & 88.5 (2.16) & 94.85 (0.57)\\ 
        \midrule
        halfcheetah-medium-expert & 90.7 &  89.1  &  38.9   & 91.6 & 95.5  &95.9& 96.8 (0.33) & 92.7 (0.2) & \cellcolor{green!16}\textbf{97.88} (0.74)\\ 
        hopper-medium-expert & 98.0 & 81.8 & 17.7   & 105.4 & 97.4 &108.6 & \cellcolor{green!16}\textbf{111.1} (1.3) & 109.3 (1.49) & \cellcolor{green!16}\textbf{111.02} (1.64) \\ 
        walker2d-medium-expert & 110.1 & 109.5 & 95.4  & 108.8 & 110.2 &\cellcolor{green!16}\textbf{112.7} & 110.1 (0.33) & 110.0 (0.07) & \cellcolor{green!16}\textbf{112.03} (0.62) \\ 
        \midrule
        {\bf{Gym Aggregate}} & 677.4 & 602.5 & 411.3 & 698.5 & 769.5 & 739.2& 792.0 & \cellcolor{gray!20}\underline{798.3} &\cellcolor{green!16}\textbf{831.6} \\ 
        \midrule
        antmaze-umaze & 91.3 & 0.0 & 73.0  & 84.8 & 96.6 &94.0& 93.4 (3.4) & 94.8 (1.00) & \cellcolor{green!16}\textbf{100.00} (0.00) \\ 
        antmaze-umaze-diverse & 54.6 & 61.0 & 61.0 & 43.3 & 69.5 &80.2& 66.2 (8.6) & 78.8 (1.83) & \cellcolor{green!16}\textbf{92.00} (4.47)\\ 
        antmaze-medium-play & 0.0 & 0.0 & 0.0  & 65.2 & 0.0 &84.5& 76.6 (10.8) & 79.6 (1.8) & \cellcolor{green!16}\textbf{86.70} (4.31) \\ 
        antmaze-medium-diverse & 0.0 & 0.0 & 8.0 & 54.0 & 6.4 &84.8& 78.6 (10.3) & 82.8 (1.71) & \cellcolor{green!16}\textbf{86.70} (12.47)\\ 
        antmaze-large-play & 0.0 & 6.7 & 0.0 & 18.8 & 1.6 &63.5& 46.6 (8.3) & 52.0 (2.23) & \cellcolor{green!16}\textbf{58.00} (8.37)\\ 
        antmaze-large-diverse & 0.0 & 2.2 & 0.0 & 31.6 & 4.4 &\cellcolor{green!16}\textbf{67.9} & 56.6 (7.6) & 54.0 (2.23) & 63.33 (4.71) \\ 
        \midrule
        {\bf{AntMaze Aggregate}} & 145.9 & 69.9 & 142.0 & 297.7 & 178.8 & \cellcolor{gray!20}\underline{474.6} &417.6 &441.6 &\cellcolor{green!16}\textbf{486.7} \\ 
        \midrule
        {\bf{Total Aggregate}} & 823.3 & 672.4 & 553.3 & 996.2 & 948.3 & 1213.8 &1209.6 & \cellcolor{gray!20}\underline{1239.9} &\cellcolor{green!16}\textbf{1318.3} \\ 
        \bottomrule
      \hline
    \end{tabular}}
    \vspace{-1em}
\end{table*}

DMs' forward process is a parameter-free Markov chain: at timestep $k$, Gaussian noise is added to the original action $a_t^0$, yielding the noisy action $a_t^k$. And the reverse process uses a neural network to learn the mapping from $a_t^k$ back to $a_t^0$, conditioned on the environmental state $s_t$—forming the core representation of RL policy as follows:
\begin{equation}
\begin{medsize}
a_{t}^{k-1} \sim p_{\theta}\left(a_{t}^{k-1} \mid a_{t}^{k}, s_{t}\right) = \mathcal{N}\left( a_{t}^{k-1} ; \mu_k\left(a_{t}^{k}, \varepsilon_{\theta}(a_{t}^{k}, s_{t}, k)\right), \sigma_k^2 \mathbf{I} \right),
\label{eq:diffusion_rl_policy}
\end{medsize}
\end{equation}
where $a_t^k$ is the action after $k$ noising steps in the $t$-th environmental interaction ($k=K$ corresponds to pure Gaussian noise and $k=0$ denotes the final output action); $\varepsilon_{\theta}(\cdot)$ is the state-conditioned noise prediction network, taking $a_t^k$, $s_t$, and denoising step $k$ as inputs to estimate the true noise added during the forward process; and $\mu_k(\cdot)$ denotes the fixed mean function, which can be calculated as
\begin{equation}
\begin{medsize}
\mu_k(a_t^k, \varepsilon) = \frac{\sqrt{\alpha_k}(1-\bar{\alpha}_{k-1})}{1-\bar{\alpha}_k}a_t^k + \frac{\sqrt{\bar{\alpha}_{k-1}}(1-\alpha_k)}{1-\bar{\alpha}_k}(a_t^k - \sqrt{1-\bar{\alpha}_k}\varepsilon),
\label{eq:diffusion_mean_func}
\end{medsize}
\end{equation}
where $\alpha_k$ is a predefined coefficient and $\bar{\alpha}_k = \prod_{i=1}^k \alpha_i$. This definition avoids prior assumptions about action distributions, fitting any continuous action distribution via iterative denoising, making it ideal for multi-modal action scenarios.

\begin{figure}[t]
    \centering
    \includegraphics[width=\linewidth]{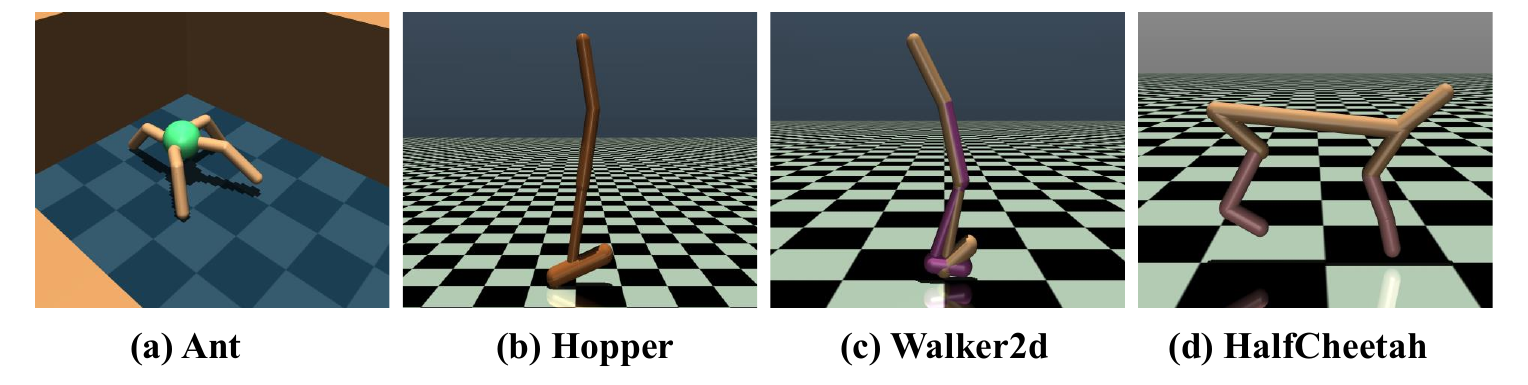}
    \vspace{-2.25em}
    \caption{The visualization of four offline RL benchmarks for robot continuous control, including Ant, Hopper, Walker2d, and HalfCheetah.}
    \vspace{-1.8em}
    \label{fig:simu}
\end{figure}

\subsection{Diffusion Policy-based RL Training}
\label{sec:app_RLinDP}
\noindent\textbf{Online Diffusion RL.}
Online RL strives to learn the optimal policy via continuous interactions with the target environment and the integration of online RL with DMs has become a prominent direction in robotic learning.

Existing research on online diffusion RL can be broadly categorized into two paradigms. The first paradigm treats generative policies (\textit{e.g.}, DMs) as black-box function approximators and optimizes them using policy gradient methods. Representative works include DACER~\cite{wang2024diffusion}, CPQL~\cite{chen2023boosting}, and FlowPG~\cite{brahmanage2023flowpg}, which directly utilize deterministic or stochastic policy gradient updates to generative model policies. Moreover, the second paradigm leverages the inherent structural properties and denoising mechanisms of generative models to drive policy optimization: for example, DIPO~\cite{yang2023policy} uses the Q-function's gradient and value magnitude to identify the optimal action, then incorporates them into DMs' variational loss for training; QSM~\cite{psenka2023learning} and MaxEntDP~\cite{dong2025maximum} further treat DMs' noise network as the score function of the target policy distribution, employing the Q-function's gradient and variants to update the network.

As a complementary flow-based approach, VINE~\cite{yang2026vine} redesigns the action-sampling path to stabilize end-to-end value-gradient optimization of generative control policies, with evaluations in both offline RL and real-world online robot learning. DPPO~\cite{ren2024diffusion} fine-tunes behavior-cloned diffusion policies with PPO using newly collected environment rollouts, enabling on-policy optimization over denoising trajectories. Building on PPO-based diffusion policy optimization, D3P~\cite{Yu2025D3PDD} jointly optimizes a diffusion policy and an adaptive denoising-step controller with online rollouts, allocating computation according to action difficulty. Beyond the choice of policy-update objective, ADPO~\cite{jiang2025adaptive} applies adaptive gradient optimization to RL fine-tuning of diffusion policies during environmental interaction. For hybrid discrete--continuous action spaces, HyDo~\cite{le2025enhancing} combines a diffusion policy for continuous motion parameters with Q-value-based discrete contact-point selection in an online off-policy maximum-entropy RL framework, enabling exploration over both action components.

Overall, current methods mostly focus on integrating generative models with off-policy RL algorithms, emphasizing adaptive policy refinement via real-time environmental feedback.
Multi-Loco~\cite{yang2025multi} proposes a shared residual RL policy to refine action outputs of pretrained DMs via online optimization, significantly narrowing the sim2real gap in online robotic control tasks. Beyond environmental feedback-driven refinement, preference-based optimization has also emerged as a key direction: FDPP~\cite{chen2025fdpp} first learns a reward function via preference-based learning, then uses RL to fine-tune pretrained policies, ensuring the policy aligns with human preference while preserving its original ability.
EvolvingGrasp~\cite{zhu2025evolvinggrasp} introduces Handpose-wise Preference Optimization (HPO), formulating grasp adaptation as a posterior probability optimization problem, and converges iteratively toward human-preferred grasp distributions with online preference feedback, enhancing the real-world practicality of diffusion-based grasp policies.

Moreover, several works combine imitation-pretrained policies with RL-based refinement.
Policy Decorator~\cite{yuan2024policy} optimizes and refines offline IL models via online environmental interactions using a model-agnostic residual policy, ensuring IL-policy stability while reducing erratic RL behaviors.
DSRL~\cite{wagenmaker2025steering} leverages behavior cloning for rapid adaptation of pretrained policies through efficient online RL, enabling quick deployment in real-world scenarios.
And DP-RRL~\cite{11114901} designs a diffusion-based IL policy learned from expert demonstration data, paired with a residual RL module using real-time environmental feedback for complex dynamics adaptation.

\begin{table*}[t]
\setlength{\tabcolsep}{1pt}
\centering
\caption{Diffusion-/Flow-based VLA methods with RL or RL-relevant fine-tuning signals.}
\vspace{-1em}
\label{tab:vla_rl_aggregation}
\renewcommand{\arraystretch}{1.1}
\scalebox{0.64}{\begin{tabular}{cccccc}
\toprule
\textbf{Methods} & \textbf{Date} & \textbf{Backbone} & \textbf{Action Expert} & \textbf{Key Contributions / RL Signal} & \textbf{Primary Setting} \\
\midrule
\multicolumn{6}{c}{\cellcolor{lightblue}\textbf{\textit{Foundation VLA and Generative Action Experts}}}\\
\midrule
RDT-1B~\cite{liu2024rdt} & 2024.10 & Diffusion Transformer & MLP action head & Scalable diffusion foundation model for bimanual manipulation & Bimanual manipulation  \\
$\pi_0$~\cite{black2024pi0visionlanguageactionflowmodel} & 2024.10 & VLM / LLM & Flow Matching & Continuous action generation conditioned on VLM features & General robot control  \\
AgiBot World~\cite{agibotworldcontributors2025agibotworldcolosseolargescale} & 2025.03 & VLM / LLM & Diffusion-based  & Large-scale embodied data and generalist manipulation policy & Large-scale manipulation  \\
$\pi_{0.5}$~\cite{intelligence2025pi05visionlanguageactionmodelopenworld} & 2025.04 & VLM / LLM & Flow Matching & Co-training for open-world VLA generalization & Open-world robot control  \\
\midrule
\multicolumn{6}{c}{\cellcolor{lightblue}\textbf{\textit{RL, RFT, and World-Model-Based VLA Optimization}}}\\
\midrule
Beyond~\cite{yang2025humandemonstrationsdiffusionbasedreinforcement} & 2025.09 & Diffusion Policy & Diffusion action policy & Diffusion RL generates high-quality VLA training trajectories & VLA data generation  \\
VLA-RFT~\cite{li2025vlarftvisionlanguageactionreinforcementfinetuning} & 2025.10 & VLM / LLM & Flow Matching & World-simulator RFT with verified trajectory rewards & Robot policy refinement  \\
VDRive~\cite{guo2025vdriveleveragingreinforcedvla} & 2025.10 & VLM / LLM & Diffusion Policy & Actor-critic feedback for reinforced VLA driving decisions & Autonomous driving  \\
$\pi_{\mathrm{RL}}$~\cite{chen2025pitextttrlonlinerlfinetuning} & 2025.10 & VLM / LLM & Flow Matching & Online PPO with Flow-Noise / Flow-SDE likelihood modeling & Online VLA RL  \\
WMPO~\cite{zhu2026wmpo} & 2025.11 & VLM / LLM & MLP action head & On-policy GRPO with pixel-based world-model feedback & World-model VLA RL  \\

{SA-VLA~\cite{pan2026sa}} & {2026.02} & {VLM / LLM} & {Flow Matching} & {Spatially-aware flow matching preserving spatial inductive bias} & {Spatial-aware VLA RL}  \\
{World-VLA-Loop~\cite{liu2026world}} & {2026.02} & {VLM / LLM} & {MLP action head } & {Closed-loop co-evolution of video world model and VLA policy} & {World-model VLA RL}  \\
{OmniVLA-RL~\cite{jie2026omnivla}} & {2026.04} & {VLM / LLM} & {Flow Matching} & {SDE with group segmented policy optimization} & {Online VLA RL}  \\
{TMRL~\cite{hong2026tmrl}} & {2026.05} & {VLM / LLM} & {Flow Matching} & {Timestep-modulated pretraining enabling exploration} & {Robot policy finetuning}  \\
{Q-VGM~\cite{wang2026q}} & {2026.06} & {VLM / LLM} & {Flow Matching} & {Q-value-gradient matching for offline-to-online RL} & {Offline-to-online VLA RL}  \\
{FlowPRO~\cite{wu2026flowpro}} & {2026.06} & {VLM / LLM} & {Flow Matching} & {Proximalized preference optimization avoiding reward hacking} & {Preference-based VLA RL}  \\
{SafeDojo~\cite{tang2026safedojo}} & {2026.06} & {VLM / LLM} & {MLP action head} & {Lagrangian-constrained GRPO for safe RL} & {World-model safe VLA RL}  \\
{dVLA-RL~\cite{wu2026dvla}} & {2026.06} & {Diffusion VLA} & {Discrete Diffusion} & {MDP over denoising trajectories with variable-step scheduling} & {DVLA RL}  \\
{RedFlow~\cite{yan2026redflow}} & {2026.07} & {VLM / LLM} & {Flow Matching} & {Redirect failure into action-level corrections} & {Offline VLA RL}  \\
{WCM~\cite{fei2026wcm}} & {2026.07} & {VLM / LLM} & {Flow / MLP} & {World critic model predicting future latents and value estimates} & {VLA RL}  \\
{StructRL~\cite{yang2026structrl}} & {2026.08} & {VLM / LLM} & {Flow Matching} & {Structured action-space exploration mitigating noise dilution} & {Online VLA RL}  \\
\bottomrule
\end{tabular}}
\vspace{-12pt}
\end{table*}

More recent studies further extend online diffusion RL to specialized and energy-constrained embodied settings.
Specifically, L-SDPPO~\cite{zhang2026sdppo} formulates the optimization of a \textit{spiking} diffusion policy via RL for intra-vehicular robotic manipulation, jointly handling the multimodal action distributions and the strict power budgets of spacecraft environments.
X-NavDP~\cite{yang2026x} targets data-efficient diffusion RL post-training for navigation, introducing group Q-score reweighted matching (GQRM) to generalize a single navigation diffusion policy across novel behaviors and embodiments.
Beyond action-space policies, CreFlow~\cite{ni2026creflow} introduces an online RL framework for embodied video diffusion models, combining a credit-aware normalized flow trajectory loss along with corrective reflow to enforce compositional task constraints under sparse rewards.

\noindent\textbf{Offline Diffusion RL.}
Compared with online diffusion RL, offline diffusion RL attracts more research attention, as it can avoid the high cost and safety risks of real-time environmental interactions by relying solely on pre-collected offline datasets.
The core challenge of offline RL is preventing policy drift beyond the support of offline datasets.
To address this, SPOT~\cite{wu2022supported} adopts a VAE-based policy regularizer to constrain policy optimization within the data distribution, improving the generalization of offline policies.

Another direction focuses on using diffusion or flow policies to model the policy distribution of offline datasets more accurately:
diffusion-QL~\cite{wang2022diffusion}, EDP~\cite{kang2023efficient}, SRDP~\cite{ada2023diffusion}, CEP~\cite{lu2023contrastive}, and FQL~\cite{park2025flow} are representative works, demonstrating that generative models can capture complex action distributions of offline datasets effectively, thereby yielding policies with superior generalization capability. Beyond these, DIPOLE~\cite{liang2026dichotomous} uses dichotomous policy regularization to diffusion-policy optimization on offline datasets and additionally evaluates offline-to-online adaptation.

Building on these foundational advances, recent works have enhanced the performance and applicability of offline diffusion RL.
Diffuser~\cite{janner2022planning} pioneers this research by training a DM on offline datasets for trajectory generation, using guided sampling to plan future trajectories.
This study first demonstrates DMs can unify the functionalities of policy models and dynamic models, laying a foundation not only for trajectory-level modeling in IL but also for subsequent offline diffusion RL research. DiffusionRL~\cite{makarova2025diffusionrl} takes a data-centric approach, using RL in simulation to refine grasp samples from a pre-existing large-scale dataset before training a diffusion policy on the resulting trajectories via offline supervised learning. DiWA~\cite{Chandra2025DiWADP} further pushes offline refinement by introducing a world model for full offline fine-tuning of diffusion-based robotic skills, eliminating reliance on real-time feedback. 

Beyond the standard return-maximizing objectives, risk-sensitive planning has emerged as a safety-oriented offline direction: RS-Diffuser~\cite{gong2026rs} couples diffusion trajectory generation along with a distributional value critic and injects tail-aware (\textit{e.g.}, CVaR) guidance into the denoising process, enabling a single policy to trade off risk-averse, risk-neutral, and risk-seeking behaviors during inference time.
In parallel, MODIP~\cite{asri2026modip} addresses offline-to-online adaptation by leveraging a learned world model and MPC to synthesize high-quality trajectories that supervise a diffusion policy, thereby retaining the simplicity of behavior cloning while outperforming pure RL fine-tuning. For offline flow-policy RL, Causal Flow Q-Learning~\cite{li2026causal} tackles observational confounding from a causal perspective to provide worst-case robust offline optimization for flow policies.

Fig.~\ref{fig:simu} visualizes four offline RL benchmarks for robot continuous control, and Table~\ref{experi:d4rl-result} compares the performance of non-diffusion-based and diffusion-based RL methods.

\subsection{Generative Experts in VLA Models}
\label{sec:RLindVLA}

In robotics, diffusion-based policy models hold promise for generative control, facilitating flexible trajectory generation via the modeling of action distributions. However, they face challenges in integrating visual-linguistic information—a core limitation that has spurred the development of Vision-Language-Action (VLA) models.
VLA models leverage pretraining from Vision-Language Models (VLMs) to advance practical learning-enabled robotics; specifically, they draw on large-scale VLM pretraining to make such robotic systems a reality. Nevertheless, the use of discrete action tokens impedes their ability to achieve fine-grained trajectory control. However, the design of discrete action tokens is not conducive to fine-grained control of robotic trajectories. $\pi_0$~\cite{black2024pi0visionlanguageactionflowmodel} first proposes an FM model as an action expert conditioned on the output of the VLM. Subsequently, numerous studies~\cite{liu2024rdt,agibotworldcontributors2025agibotworldcolosseolargescale,intelligence2025pi05visionlanguageactionmodelopenworld} have further explored the integration of DMs with VLA frameworks. Nevertheless, IL-based approaches often limit the scalability of such models, whereas RL has emerged as a potential alternative.

In robotic tasks, RL optimizes policies by iteratively maximizing cumulative rewards, thereby discovering strategies that surpass the upper bound imposed by merely imitating human demonstrations. Specifically, Yang et al.~\cite{yang2025humandemonstrationsdiffusionbasedreinforcement} employs task-completion-based rewards to guide the update of the policy network. VDRive~\cite{guo2025vdriveleveragingreinforcedvla} leverages RL to enhance interpretable and robust decision-making in dynamic autonomous driving environments. And $\pi_{rl}$~\cite{chen2025pitextttrlonlinerlfinetuning} introduces RL into \textbf{\textit{flow-based VLA models}} through noise injection and MDP modeling for online finetuning. dVLA-RL~\cite{wu2026dvla} further extends RL to discrete diffusion VLAs by reparameterizing the objective from the intractable marginal action probability to the joint probability of the sampled denoising path, formulated as an MDP with variable denoising-step scheduling.
StructRL~\cite{yang2026structrl} proposes structured action-space exploration for flow-based VLAs. Moreover, end-to-end value-gradient methods have further matured generative-policy RL: Q-VGM~\cite{wang2026q} matches Q-value gradients for offline-to-online RL of flow-based VLA policies. For VLA-specific RL fine-tuning, SA-VLA~\cite{pan2026sa} injects spatial awareness into flow-matching VLA RL, while OmniVLA-RL~\cite{jie2026omnivla} couples spatial understanding with online RL. FlowPRO~\cite{wu2026flowpro} achieves reward-free reinforced fine-tuning of flow-based VLAs via proximalized preference optimization; RedFlow~\cite{yan2026redflow} derives action-level correction targets from progress and outcome signals in a fixed buffer of rollouts for offline corrective post-training of flow-based VLA policies; and TMRL~\cite{hong2026tmrl} modulates pretraining by diffusion timesteps for efficient policy finetuning.

\textbf{\textit{World model-based VLA RL}} has also gradually gained attention.
Specifically, VLA-RFT~\cite{li2025vlarftvisionlanguageactionreinforcementfinetuning} utilizes world models as simulators to generate future states and performs end-to-end VLA updates using verifiability-based rewards. World-VLA-Loop~\cite{liu2026world} jointly learns a video world model and a VLA policy in a closed loop, while SafeDojo~\cite{tang2026safedojo} focuses on task-safety co-optimization via Lagrangian-constrained GRPO over interactive world-model rollouts. WCM~\cite{fei2026wcm} learns a world critic model for VLA RL, complementing rollout-based approaches with predictive value estimation.
Table~\ref{tab:vla_rl_aggregation} shows representative RL-driven dVLA methods.
\vspace{-10pt}

\section{RL for Diffusion Science Models}
\label{sec:DSci}

\begin{figure}[t]
\centering
\includegraphics[width=0.48\textwidth]{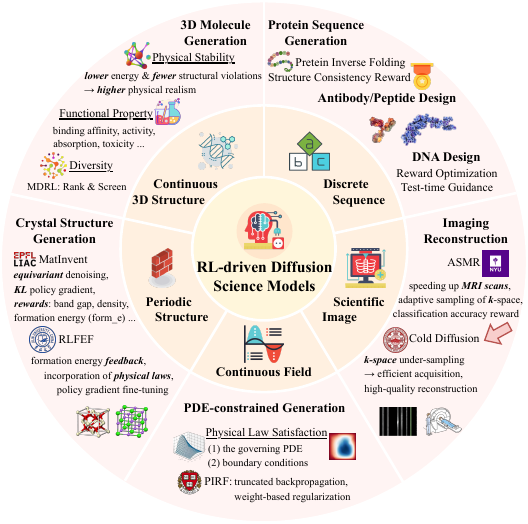}
\vspace{-4pt}
\caption{A summary of the overall five RL-driven scientific generation domains. Note that the representative methods listed here are not exhaustive; please refer to our main text for more details.
}
\vspace{-1.5em}
\label{fig:RL4dSci}
\end{figure}

Beyond the above domains, RL has also been introduced to scientific generative tasks for explicit control over physical laws or function-specific objectives through well-designed reward signals, thereby producing desirable outcomes.
In this section, we systematically review recent advances in RL-driven diffusion-based science models from a novel perspective of data modality, as shown in Fig.~\ref{fig:RL4dSci}.

\vspace{-1em}
\subsection{Discrete Molecular Sequence Generation}
\label{sec:Discrete Sequence Generation}
Discrete biomolecular sequences (\textit{e.g.}, proteins and DNA) are typically modeled by discrete diffusion frameworks (\textit{e.g.}, D3PM~\cite{austin2021structured}), yet these inherently lack explicit additional control over structural or functional objectives.
However, RL effectively bridges this gap by enabling reward-guided optimization toward desired biomolecular properties.
For structural design, RL-DIF~\cite{ektefaie2024reinforcement} leverages DDPO~\cite{black2024training} with sc-TM~\cite{trippe2022diffusion} to enhance protein foldability, whereas SDPO~\cite{han2025discrete} employs stepwise offline preference learning to align diffusion trajectories with structural targets.
In functional optimization, RL-D$^2$~\cite{ma2025reinforcement} trains diffusion policies via policy mirror descent (PMD) for efficient high-fitness DNA design, while other methods~\cite{yang2025steering,zhuang2025moleditrl} optimize non-differentiable experimental rewards through RL fine-tuning with KL regularization to maintain prior proximity.
Similarly, VIDD~\cite{su2025iterative} achieves off-policy optimization via value-weighted likelihood distillation.
Furthermore, antibody design frameworks such as BetterBodies~\cite{vogt2024betterbodies} integrate LDMs with offline RL, guided by binding-oriented rewards for experimentally validated outputs.
Collectively, these studies demonstrate that RL can expose non-differentiable biological objectives while retaining the broad support of pretrained sequence priors.

More recent literature extends this paradigm along two complementary directions.
DoMinO~\cite{su2026discrete} reformulates discrete flow matching (DFM) as a continuous-time MDP policy and introduces distributional regularization to limit drift from the pretrained generators.
RIDER~\cite{hu2026rider} instead aligns RNA sequences with target three-dimensional structures via self-consistency rewards.
These works suggest that progress in biological sequence design increasingly relies on jointly optimizing function, structural validity, and sequence naturalness, rather than maximizing a single proxy reward.


\subsection{Continuous 3D Structure Generation}
\label{sec:Continuous 3D Structure Generation}
For continuous 3D molecular design, RL converts physics- and property-based evaluators into trajectory-level supervision.
Representative approaches include RLPF~\cite{zhou2025guiding}, which applies PPO fine-tuning to diffusion trajectories with force field feedback for stable conformations.
And for multi-target compounds with desired properties, MDRL~\cite{yuan20253d} jointly employs composite scores (\textit{e.g.}, ligand efficiency, docking score, and LogP) to steer pretrained models towards pharmacologically viable regions, while Chen et al.~\cite{chen2025uncertainty} extends this by integrating uncertainty-aware reward shaping to balance conflicting objectives and avoid high-risk chemical spaces.
Building on this foundation, FTDiff~\cite{lin2026fine} introduces GRPO and compresses sampling for efficient pocket-conditioned ligand design, while DePPA~\cite{xue2026fine} applies denoising policy optimization to a pocket-aware diffusion prior with multi-property reward signals.
In short, the emerging direction is thus not simply stronger reward maximization, but efficient and uncertainty-aware post-training that balances binding quality, drug-likeness, synthesizability, and diversity.

\subsection{Periodic Crystal Structure Generation}
\label{sec:Periodic Structure Generation}
Crystal generation would impose additional geometric constraints, as the composition, lattice parameters, atomic coordinates, and periodic boundaries must remain jointly valid.
Therefore, existing methods have attempted to introduce RL for stable, low-energy outputs while maintaining the crystallographic symmetry.
Specifically, MatInvent~\cite{chen2025matinvent} uses policy gradient optimization with reward-weighted KL regularization and diversity filtering to enhance success rates for low-formation-energy crystals.
RLFEF~\cite{huang2025reinforcement} reframes RL fine-tuning as imposing physical priors, using off-policy RL to bias sampling towards stable structures while provably preserving the space-group symmetry.
Both methods utilize formation energy as the reward signal along with KL regularization and experience replay to prevent mode collapse.

Recent work has extended this paradigm to continuous-time flow baselines.
OMatG-IRL~\cite{hollmer2026open} performs inference-time policy optimization directly over velocity-field generators, achieving energy-guided exploration without explicit score recovery.
CrystalGRPO~\cite{su2026crystalgrpo} further introduces joint coordinate-lattice policies and coverage-aware optimization, showing the improvements in Top-1 quality conflict with the Top-K candidate coverage required for polymorph discovery. These motivate crystal-specific evaluations that jointly report physical quality, target recovery, and candidate-set diversity, rather than relying on formation energy alone.

\subsection{Continuous Physics-Constrained Field Modeling}
\label{sec:Continuous Fields}
In continuous physics-constrained field modeling, reward design must reconcile empirical data fidelity with governing equations and boundary conditions.
Specifically, PIRF~\cite{yuan2025pirf} treats PDE residuals and boundary conditions as trajectory-level rewards for lower physical residuals with fewer steps, while RTB-DiffInv~\cite{scimeca2025solving} uses off-policy distribution matching to adapt unconditional diffusion priors to uncertainty-aware, conditional inverse-problem posteriors.

Closely studies further broaden this view: PCFT~\cite{tauberschmidt2026physics} directly fine-tunes flow-matching models with differentiable weak-form constraints, whereas PBFM~\cite{baldan2026physics} treats distributional accuracy and physical consistency as competing objectives on a Pareto frontier.
Although PCFT and PBFM are not conventional policy-gradient methods, together these works reveal three complementary routes for physics alignment: on-policy reward optimization, off-policy posterior matching, and direct differentiable constraint optimization.


\subsection{Scientific Imaging and Inverse Reconstruction}
\label{sec:Scientific Imaging}
Scientific imaging involves two distinct targets: the generative reconstruction and the measurement-acquisition policy.
Cold Diffusion~\cite{shen2024learning} belongs to the former, replacing Gaussian corruption with physically meaningful \textit{k}-space degradation for MRI reconstruction without RL.
ASMR~\cite{yen2024adaptive} belongs to the latter, learning an RL policy that selects diagnostically informative \textit{k}-space measurements while bypassing image reconstruction. They should therefore be viewed as complementary baselines rather than a unified RL-diffusion framework.
RL has also begun to optimize medical diffusion generators directly.
Specifically, RL4Med-DDPO~\cite{saremi2025rl4med} aligns dermoscopic image generation with clinical attributes, while recent multi-scale reward learning~\cite{tian2026optimizing} extends PPO-based diffusion fine-tuning to 3D volumetric MRI via combining global anatomical and local slice-level feedback.

\section{Conclusion}
This survey presents a comprehensive review of RL-driven DMs, systematically mapping the technical landscape across four critical research domains: diffusion-based LLMs, multimodal frameworks, robotic systems, and scientific modeling.
By centering on the post-training era, we emphasize pivotal dimensions, including test-time scaling, experience-driven adaptation, autonomous exploration, and reward mechanism design, which collectively redefine the integration paradigm of RL and DMs.
Our core objective is to synthesize foundational principles, identify underexplored research frontiers, and propose actionable pathways for future investigations, further aiming to catalyze both theoretical innovation and practical deployment in this rapidly advancing field.
Given the dynamic evolution of RL-DM research, this survey may not capture the most recent breakthroughs.
Therefore, we will establish a community-maintained platform leveraging crowdsourcing to ensure real-time updates, fostering continuous growth in this knowledge ecosystem.


\section*{Acknowledgment}
This work was supported by the National Science and Technology Major Project (2023ZD0121403), National Natural Science Foundation of China (NSFC) under Grants 62406161 and 62441615.

{\small
\bibliographystyle{IEEEtran}
\bibliography{ref}
}
\vspace{-40pt}
\begin{IEEEbiography}[{\includegraphics[width=0.6in,height=0.78in,clip,keepaspectratio]{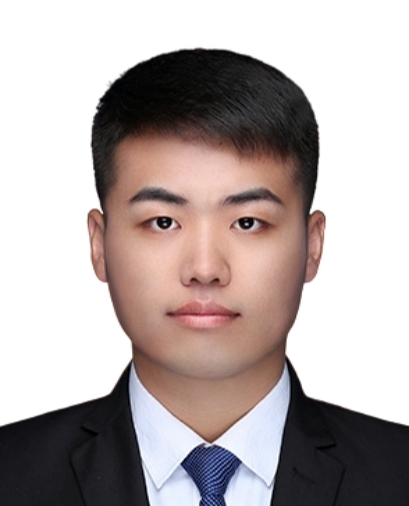}}]
{Zhiyuan Ma} received Ph.D. degree from the Huazhong University of Science and Technology (HUST) in 2023. He is currently an Associate Professor with the School of Computer Science and Technology, HUST. He completed his Postdoctoral Research with Tsinghua University in 2025. His research interests include AIGC, generative models, and multimodal large models. He received the 2025 Xiaomi Young Scholar Program, the  2026 CCF-Tencent RAGR Fund, the 2025 CCF-Kuaishou Large Model Explorer Fund, and the 2024 National Natural Science Foundation (NSFC) Youth Project.
\end{IEEEbiography}
\vspace{-30pt}
\begin{IEEEbiography}[{\includegraphics[width=0.6in,height=0.78in,clip,keepaspectratio]{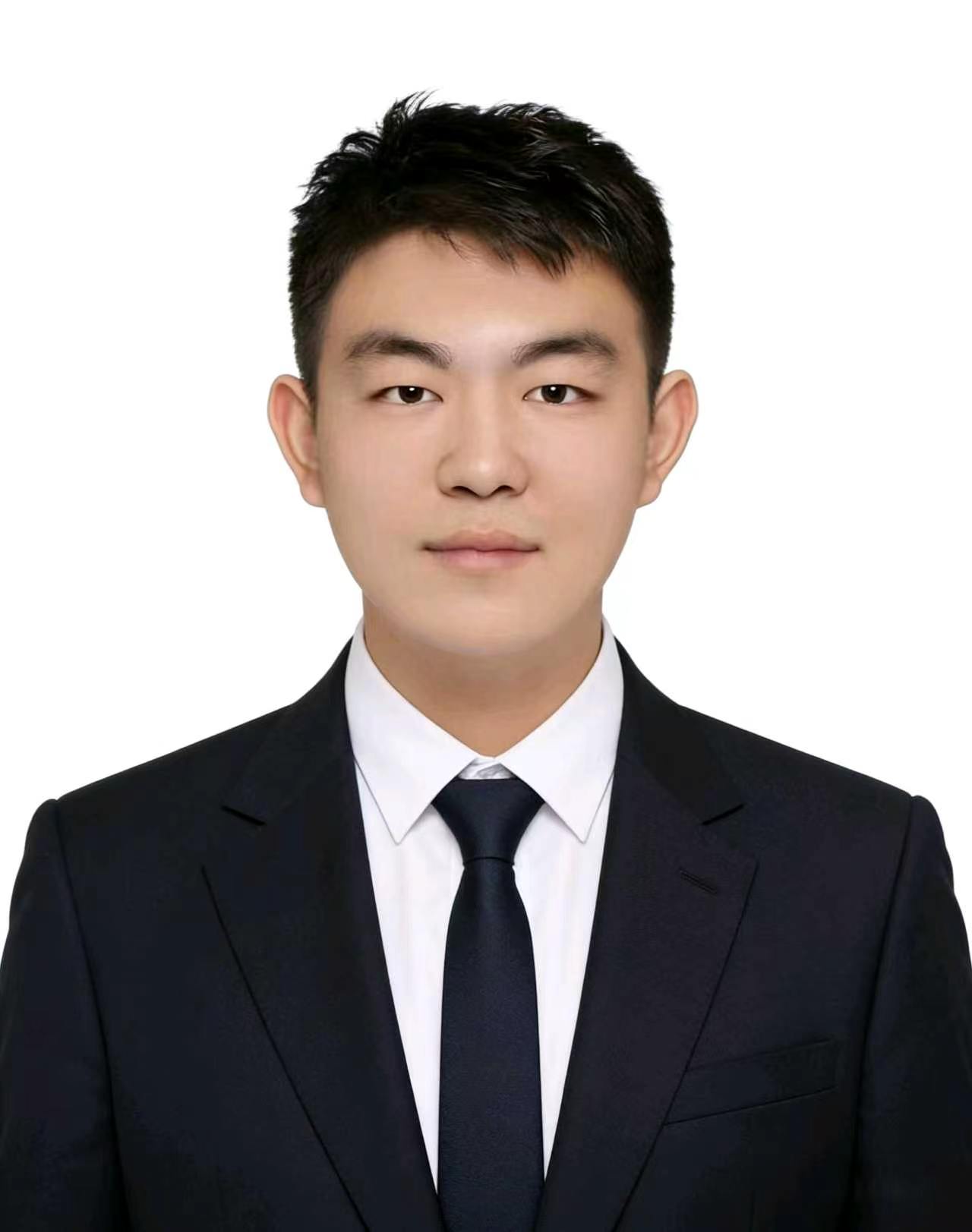}}]
{Jiabao Wei} received the bachelor’s degree from the School of Computer Science \& Technology of Beijing Institute of Technology (BIT) in 2024. He is currently working toward the third-year master’s degree with BIT, majoring in Computer Technology. He has interned at TsinghuaC3I, Shanghai AI Lab, Frontis AI, and HUST-MAIR. His research interests include AIGC and AGI4Science.
\end{IEEEbiography}
\vspace{-30pt}
\begin{IEEEbiography}[{\includegraphics[width=0.6in,height=0.78in,clip,keepaspectratio]{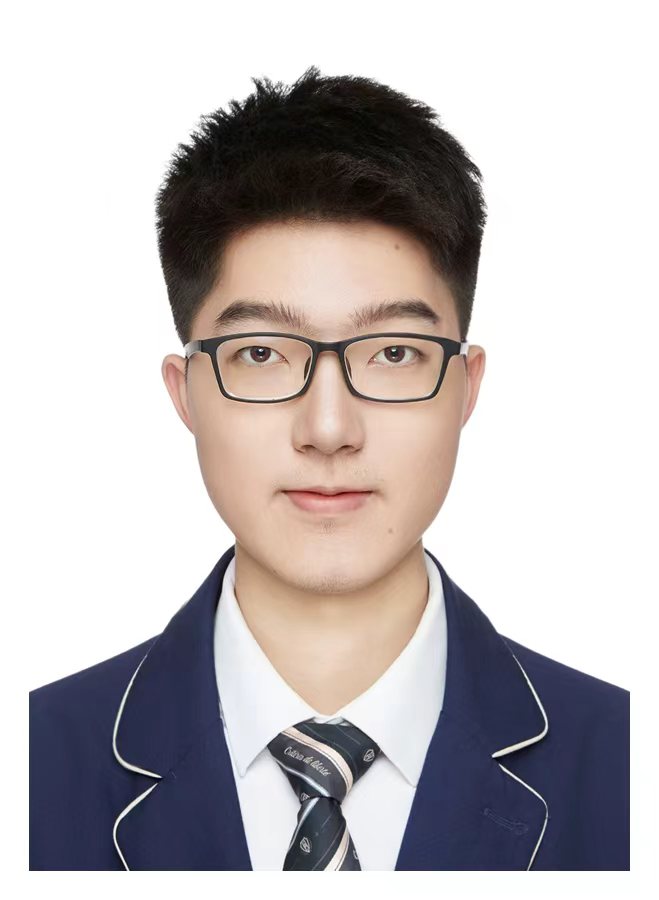}}]
{Tianyu Wang} is currently a first-year PhD in IIIS Institute, Tsinghua University. He received his bachelor's degree in the honor program of John Hopcroft Class, Zhiyuan College, Shanghai Jiao Tong University, in 2026. His research interest mainly contains Large Language Models, Agents, and  their applications to downstream tasks. His representative work includes MegaAgent, and Agents' Last Exam.
\end{IEEEbiography}
\vspace{-30pt}
\begin{IEEEbiography}[{\includegraphics[width=0.6in,height=0.78in,clip,keepaspectratio]{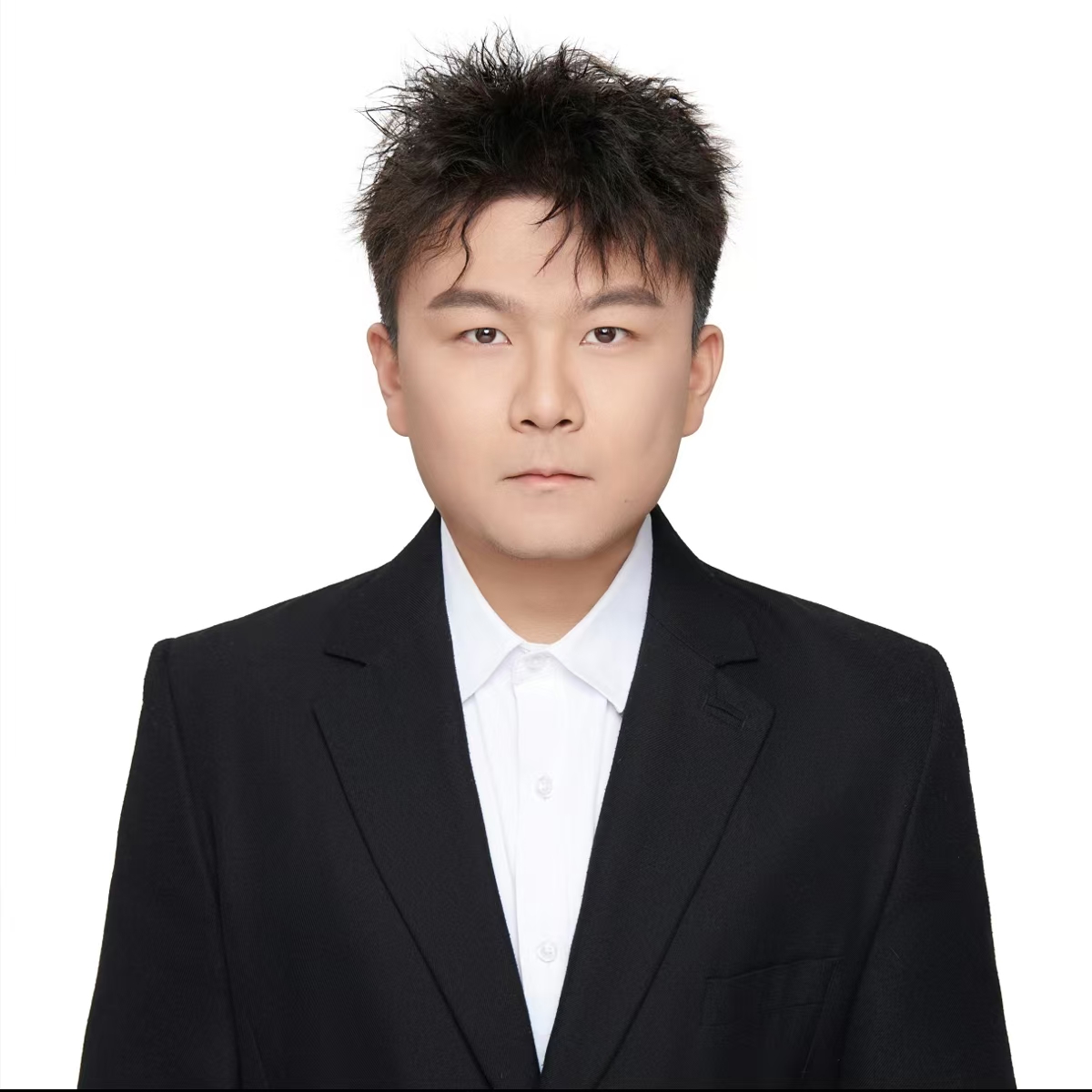}}]
{Hanqing Wang} received his bachelor’s degree from the School of Artificial Intelligence and Automation at Huazhong University of Science and Technology. He is currently an MPhil student and an incoming PhD student at the Hong Kong University of Science and Technology (Guangzhou), advised by Prof. Hui Xiong. His research interests include 3D vision, multimodal large language models, and generative world models for embodied intelligence, with the goal of developing generalizable agents that can perceive, reason about, and interact with the physical world.
\end{IEEEbiography}
\vspace{-30pt}
\begin{IEEEbiography}[{\includegraphics[width=0.6in,height=0.78in,clip,keepaspectratio]{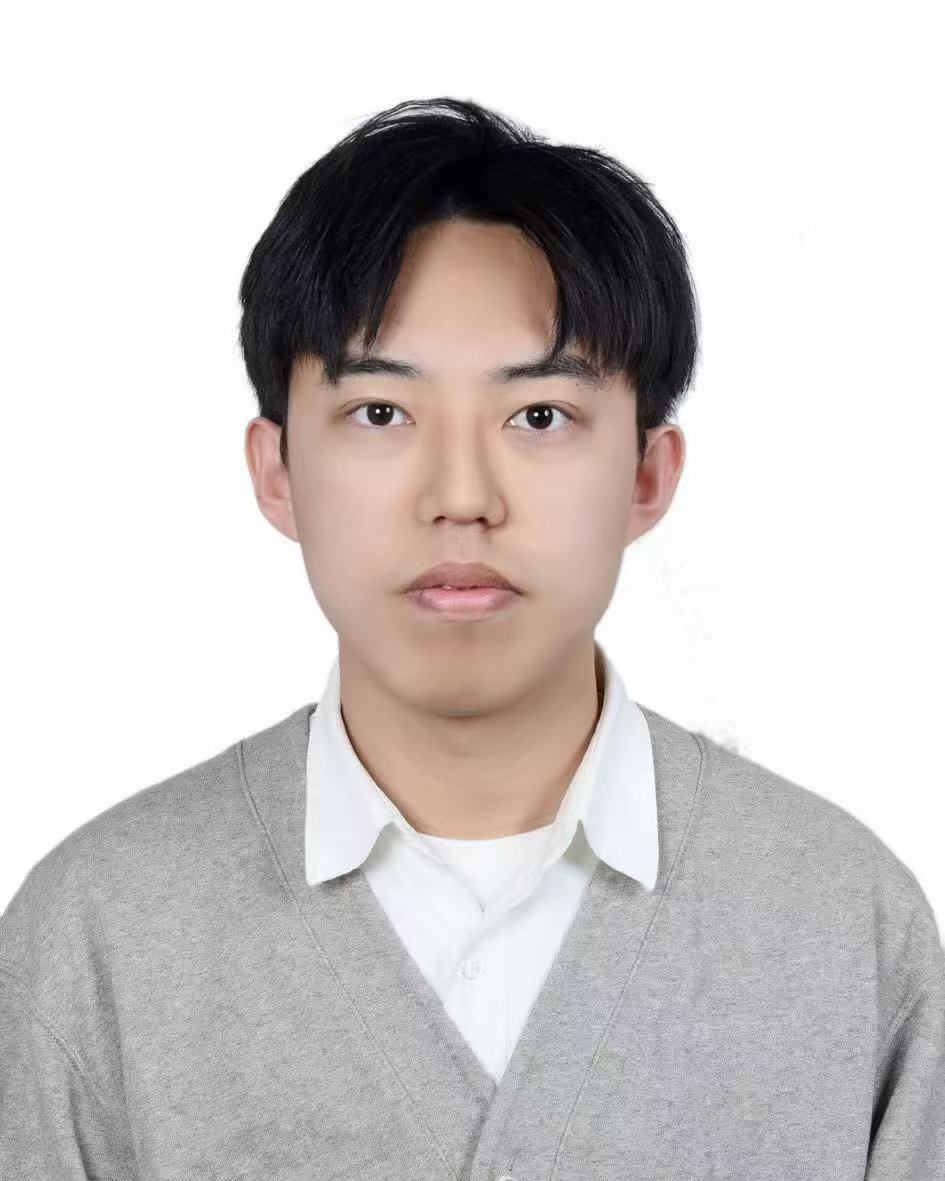}}]
{Jiadong Pan} received the B.S. degree in physics and the M.S. degree in computer science and technology from the University of Chinese Academy of Sciences, Beijing, China, in 2022 and 2026, respectively. He is currently pursuing the Ph.D. degree in computer science and engineering at The Chinese University of Hong Kong, Hong Kong. His research interests include multimodal understanding and generation, embodied intelligence, and robotics.
\end{IEEEbiography}
\vspace{-30pt}
\begin{IEEEbiography}[{\includegraphics[width=0.6in,height=0.78in,clip,keepaspectratio]{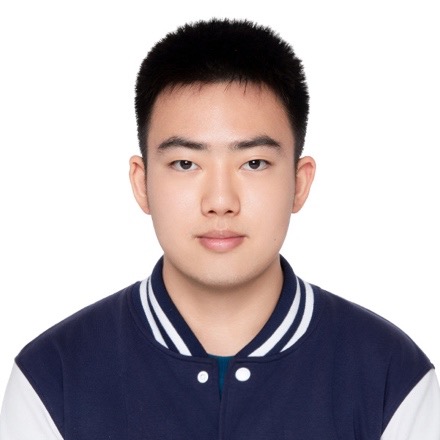}}]
{Haowen Xu} is a third-year undergraduate student at the School of Artificial Intelligence, Shanghai Jiao Tong University. His research interests include multimodal generative models and world models, with a particular focus on controllable video generation and physics-aware video generation.
\end{IEEEbiography}
\vspace{-30pt}
\begin{IEEEbiography}[{\includegraphics[width=0.6in,height=0.78in,clip,keepaspectratio]{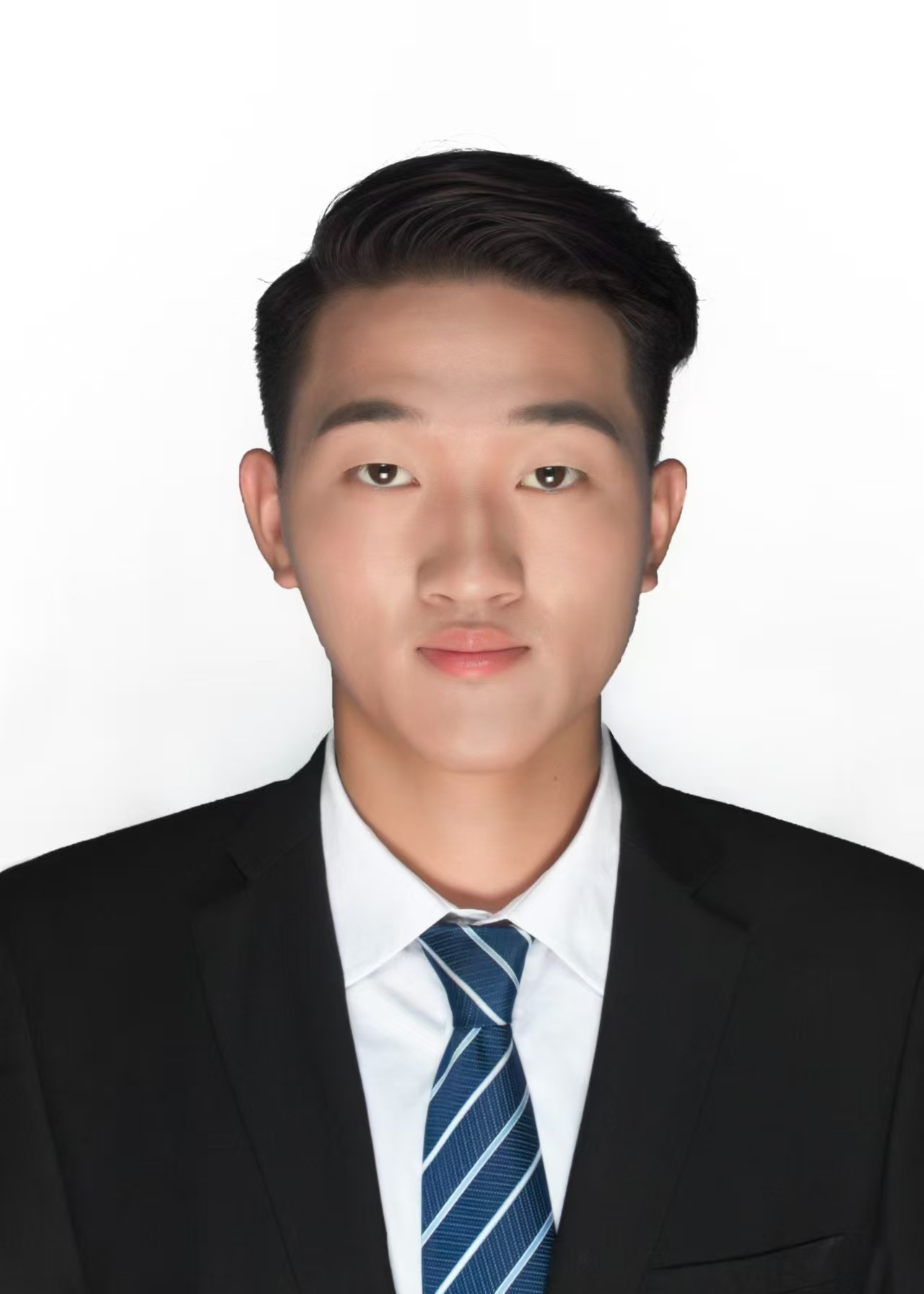}}]
{Liangliang Zhao} received the Master’s degree in Computer Science and Technology from Northwestern Polytechnical University in 2024. He is currently a second-year Ph.D. student in the College of Computer Science and Artificial Intelligence at Fudan University, in a joint training program with Shanghai Artificial Intelligence Laboratory. His research interests include multimodal generation and understanding and reinforcement learning, with a particular focus on developing generative world models and embodied agents capable of perceiving, reasoning, and interacting with the physical world.
\end{IEEEbiography}
\vspace{-30pt}
\begin{IEEEbiography}[{\includegraphics[width=0.6in,height=0.78in,clip,keepaspectratio]{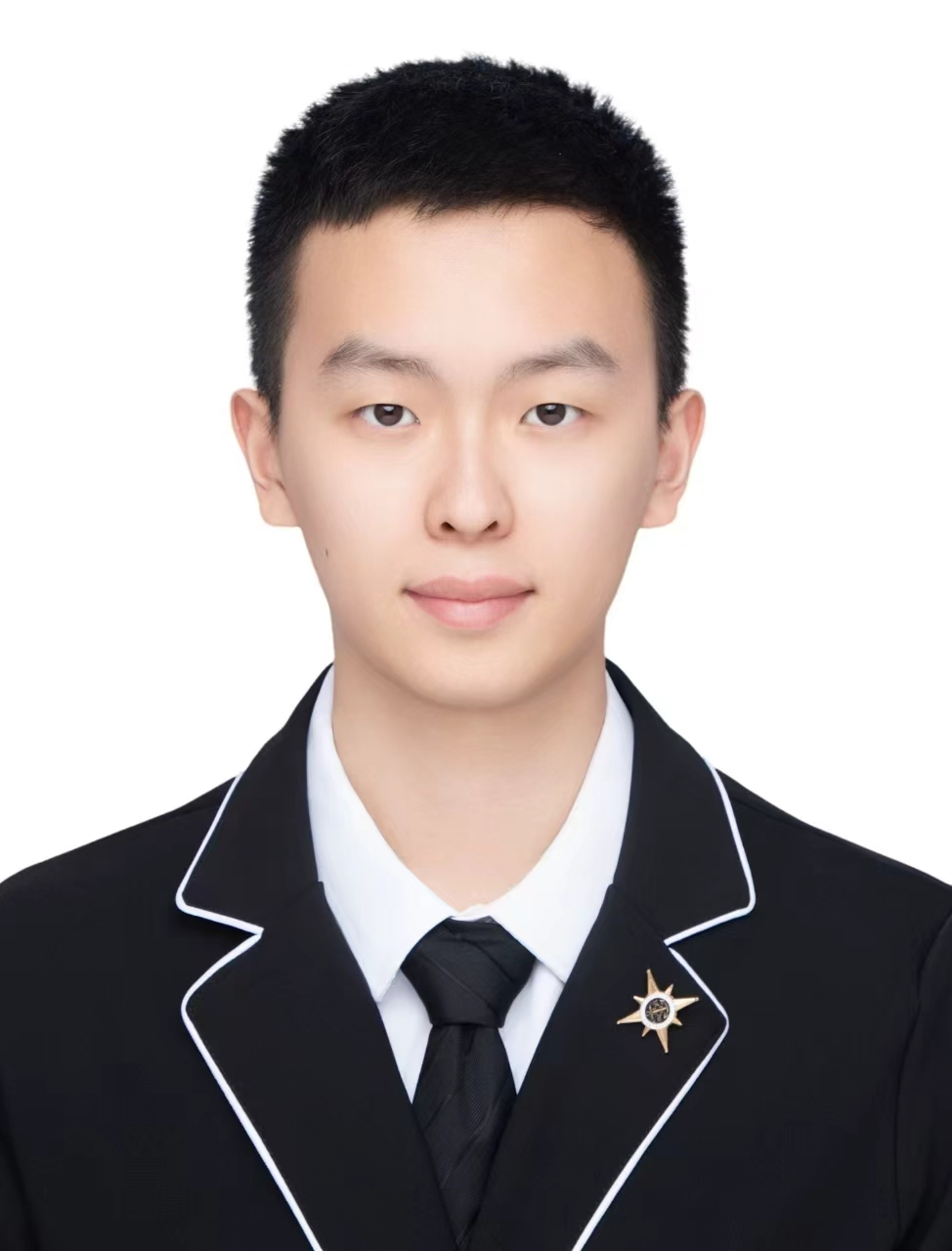}}]
{Ruixun Liu} received the B.Eng. degree in automation from Xi’an Jiaotong University, Xi’an, China, in 2026. He is currently pursuing the Ph.D. degree at Peking University, Beijing, China. His research interests include multimodal large language models and embodied artificial intelligence.
\end{IEEEbiography}
\vspace{-30pt}
\begin{IEEEbiography}[{\includegraphics[width=0.6in,height=0.78in,clip,keepaspectratio]{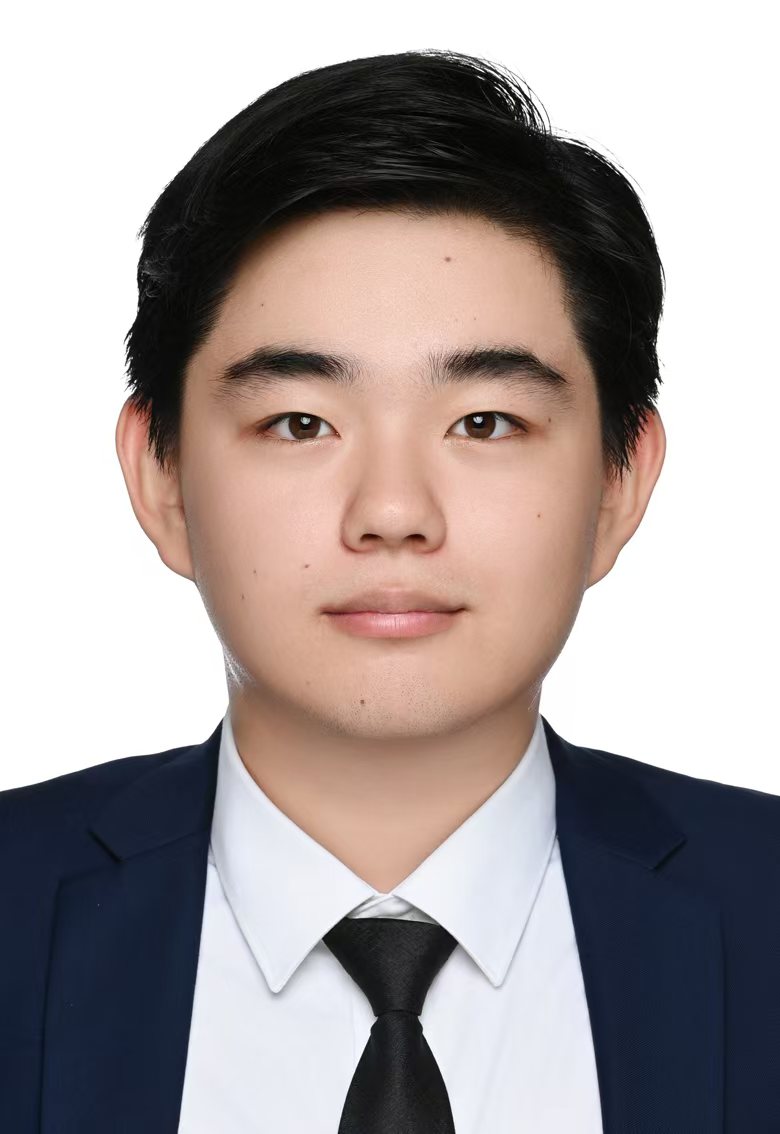}}]
{Yuhan Wang} received the Bachelor’s degree in Mathematics and Applied Mathematics from Tsinghua University’s Zhi Li College. He is currently a first-year Master’s student in Artificial Intelligence at Tsinghua Shenzhen International Graduate School. His research interests include computer vision, reinforcement learning for multimodal models, world models, and human-computer interaction.
\end{IEEEbiography}
\vspace{-30pt}
\begin{IEEEbiography}[{\includegraphics[width=0.6in,height=0.78in,clip,keepaspectratio]{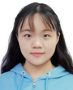}}]
{Yuwei Zhang} was a graduate student at Tongji University during this work. She is currently a first-year Ph.D. student in the Department of Computer Science and Engineering at The Hong Kong University of Science and Technology. Her research interests include computer vision, video generation, and world models.
\end{IEEEbiography}
\vspace{-30pt}
\begin{IEEEbiography}[{\includegraphics[width=0.6in,height=0.78in,clip,keepaspectratio]{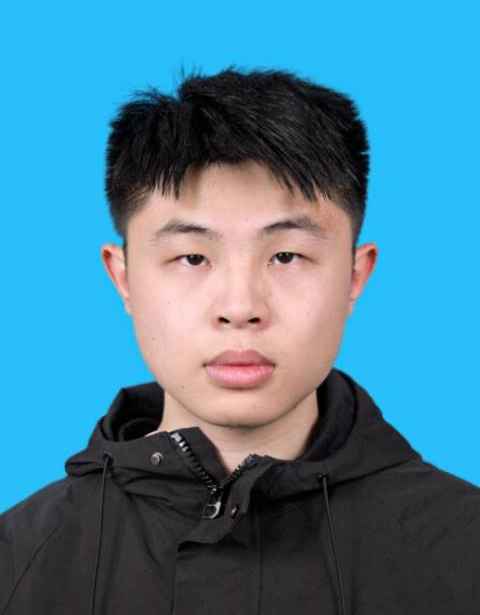}}]
{Liang Feng} received the Bachelor's degree in Economics and Management from UESTC. He is currently a third-year Master's student in Economics at Fudan University. His research interests include computer vision, agent.
\end{IEEEbiography}
\vspace{-30pt}
\begin{IEEEbiography}[{\includegraphics[width=0.6in,height=0.78in,clip,keepaspectratio]{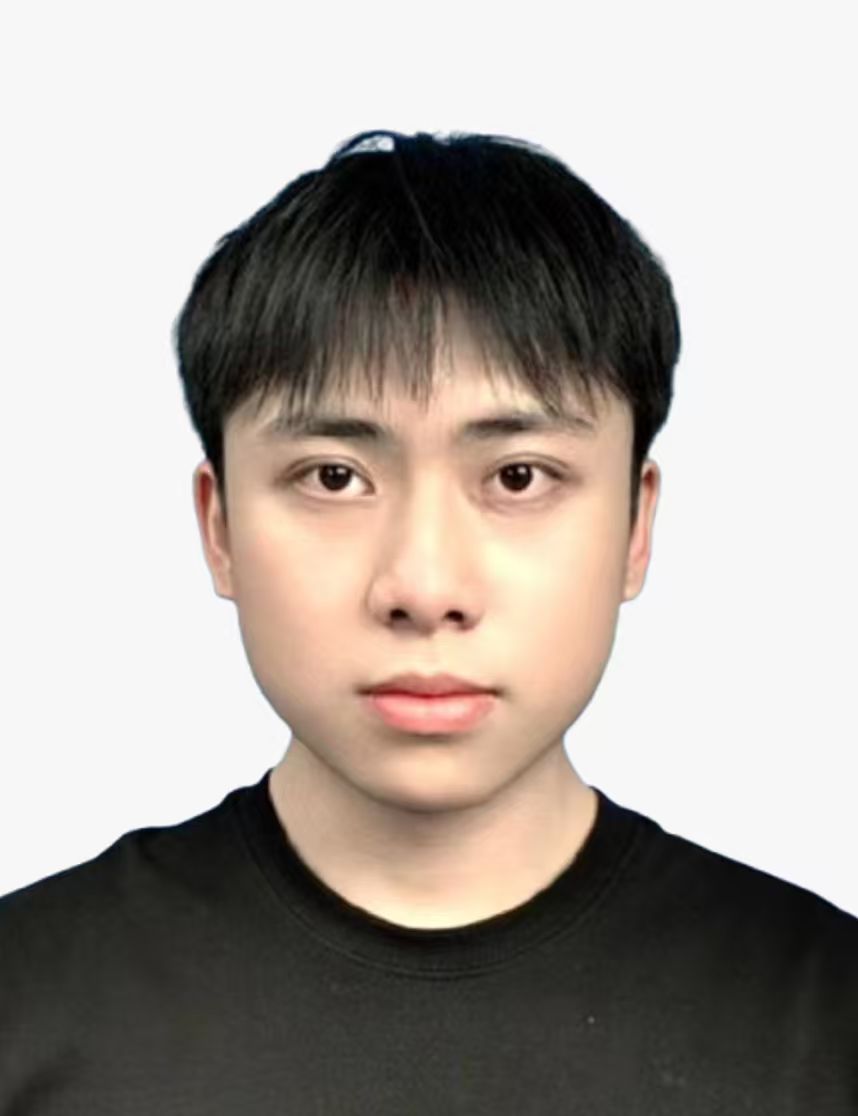}}]
{Junhao Xiao} received the M.S. degree in computer technology from Central China Normal University, Wuhan, China, in 2026. He is currently pursuing the Ph.D. degree in information and communication engineering at the Future Information Innovation Institute, Fudan University, Shanghai, China. His research interests include multimodal understanding and reasoning, medical world models, and image generation and editing.
\end{IEEEbiography}
\vspace{-30pt}
\begin{IEEEbiography}[{\includegraphics[width=0.6in,height=0.78in,clip,keepaspectratio]{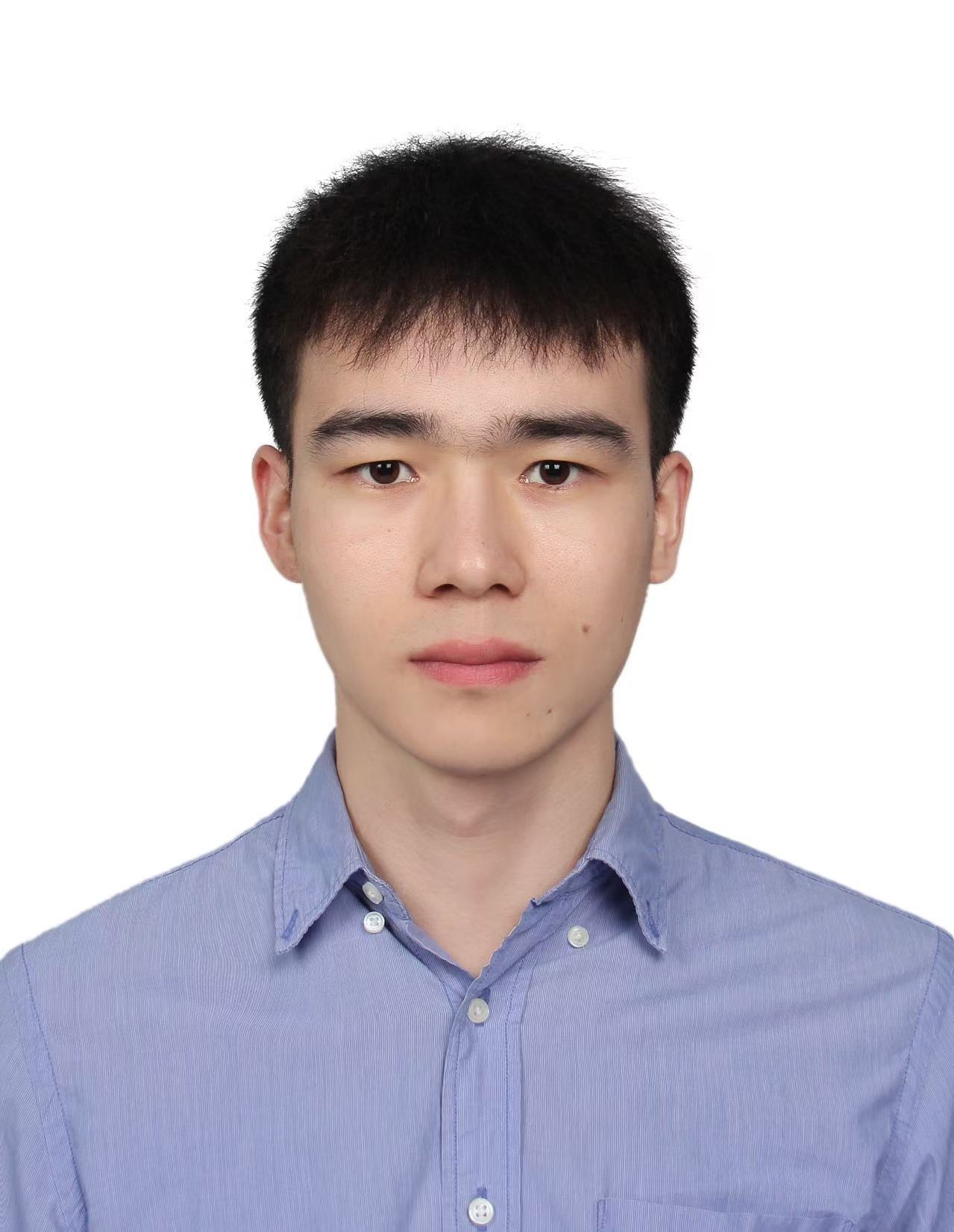}}]
{Kaiyan Zhang} received the B.S. and M.S. degrees in computer science and technology from Harbin Institute of Technology, Harbin, China, in 2020 and 2022, respectively, and the Ph.D. degree in electronic engineering from Tsinghua University, Beijing, China, in 2026. He is currently the Chief Technology Officer of Frontis AI, Beijing, China. His research interests include large language models, reinforcement learning, multi-agent systems, and self-evolving agents.
\end{IEEEbiography}
\vspace{-30pt}
\begin{IEEEbiography}[{\includegraphics[width=0.6in,height=0.78in,clip,keepaspectratio]{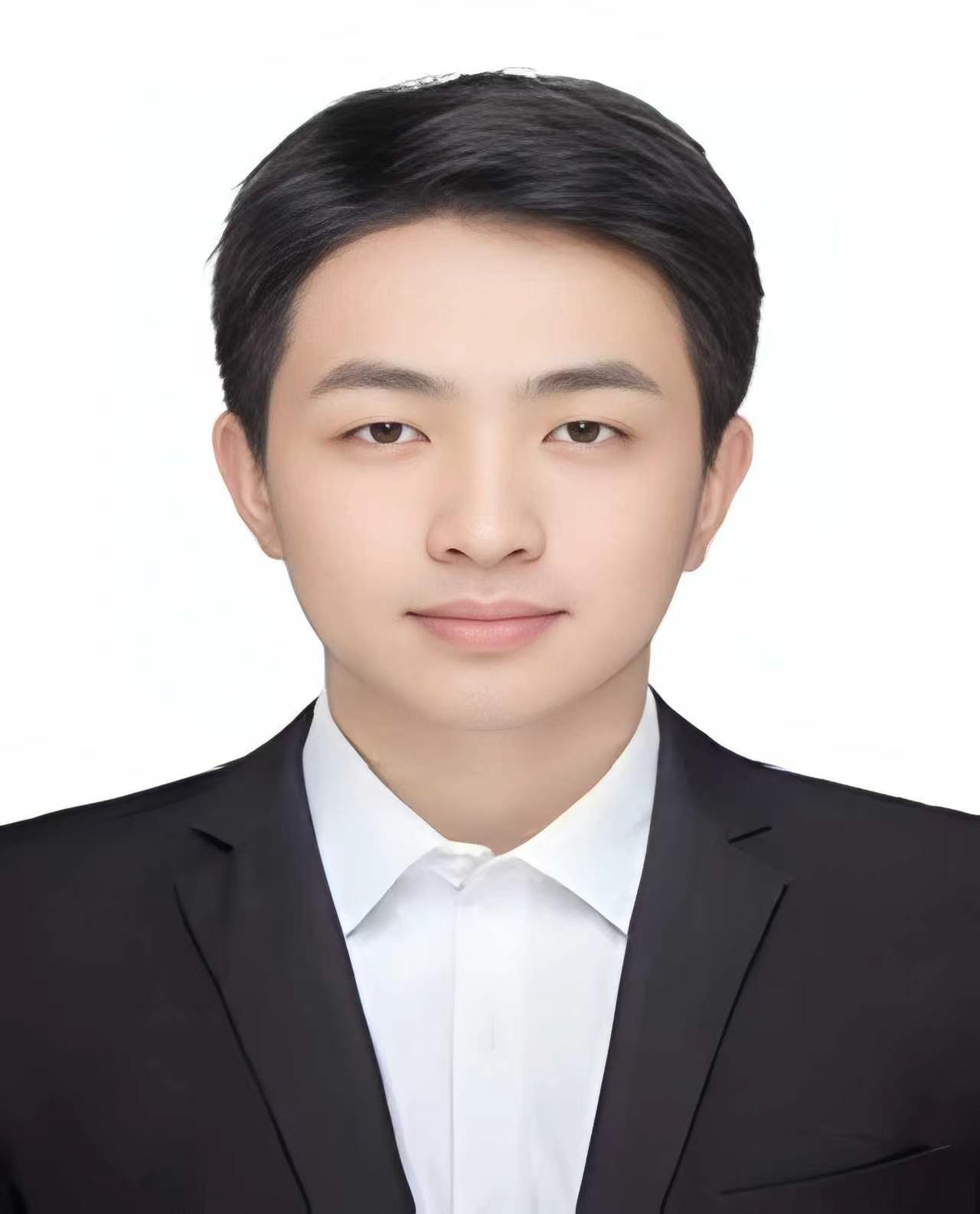}}]
{Xuekai Zhu} is a Ph.D. student in computer science and technology at Shanghai Jiao Tong University, Shanghai, China. His research interests include large language models, reinforcement learning, and self-evolving agents.
\end{IEEEbiography}
\vspace{-30pt}
\begin{IEEEbiography}[{\includegraphics[width=0.6in,height=0.78in,clip,keepaspectratio]{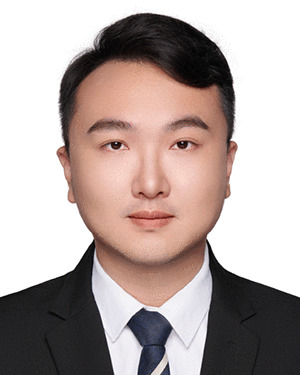}}]
{Daizong Liu} is currently a Tenure-track Assistant Professor in the Institute for Math \& AI at
Wuhan University. He received the M.S. degree
in Electronic Information and Communication of
Huazhong University of Science and Technology in
2021, and the Ph.D. degree at Wangxuan Institute of
Computer Technology of Peking University in 2025.
His research interests include 3D adversarial attacks,
multi-modal learning, LVLM robustness, etc. He has
published more than 40 papers in refereed conference proceedings and journals such as TPAMI, NeurIPS, CVPR, ICCV, ECCV, SIGIR, AAAI, ACL. He regularly serves on
the program committees of top-tier AI conferences such as NeurIPS, ICML, ICLR, CVPR, ICCV and ACL.
\end{IEEEbiography}
\vspace{-30pt}
\begin{IEEEbiography}[{\includegraphics[width=0.6in,height=0.78in,clip,keepaspectratio]{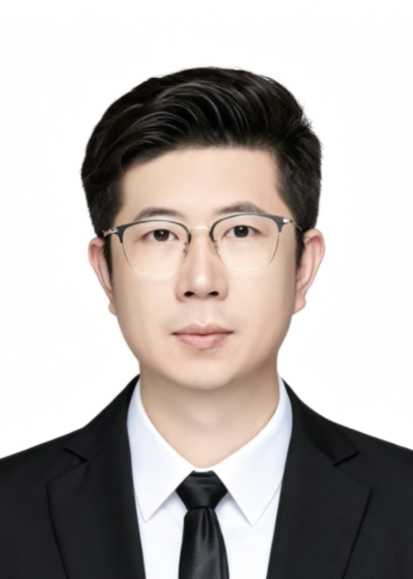}}]
{Biqing Qi} is currently a researcher at Shanghai AI Lab and a postdoctoral fellow at the University of Hong Kong. He received his Ph.D. from Harbin Institute of Technology, jointly supervised by Tsinghua University. His research interests include machine learning and natural language processing. He has published over 70 papers in top-tier conferences and journals, including NeurIPS, CVPR, ACL, and TPAMI, and has led four projects, including national major projects and Shanghai major projects.
\end{IEEEbiography}
\vspace{-30pt}
\begin{IEEEbiography}[{\includegraphics[width=0.6in,height=0.78in,clip,keepaspectratio]{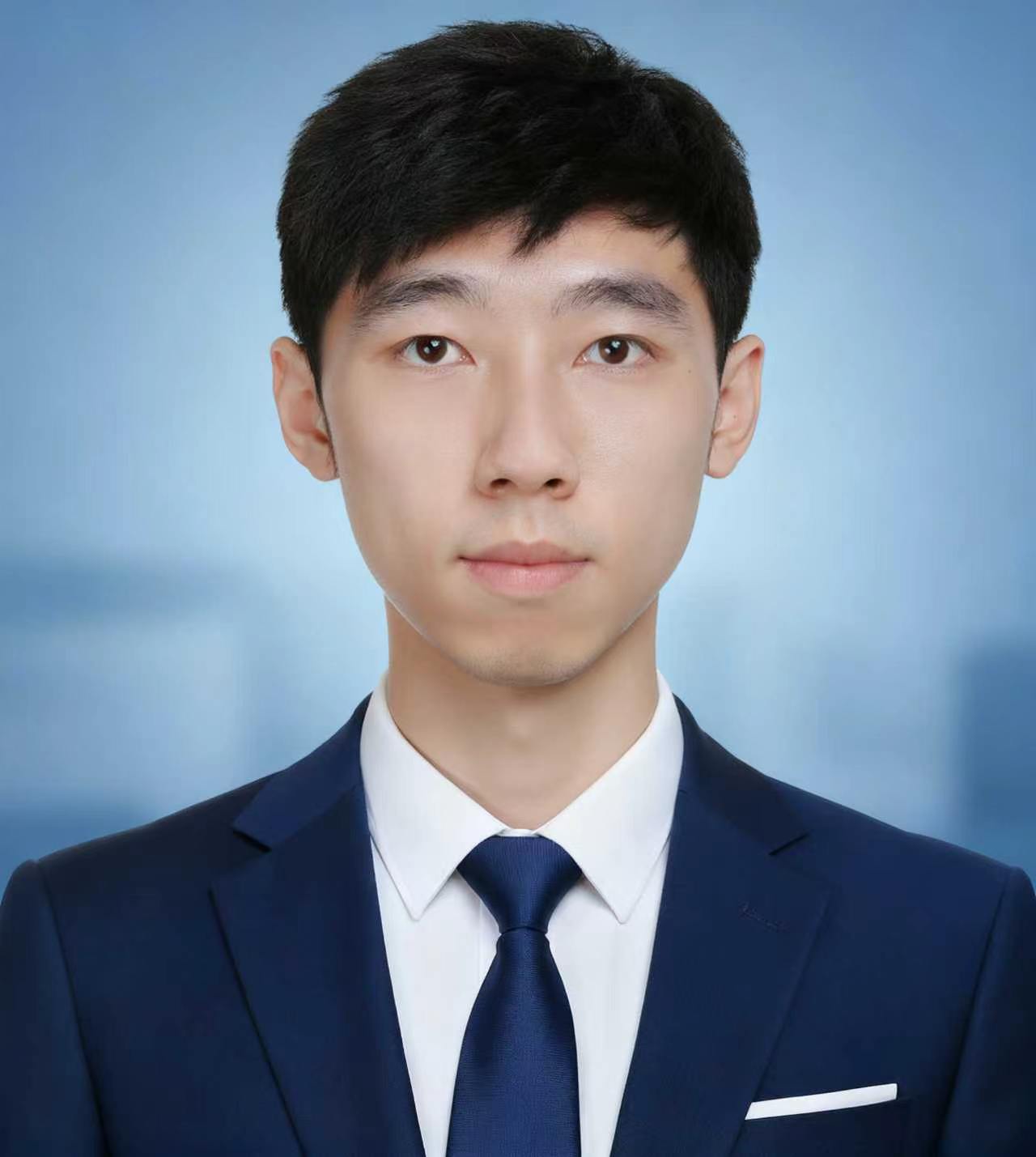}}]
{Yihao Liu} received the B.S. and Ph.D. degrees from the University of Chinese Academy of Sciences, Beijing, China. His doctoral research was conducted at the Shenzhen Institutes of Advanced Technology, Chinese Academy of Sciences. He is currently a Research Scientist with the Shanghai Artificial Intelligence Laboratory, where he leads the Multimodal Frontier Team. His current research focuses on building unified multimodal models that jointly support understanding, generation, reasoning, and self-correction. He has led the development of several unified multimodal models that bridge visual understanding and generation within a single framework. His research has been published in leading conferences and journals, including IEEE TPAMI, CVPR, ICCV, ECCV, ICML, ICLR, and NeurIPS. He has received several honors, including the Top Talent Program of the Shanghai Eastern Talent Plan, the Chinese Academy of Sciences President's Award, and the Zhu Li Yue Hua Outstanding Doctoral Student Award.
\end{IEEEbiography}
\vspace{-30pt}
\begin{IEEEbiography}[{\includegraphics[width=0.6in,height=0.78in,clip,keepaspectratio]{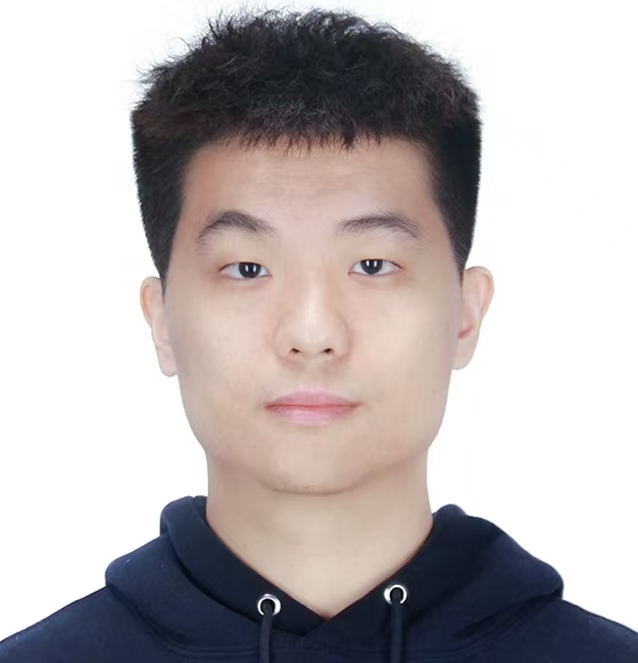}}]
{Ganqu Cui} is a Research Scientist at Shanghai Artificial Intelligence Laboratory. He received his Ph.D. from the Department of Computer Science and Technology at Tsinghua University. His research focuses on large language model alignment and reinforcement learning. He recieved the WAIC Yunfan Rising Star Award, the Tsinghua University Outstanding Doctoral Dissertation Award, and the Tsinghua University Outstanding Graduate Award.
\end{IEEEbiography}
\vspace{-30pt}
\begin{IEEEbiography}[{\includegraphics[width=0.6in,height=0.78in,clip,keepaspectratio]{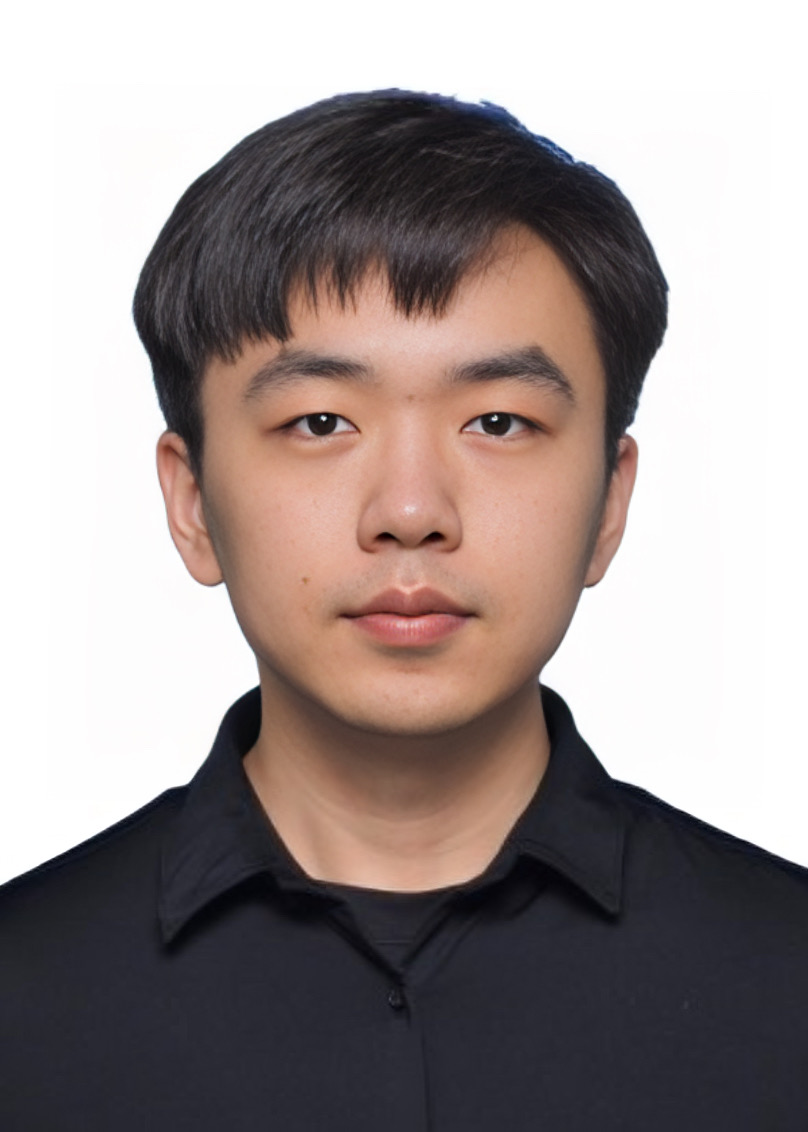}}]
{Ning Ding} is an Assistant Professor in the Department of Electronic Engineering at Tsinghua University, working on large language models, reinforcement learning, and embodied intelligence. He earned his Ph.D. from the Department of Computer Science and Technology at Tsinghua University, and has received the ACL Best System Demonstration Paper Award, the WAIC Youth Outstanding Paper Award, and Tsinghua University’s Outstanding Doctoral Dissertation Award. 
\end{IEEEbiography}
\vspace{-30pt}
\begin{IEEEbiography}[{\includegraphics[width=0.6in,height=0.78in,clip,keepaspectratio]{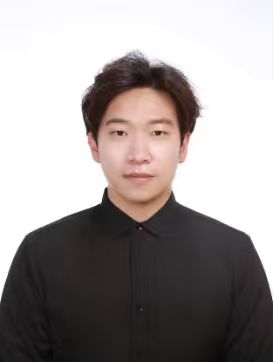}}]
{Linfeng Zhang} is an Assistant Professor in the School of Artificial Intelligence at Shanghai Jiao Tong University, working on efficient AI and the pre-training and post-training of large models. He earned his Ph.D. from the Institute for Interdisciplinary Information Sciences at Tsinghua University, and has served as an Area Chair for ICLR, NeurIPS, ACL, EMNLP, and other conferences.
\end{IEEEbiography}
\vspace{-30pt}
\begin{IEEEbiography}[{\includegraphics[width=0.6in,height=0.78in,clip,keepaspectratio]{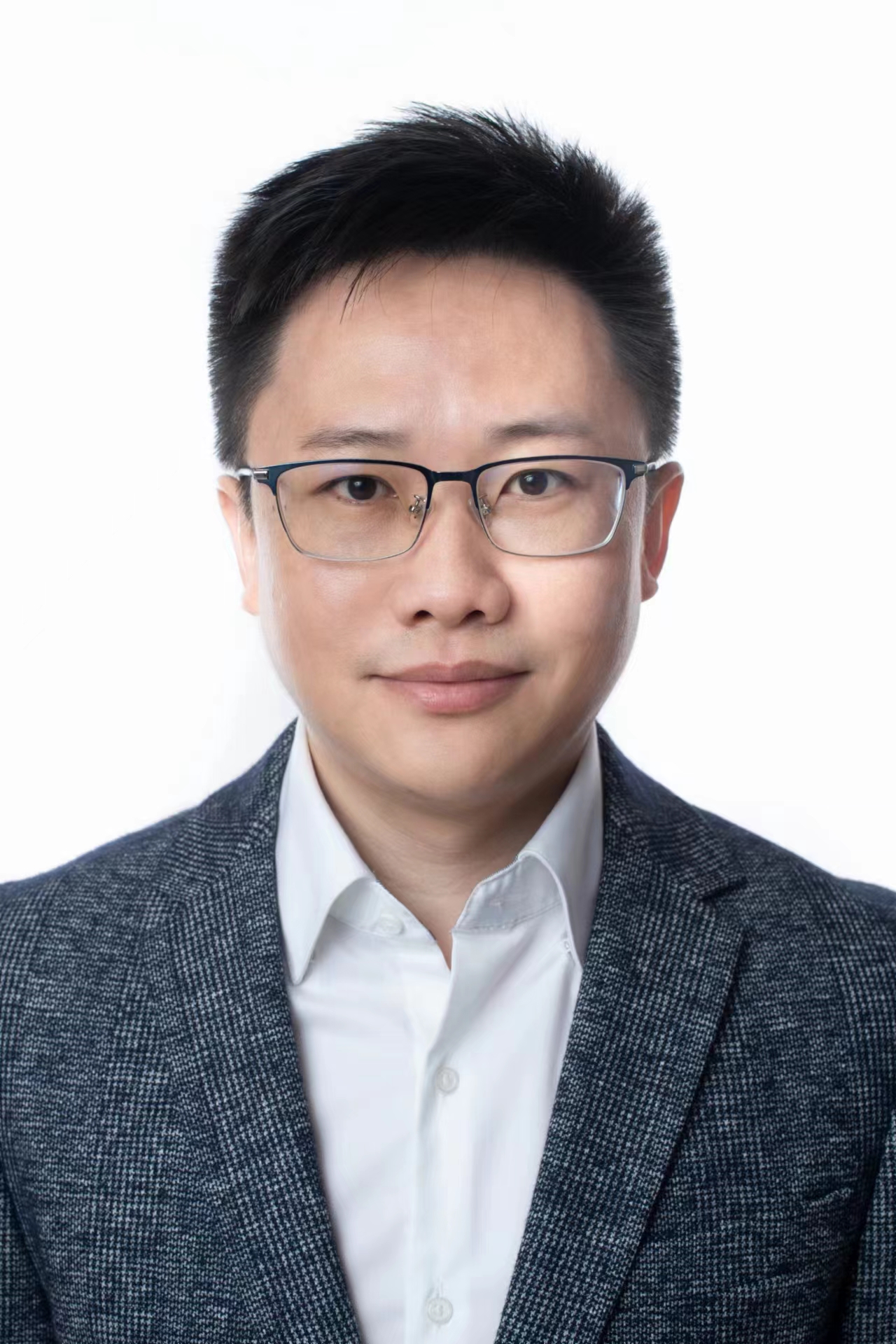}}]
{Jianjun Li} received the Ph.D. degree in computer science from Huazhong University of Science and Technology (HUST), Wuhan, China in 2012. He was a Senior Research Associate with the Department of Computer Science, City University of Hong Kong. He is currently a full professor at the School of Computer Science and Technology in HUST. His research interests include recommender systems, natural language processing, cyber-physical systems, and advanced data management. 
\end{IEEEbiography}
\vspace{-30pt}
\begin{IEEEbiography}[{\includegraphics[width=0.6in,height=0.78in,clip,keepaspectratio]{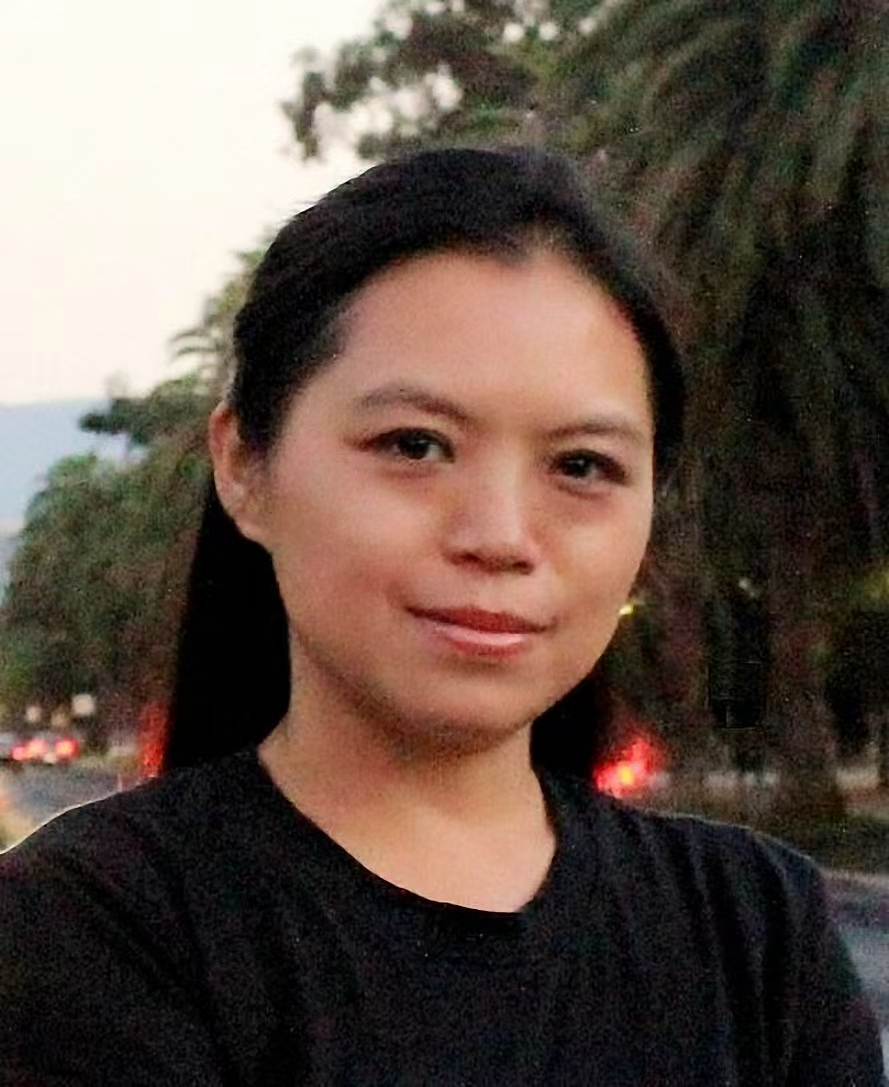}}]
{Kun He} (SM18) received the Ph.D. degree in system engineering from Huazhong University of Science and Technology, Wuhan, China, in 2006. She is currently a Professor in School of Computer Science and Technology, Huazhong University of Science and Technology, Wuhan, China. She had been with the Department of Management Science and Engineering at Stanford University in 2011-2012 as a visiting researcher. 
She had been with the department of Computer Science at Cornell University in 2013-2015 as a visiting associate professor, in 2016 as a visiting professor, and in 2018 as a visiting professor.  She was honored as a Mary Shepard B. Upson visiting professor for the 2016-2017 Academic year in Engineering, Cornell University, New York. Her research interests include machine learning, deep learning, and optimization methods.
\end{IEEEbiography}
\vspace{-30pt}
\begin{IEEEbiography}[{\includegraphics[width=0.6in,height=0.78in,clip,keepaspectratio]{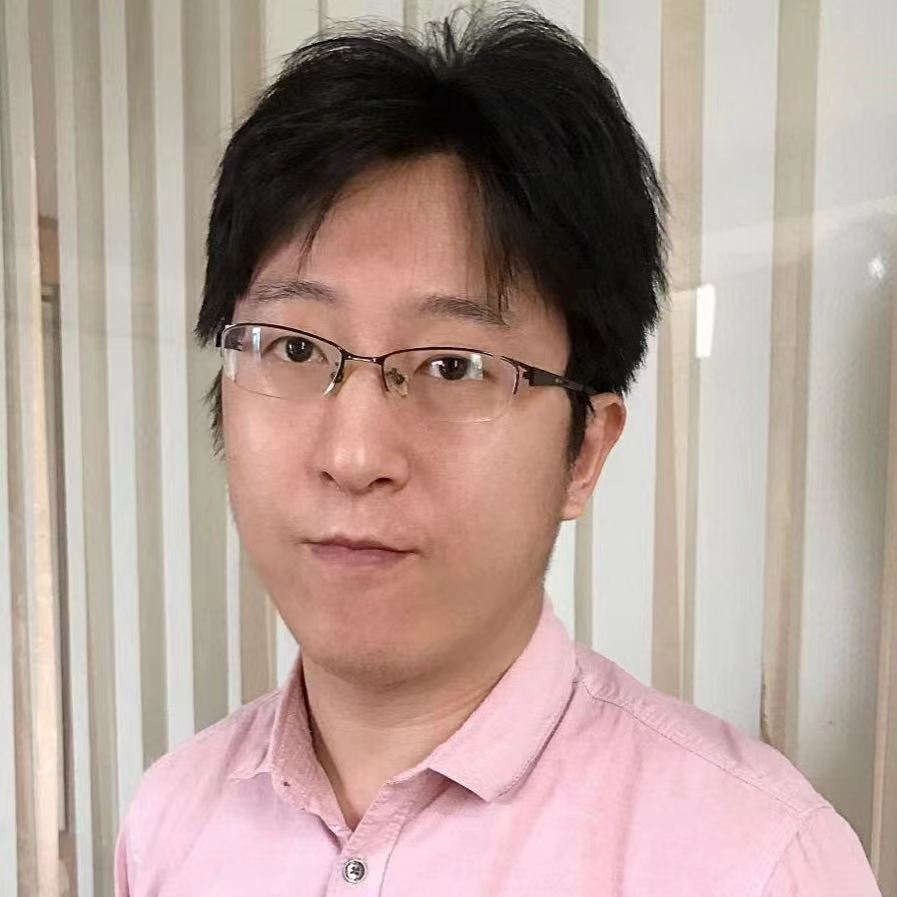}}]
{Xiang Bai} (Fellow, IEEE) received his B.S., M.S., and Ph.D. degrees from the Huazhong University of Science and Technology (HUST), Wuhan, China, in 2003, 2005, and 2009, respectively. He is currently a Professor with the School of Software Engineering, HUST. He is also the Vice-director of the National Center of AntiCounterfeiting Technology, HUST. His research interests include OCR, LMM, autonomous driving, and world model. He serves as an associate editor for IEEE Transactions on Pattern Analysis and Machine Intelligence (TPAMI) and as an Associate Editor-in-Chief for Pattern Recognition.
\end{IEEEbiography}
\vspace{-30pt}
\begin{IEEEbiography}[{\includegraphics[width=0.6in,height=0.78in,clip,keepaspectratio]{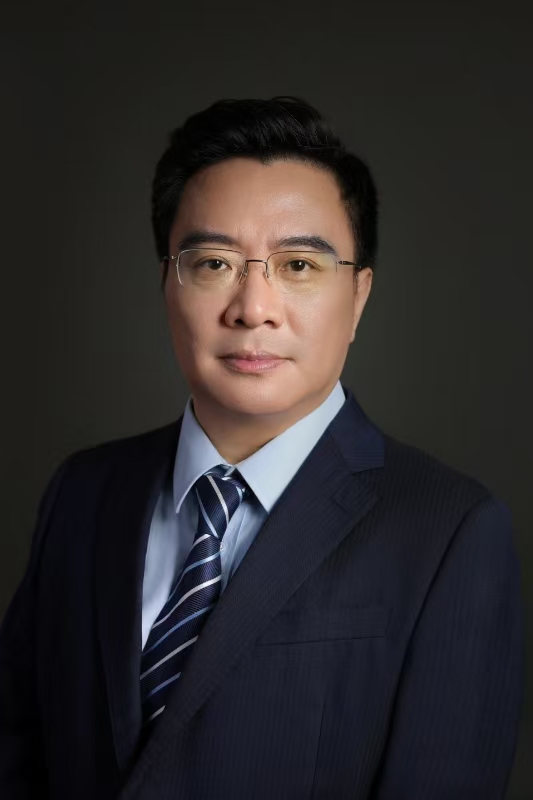}}]
{Bowen Zhou} (Fellow, IEEE) is the director \& Chief Scientist of Shanghai AI Lab, Huiyan Chair Professor of Department of Electrical Engineering at Tsinghua University, IEEE/CAAI Fellow, the chief scientist of the scientific and technological innovation 2030 “New Generation Artificial Intelligence” major project. Prior to that, he was the Director of AI Foundations of IBM T. J. Watson Research Center, Chief Scientist of IBM Watson Group, a Distinguished Engineer of IBM, SVP of JD.Com, the Chairman of JD.Com’s Technology Committee, and the President of JD Cloud \& AI BU. Dr. Zhou won various influential awards including “Wu Wenjun Artificial Intelligence Science and Technology Award - Outstanding Contribution Award”. 
\end{IEEEbiography}

\end{document}